\documentclass{MITcsail}

\usepackage{amsmath,amsfonts,bm}

\def\eqref#1{equation~\ref{#1}}

\def\1{\bm{1}}

\DeclareMathAlphabet{\mathsfit}{\encodingdefault}{\sfdefault}{m}{sl}
\SetMathAlphabet{\mathsfit}{bold}{\encodingdefault}{\sfdefault}{bx}{n}

\usepackage{amsmath}
\usepackage{amsfonts}
\usepackage{amssymb}
\usepackage{wrapfig}
\usepackage{subcaption}
\usepackage{multirow}
 \usepackage{mathtools} 

\usepackage{verbatim}

\usepackage{anyfontsize}

\usepackage{microtype}
\usepackage{graphicx}
\usepackage{booktabs} 
\usepackage{multirow}
\usepackage{amsmath,amssymb}
\usepackage{booktabs}
\usepackage{caption,subcaption}

\usepackage{xcolor}
\definecolor{mygreen}{HTML}{3cb44b}
\definecolor{skyblue}{HTML}{beffff}
\definecolor{lightgreen}{HTML}{90ee90}

\usepackage{color, colortbl}

\definecolor{emerald}{rgb}{0.31, 0.78, 0.37}

\usepackage{tcolorbox}
\usepackage{enumitem}
\usepackage{colortbl}

\usepackage{xcolor}
\definecolor{mygreen}{HTML}{3cb44b}
\colorlet{myyellow}{green!10!orange!90!}
\makeatletter

\usepackage{tikz}
\usetikzlibrary{arrows,shapes,snakes,automata,backgrounds,fit,petri}
\usepackage{adjustbox}

\newcommand{\RN}[1]{%
	\textup{\lowercase\expandafter{\it \romannumeral#1}}%
}
\usepackage{tabu}

\newcommand{\beq}{\vspace{0mm}\begin{equation}}
\newcommand{\eeq}{\vspace{0mm}\end{equation}}
\newcommand{\beqs}{\vspace{0mm}\begin{eqnarray}}
\newcommand{\eeqs}{\vspace{0mm}\end{eqnarray}}
\newcommand{\barr}{\begin{array}}
\newcommand{\earr}{\end{array}}

\usepackage{color, colortbl}
\definecolor{Gray}{gray}{0.93}

\usepackage{lipsum}

\usepackage{pifont}

\usepackage{makecell}

\usepackage{xcolor,amsmath}
\usepackage[linesnumbered,ruled,vlined]{algorithm2e}
\DontPrintSemicolon

\usepackage{xcolor}
\definecolor{mygreen}{HTML}{3cb44b}

\SetKwComment{Comment}{\color{green!50!black}\# }{}

\SetKwProg{Function}{def}{:}{}

\SetKwProg{For}{for}{:}{}
\SetKwProg{If}{if}{:}{}

\usepackage{hyperref}
\hypersetup{
    colorlinks=true,
    linkcolor=orange,
    citecolor=lightgray,
    filecolor=darkred,
    urlcolor=orange
}
\usepackage[numbers, sort&compress]{natbib}

\usepackage{enumerate}
\usepackage{colortbl}
\usepackage{pgf} 
\usepackage{float}
\usepackage{epsfig}
\usepackage{graphicx}
\usepackage{amsmath}
\usepackage{amssymb}
\usepackage{xcolor}
\usepackage{multirow}
\usepackage{multicol}
\usepackage[utf8]{inputenc}

\definecolor{darkred}{RGB}{140, 21, 21}
\definecolor{lightgray}{gray}{0.7}
\definecolor{orange}{HTML}{F58025}

\newcommand{\data}{\textbf{Video-MMLU}}
\newcommand{\benchmark}{\textbf{Video-MMLU}}

\definecolor{small}{RGB}{255, 255, 255} 
\definecolor{big}{RGB}{148, 153, 192} 
\newcommand{\rgbrank}[2]{
  \pgfmathsetmacro{\percent}{%
    \ifnum#2<10
      0 + #2
    \else
      \ifnum#2<30
        10 + (#2-10) * 0.5
      \else
        \ifnum#2<45
          20 + (#2-30) * 1.33
        \else
          50 + (#2-45) * 0.91
        \fi
      \fi
    \fi
  }%
  \edef\temp{\noexpand\cellcolor{big!\percent!small}}\temp
  \temp #1%
}

\title{Video-MMLU: A Massive Multi-Discipline Lecture Understanding Benchmark}

\author
{Enxin Song~$^{1}$, Wenhao Chai~$^{2,\dagger}$, Weili Xu~$^{1,3}$, Jianwen Xie~$^{4}$, Yuxuan Liu~$^{3}$, Gaoang Wang~$^{1,*}$\\
\vspace{1em}
\normalfont{\small $^{1}$ Zhejiang University}\\
\normalfont{\small $^{2}$ University of Washington}\\
\normalfont{\small $^{3}$ University of Illinois Urbana-Champaign}\\
\normalfont{\small $^{4}$ Lambda, Inc.}\\
\vspace{1em}
\texttt{Link: 
\href{https://enxinsong.com/Video-MMLU-web/}{Project Page}
~|~
\href{https://huggingface.co/datasets/Enxin/Video-MMLU}{Dataset}
~|~
\href{https://github.com/Espere-1119-Song/Video-MMLU/tree/main}{Code}}
\vspace{1em}
}

\footnotetext{Project Lead.}
\footnotetext{Corresponding author.}

\begin{document}

\maketitle
\thispagestyle{firstpagestyle}

   \begin{center}
    \centering
    \includegraphics[width=\textwidth]{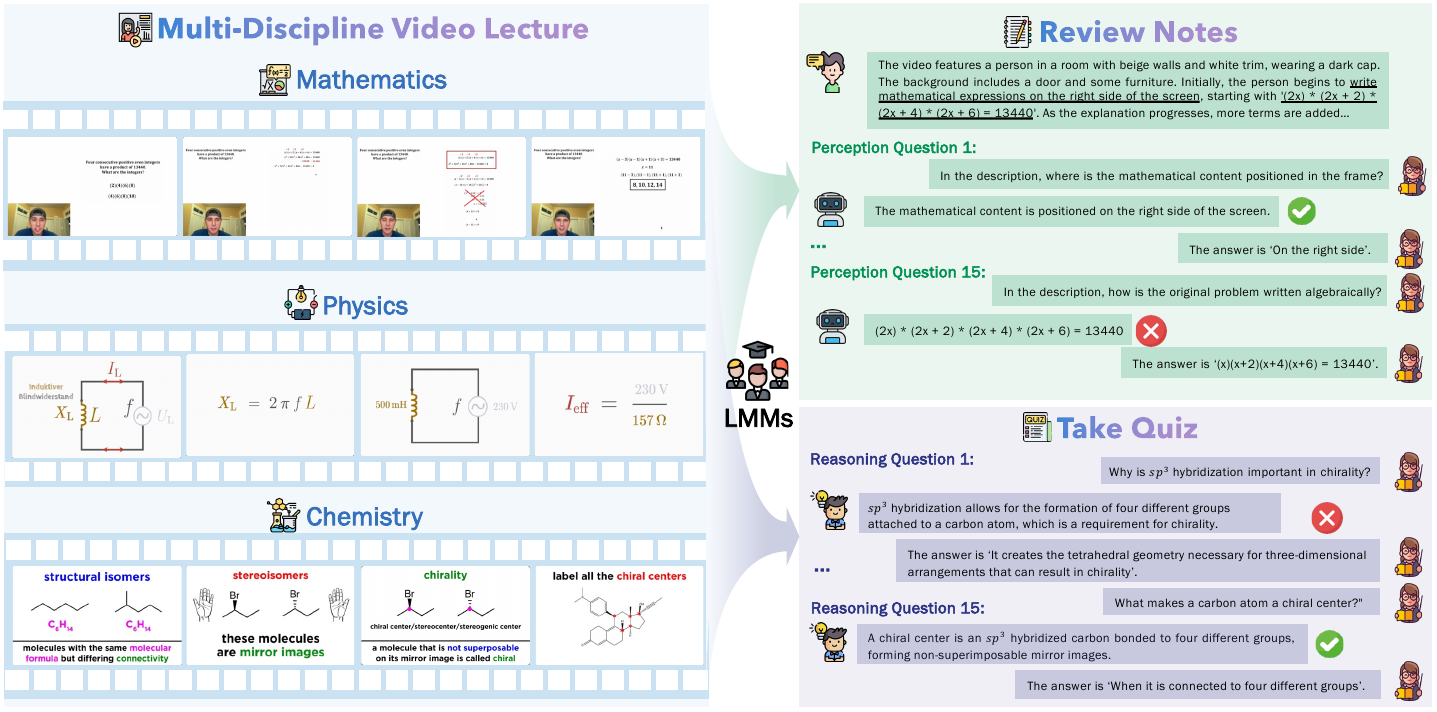}
    \captionof{figure}{Overview of~\benchmark.
The benchmark includes multi-discipline lecture videos in mathematics, physics, and chemistry, featuring theorem demonstrations and problem-solving. Evaluation consists of: (1)\textbf{Review Notes}, where models generate detailed video captions to assess visual perception, and (2)\textbf{Take Quiz}, where models answer reasoning questions to test comprehension.}
    \label{fig:teaser}
   \end{center}%

\begin{abstract}
Recent advancements in language multimodal models (LMMs) for video have demonstrated their potential for understanding video content, yet the task of comprehending multi-discipline lectures remains largely unexplored. We introduce Video-MMLU, a massive benchmark designed to evaluate the capabilities of LMMs in understanding Multi-Discipline Lectures. We evaluate over 90 open-source and proprietary models, ranging from 0.5B to 40B parameters. Our results highlight the limitations of current models in addressing the cognitive challenges presented by these lectures, especially in tasks requiring both perception and reasoning. Additionally, we explore how the number of visual tokens and the large language models influence performance, offering insights into the interplay between multimodal perception and reasoning in lecture comprehension.
\end{abstract}
\vspace{24pt}
\clearpage

\section{Introduction}

Large Language Models (LLMs)~\cite{cai2024internlm2, bai2025qwen2, qwq32b} have demonstrated remarkable capabilities in encoding vast amounts of world knowledge~\cite{wei2024measuring}. At the same time, many Large Multimodal Models (LMMs) leverage LLMs for multimodal understanding of images, videos, and even audio, inheriting the world knowledge and reasoning capabilities of LLMs. In the video domain, numerous benchmarks have emerged in recent years to evaluate various aspects of model capabilities, such as general video understanding~\cite{zhou2024mlvu, li2024mvbench}, long video understanding~\cite{wu2024longvideobench, mangalam2024egoschema, fu2024video, song2023moviechat, song2024moviechat+}, and detailed video captioning~\cite{chai2024auroracap, wei2025longcaptioning, kim2024hicm}. These benchmarks derive their data from various sources including movies, daily activities, and short-form videos. They all evaluate different aspects of video understanding capabilities, such as temporal reasoning and visual-linguistic alignment.

Despite advances in performance of LMMs on artificially constructed benchmarks, numerous real-world use cases, particularly those involving knowledge-intensive or reasoning-heavy content, remain insufficiently tested. Previous benchmarks~\cite{wang2024lvbench, xiao2021next, xu2016msr} have predominantly focused on simple factual questions about video content, leaving a significant gap in questions and problems which are knowledge-intensive or require strong reasoning capabilities. 
Educational videos, specifically lecture content across academic disciplines, pose an especially challenging domain for multimodal understanding. These videos contain dense information through text, equations, and visual demonstrations that require both strong perception capabilities and domain-specific reasoning. The ability to process and comprehend such content would significantly advance the practical applications of LMMs in educational settings. Existing visual knowledge reasoning benchmarks~\cite{zou2024dynamath, yue2023mmmu, jiang2025mme} are limited to static images and inadequate for assess dynamic problem-solving, evolving visuals, and continuous reasoning in real-world educational scenarios.  

To address this gap, we introduce~\data, a video-based massive multi-discipline lecture understanding benchmark. Unlike previous benchmarks that focus on general video content, \data~specifically targets lecture videos that involve theorem demonstrations and problem-solving across disciplines including mathematics, physics, and chemistry. This benchmark requires models to not only recognize visual content but also reason through complex educational material. 
We construct~\data~using a multi-stage annotation pipeline that integrates video captions as the backbone and enriches them with frame-level captions. We conduct extensive experiments to evaluate existing LMMs, including vision-blind baselines, proprietary models, open-source LMMs and token-compressed models on captioning and question-answering tasks. Our findings reveal the intricate impact of model type, scale, architecture, and token compression on LMM performance in~\benchmark.

In summary, our contributions are three-fold:

\begin{itemize}
    \item We build~\data, which requires strong reasoning capabilities and world knowledge compared to the previous benchmarks for video LMMs.
    \item We evaluate more than 90 proprietary models and open-source models of varying sizes on~\benchmark. Our findings indicate that existing models generally perform poorly, with accuracy ranging from only 10\% to 50\%.
    \item We explore how the number of visual tokens and the base LLMs influence performance, offering insights into the interplay between multimodal perception and reasoning in lecture comprehension.
\end{itemize}

\begin{table*}[t]
\centering
\renewcommand{\arraystretch}{0.92}
\caption{\textbf{Benchmark comparison} for video understanding tasks. Ave.\ Length indicates the average number of words per caption.}
\resizebox{\textwidth}{!}{%
\begin{tabular}{l| ccc |cccc| cc }
\toprule
\multirow{2}{*}{\textbf{Dataset}} & \multirow{2}{*}{\textbf{Theme}}& \multirow{2}{*}{\textbf{\#~Video}} & \multirow{2}{*}{\textbf{\#~Ave. Duration (s)}} & \multicolumn{4}{c|}{\textbf{Caption}} & \multicolumn{2}{c}{\textbf{Question-answering}} \\
& & & & \textbf{Number} & \textbf{\#~Word} & \textbf{\#~Vocab.} & \textbf{Ave. Length} & \textbf{Number} & \textbf{Type} \\
\midrule
MovieChat-1K~\cite{song2023moviechat} & Movie & 1,000 & 564 & 1,000 & 121,077 & 102,988 & 121 & 13,000 & Open-ended \\
MMWorld~\cite{he2024mmworld} & Professional & 1,910 & 107 & 1,910 & - & - & 66 & 6,627 & Multiple-choice \\
MLVU~\cite{zhou2024mlvu} & Open & 1,730 & 930 & 247 & - & - & -& 3,102 & Multiple-choice \\
MVBench~\cite{abellan2023probing} & Open & 4,000 & 16 & \multicolumn{4}{c|}{\textcolor{red}{\textbf{\texttimes}}} & 4,000 & Multiple-choice \\
LongVideoBench~\cite{wu2024longvideobench} & Open &  3,763 & 473 &  \multicolumn{4}{c|}{\textcolor{red}{\textbf{\texttimes}}} & 6,678 & Multiple-choice \\
TempCompass~\cite{liu2024tempcompass} & Open & 410 & $<30$ & \multicolumn{4}{c|}{\textcolor{red}{\textbf{\texttimes}}} & 7,540 & Multiple-choice\\
Video-MMMU~\cite{hu2025video} & Professional & 300 & 506 & \multicolumn{4}{c|}{\textcolor{red}{\textbf{\texttimes}}} & 900 & Multiple-choice \\
VATEX~\cite{wang2019vatex} & Open & 41,250 & 10 & 41,250 & 4,994,768 & 44,103  & 15 & \multicolumn{2}{c}{\textcolor{red}{\textbf{\texttimes}}} \\
VDC~\cite{chai2024auroracap} & Open & 1,027 & 28 & 1,027 & 515,441 & 20,419 & 501 & \multicolumn{2}{c}{\textcolor{red}{\textbf{\texttimes}}} \\
LongCaptioning~\cite{wei2025longcaptioning} & Open & 10,000 & 93 & 10,000 & - & - & 1,198 & \multicolumn{2}{c}{\textcolor{red}{\textbf{\texttimes}}} \\

\midrule
\benchmark~(ours) & Professional & 1,065 & 109 & 1,065 & 520,679 & 27,613 & 489 & 15,746 & Open-ended \\
\bottomrule
\end{tabular}%
}
\label{tab:benchmark}
\end{table*}

\section{Related Work}
\subsection{Large Multimodal Models for Video}
Early video LMMs~\cite{dai2024instructblip, zhang2023video} primarily focused on understanding short videos within only a few seconds, employing either video-based encoders~\cite{arnab2021vivit, liu2022video} or image-based encoders~\cite{oquab2023dinov2, zhai2023sigmoid}. Subsequent studies~\cite{yin2024t2vid, chen2024sharegpt4video} demonstrated that structurally expanding LLaVA-like image-based LMMs into video-based LMMs, combined with a well-designed training strategy and high-quality training data, can yield strong performance.
Recent efforts~\cite{cheng2024videollama, wu2025longvitu, wang2024longllava, geng2024longvale, liu2024streamchat, zhang2025memory, luo2025quota, luo2024video, jiang2025token, chen2025livecc} has shifted towards long video understanding, targeting hour-long durations or streaming scenarios, either by compressing visual information~\cite{yu2024espresso} or extending the LLM’s context window~\cite{chen2024longvila}. 
Including audio inputs~\cite{fu2025vita, fu2024vita, he2024storyteller} also helps imrpove video understanding. 
Video detailed captioning is another focus~\cite{kim2024hicm, bi2025everything, yu2025eve, chu2025fine}, aiming to generate fine-grained descriptions of video content.
\cite{han2025adafv, chandak2009fastv, zhuang2024st, fu2024framefusion, huang2024prunevid, han2024rethinking} have further explored token reduction to improve training and inference efficiency by designing compression mechanisms within the vision encoder~\cite{wang2025folder}, reducing token counts in the LLM~\cite{zhang2024p}, or introducing additional compression modules~\cite{wang2024retake, yang2024visionzip}.
Additionally, video reasoning~\cite{han2024videoespresso, bigverdi2024perception, qiu2024step} has gained attention, with approaches like temporal grounding~\cite{wang2024timerefine, lu2024llava, guo2024trace} to enhance comprehension. Vlog~\cite{lin2025vlogvideolanguagemodelsgenerative} enhances video understanding by using generative retrieval of a hierarchical narration vocabulary.

\subsection{Benchmarks for Video Understanding}

Previously, video LMMs were evaluated on classical video QA tasks~\cite{xu2016msr, xu2017video} with short question-answer pairs or brief one-sentence video captions, focusing on global questions. However, with the integration of stronger LLMs~\cite{qwen2.5, cai2024internlm2}, these simple benchmarks are no longer sufficient to assess and differentiate model performance.
Recent video understanding benchmarks~\cite{agarwal2024mvtamperbench, zhang2024vinoground, he2024mmworld, cores2024tvbench, choong2024vidhal, shangguan2024tomato, jung2024consistency, hong2025worldsense, ye2025llavaction, fang2024mmbench} have shifted toward longer video durations or more complex temporal reasoning tasks. EgoSchema~\cite{mangalam2023egoschema} involves multiple-choice questions on 3-minute egocentric videos, and LongVideoBench~\cite{wu2024longvideobench} expands to hour-long video understanding.
Streaming video understanding benchmarks~\cite{lin2024streamingbench} are crucial for evaluating models’ real-time processing.
Additionally,~\cite{hu2024can, chai2024auroracap} evaluate models by requiring detailed descriptions of video content. While most benchmarks~\cite{bae2025mash, cao2025video, gao2025exploring, wang2025time} emphasize open-world understanding, others introduce diverse scenarios~\cite{hu2025video, zhou2025humanvbenchexploringhumancentricvideo, lee2022multimodal}, such as movie clips~\cite{yue2023movie101} or dynamic GUI interactions~\cite{zhao2025worldgui, bsharat2025mobile}. Our contribution introduces a video understanding benchmark for lecture comprehension across disciplines, evaluating detailed perception and reasoning through fine-grained captioning and complex QA tasks.

\section{Dataset Construction}


\subsection{Video Collection and Processing}
\label{video_collect}

\benchmark, a \textbf{M}assive \textbf{M}ulti-discipline \textbf{L}ecture \textbf{U}nderstanding benchmark, aims to evaluate the comprehension abilities on multi-discipline lectures of Large Multimodal Models (LMMs). While existing visual knowledge reasoning benchmarks~\cite{zou2024dynamath, lu2021inter, chen2021geoqa, liu2024mmbench, jiang2025mme} are mainly limited to static images, lecture videos offer richer temporal information and pose greater challenges in knowledge representation and reasoning.

To ensure high-quality data collection, we carefully consider video length, content type, and presentation quality. Most video annotation methods~\cite{zhang2024video, wei2025longcaptioning, chen2025sharegpt4video, ju2024miradata} segment long videos into shorter clips and merge segment-based annotations, which may disrupt the continuity of the reasoning flow. As observed by ~\cite{wu2024longvideobench}, proprietary models such as \texttt{GPT-4o} and \texttt{Gemini-1.5-Pro} can process long inputs up to 256 frames, but their performance beyond this limit remains uncertain. To ensure reliable and consistent annotations, we limit video length to 4 minutes. Additionally, some abstract animation lectures rely heavily on subtitles for comprehension, which may not accurately reflect the model’s ability to understand the lecture content. Therefore,~\benchmark~specifically targets videos that focus on theorem demonstrations and probleming-solving. The videos deliver dense information through numbers and formulas, pose significant challenges for video LMMs in dynamic OCR recognition and comprehension.

After retrieving videos via the YouTube Data API, we filter out those lacking transcribed or English subtitles. To ensure an appropriate level of difficulty, we exclude overly static videos or excessively challenging videos requiring strong domain-specific prior knowledge for comprehension. Ultimately, we collected 1,065 videos from 10 YouTube channels, covering mathematics, physics, and chemistry. For keyframe extraction, we employ a customized approach based on video motion pacing. Since videos from the same creator often share similar frame rates, we manually review representative videos per creator to set the optimal sampling rate. Sampling rates typically range from 1 frame per second to 1 frame every 5 seconds, capturing key visual information for annotation and evaluation.


\vspace{-3pt}
\subsection{Annotations Construction Pipeline}
\label{annotation_construction}
Imagine a classroom where a large multimodal model is the \textbf{student} and~\benchmark~acts as the \textbf{teacher}.~\benchmark~evaluates whether the student can perceive and comprehend multi-discipline lectures, much like a student taking notes and being tested later. For each video, we generate a detailed caption as the standard \textbf{``notes"} to assess the model’s visual perception. Additionally, we create 15 questions as a \textbf{``quiz"} to evaluate content reasoning, challenging the model’s ability to apply learned knowledge.

\begin{figure}[htbp]
    \centering
    \begin{minipage}{0.4\textwidth} 
        \centering
        \begin{subfigure}[b]{\textwidth}
            \includegraphics[width=\textwidth]{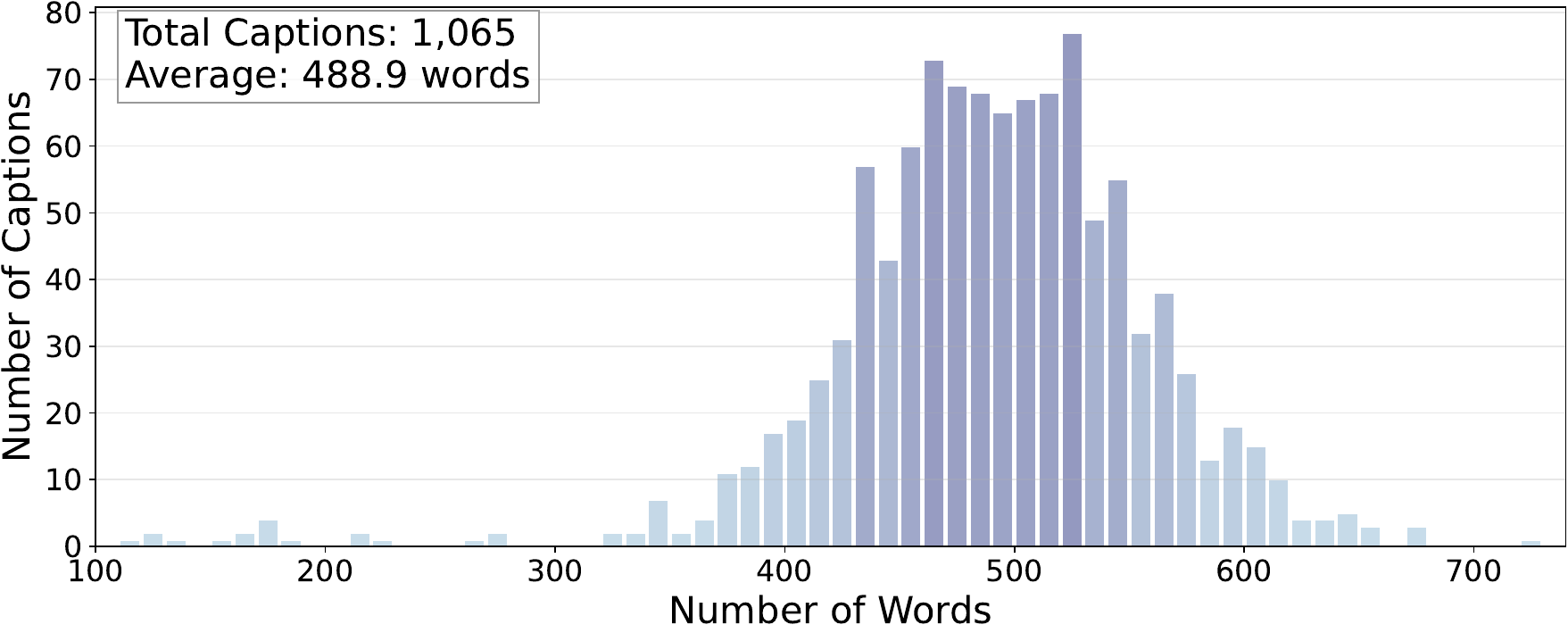}
            \caption{Video detailed captions length distribution.}
            \vspace{8pt} 
            \label{caption_length}
        \end{subfigure}
        \begin{subfigure}[b]{\textwidth}
            \includegraphics[width=\textwidth]{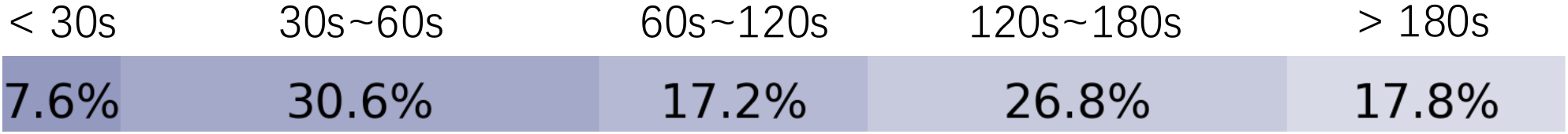}
            \caption{Video length duration.}
            \vspace{8pt}
            \label{video_length}
        \end{subfigure}
        \begin{subfigure}[b]{\textwidth}
            \includegraphics[width=\textwidth]{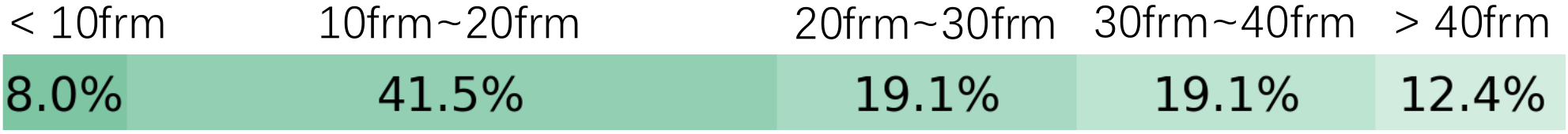}
            \caption{Keyframes number distribution.}
            \label{keyframe}
        \end{subfigure}
    \vspace{-15pt}
    \caption{Visualization of datasets statistics.}
    \end{minipage}
    \hspace{25pt}
    \begin{minipage}{0.5\textwidth} 
        \centering
        \includegraphics[width=\textwidth]{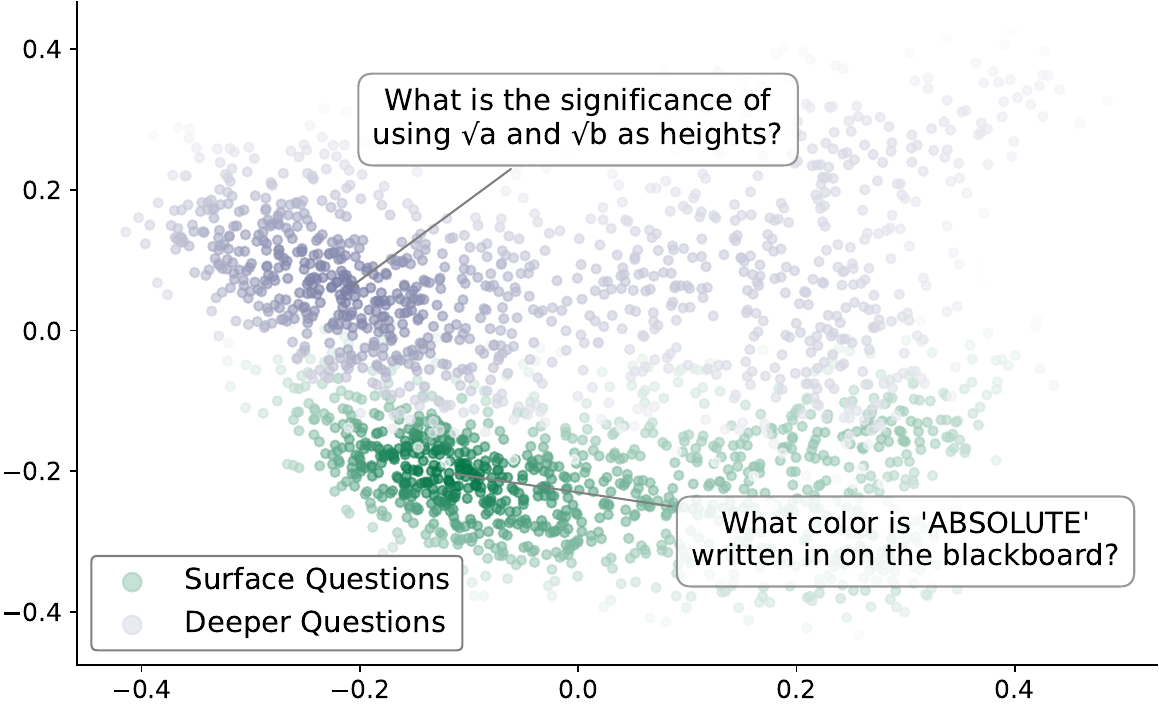}
        \caption{The text embedding space distribution of surface perception questions in green and deeper reasoning questions in purple.}
        \label{fig:embedding}
    \end{minipage}

    \label{fig:example}
\end{figure}

\vspace{-8pt}

\paragraph{Note generation.} The detailed captions in~\benchmark~describe main elements and backgrounds in the video, emphasizing formula recognition and changes in animated demonstrations. We believe that models must first accurately perceive surface visual features to effectively utilize them for reasoning tasks like question answering.

To generate detailed and accurate captions, we employ a multi-stage construction pipeline that structures the video caption as a framework, enriches frame-level details with image captions, and refines textual content with transcribed subtitles. Specifically, Aria~\cite{li2024aria} captures the temporal motion globally, \texttt{GPT-4o} generates detailed keyframe captions. Consequently, \texttt{Claude-3.5-sonnet} integrates the captions, enriching the structural framework provided by the video captions with image captions.
However, videos in~\benchmark~feature multi-disciplinary theorem demonstrations and problem-solving explanations, requiring recognition of numerous formulas and extensive text. Despite capturing fine-grained features, the combination of image captions and video captions alone cannot fully correct OCR errors. Therefore, we design an automatic refinement strategy to efficiently obtain accurate detailed captions. 
Transcribed subtitles are crucial for multimodal video understanding, providing key textual information from speech and reducing ambiguity in visual content. However, they may describe elements absent from visual frames. To ensure accuracy, we use \texttt{Claude-3.5-sonnet} to identify formulas and numbers in the initial captions, cross-checking and correcting them against the transcribed subtitles obtaining from YouTube API, while preventing the addition of new elements. 
Finally, we manually review captions to correct hallucinations and supplement omitted visual elements. The refined, detail structured captions serve as ground truth for evaluation. The multi-stage approach enables~\benchmark~to capture rich video details while minimizing hallucinations.

Following AuroraCap~\cite{chai2024auroracap}, we utilize VDCscore as the evaluation metric, which adopts a divide-and-conquer approach to transform long caption evaluation into multiple short question-answer (QA) pairs. For each ground-truth caption, we pre-generate 15 concise, open-ended QA pairs using \texttt{Claude-3.5-sonnet}.  These questions and answers are directly relevant to the video content, covering all explicit visual features in the frames.

\vspace{-8pt}

\paragraph{Quiz design.} Unlike detailed captioning, the visual question-answering task ('quiz') in~\benchmark~aims to challenge the model's ability to reason beyond surface features. Transcribed subtitles often explain the entire reasoning process, linking video elements and compensating for the lack of deeper reasoning in detailed captions. Our goal is to assess the model’s ability to reason underlying relationships by recognizing text and animations in frames without relying on subtitle guidance.
Therefore, we adopt a highly efficient automatic question-answer generation strategy, particularly for reasoning tasks. 
We combine pre-generated detailed captions with transcribed subtitles and use \texttt{Claude-3.5-sonnet} to generate high-quality, in-depth open-ended QA pairs. Since no reliable metric exists for long-form open-ended QA evaluation, we constrain \texttt{Claude-3.5-sonnet} to generate answers no longer than 15 words. Limiting the answer length ensures a fair and reliable evaluation while also assessing the model’s ability to distill key points succinctly. The average question length is as long as 10.09 words, and the average length of an answer is 13.36 words.

\subsection{Datasets Statistics}
\label{dataset_statistics}

\benchmark~comprises 1,065 multi-discipline lecture videos spanning mathematics (90.3\%), physics (3.6\%), and chemistry (6.1\%). Each topic requires a nuanced understanding of video context, foundational disciplinary knowledge, practical reasoning abilities, and logical deduction skills. The video durations range from 10 to 240 seconds, with an average duration of 109 seconds. As shown in Figure~\ref{video_length}, over 74\% videos are between 30 and 180 seconds, while 17.8\% extend beyond 180 seconds. Only 7.6\% of videos are shorter than 30 seconds. The distribution of extracted keyframes, visualized in Figure~\ref{keyframe}, reveals an average of 26 keyframes per video.

The benchmark includes 1,065 detailed captions and 15,746 reasoning question-answer pairs. For each detailed caption, we extract 15 surface question-answer pairs, yielding a total of 15,750 question-answer pairs. As indicated in Table~\ref{tab:benchmark}, our detailed captions have a competitive average length of 489 words compared to existing benchmarks, highlighting the comprehensiveness of the captions in~\benchmark. Figure~\ref{caption_length} presents the distribution of detailed captions in~\benchmark, with most being about 500 words.
To differentiate between the question types in video captioning and the “quiz,” we visualize the text embedding spaces of both question-answer pairs using the jina-embeddings-v3~\cite{sturua2024jina}. As shown in Figure~\ref{fig:embedding}, the embedding spaces for the two question types are clearly distinct, facilitating a comprehensive evaluation of models’ reasoning and perception. Additionally, we compute the Jensen-Shannon distance between the two embedding spaces, which is 0.668, further validating their distinctiveness.


\section{Evaluation of~\benchmark}

\subsection{Models and Evaluation Strategies}

\paragraph{Participating LMMs.} We evaluate a total of 87 models across three categories:
\begin{itemize}
\item \textbf{3 Proprietary Models}, including \texttt{Gemini-1.5-Flash}, \texttt{GPT-4o}, and \texttt{Claude-3.5-sonnet},
\item \textbf{78 Open-Source LMMs}, encompassing state-of-the-art video-specific LMMs and image-based LMMs capable of processing multiple images, with model sizes \textbf{ranging from 256M to 40B},
\item \textbf{6 Vision-Blind Baselines}, following~\cite{chen2024we, tong2024cambrian}.
\end{itemize}

\paragraph{Evaluation strategies.} To ensure a fair comparison, we maintain consistency by using the same 32 uniformly sampled frames across all models. Considering the limitations on input frame numbers for various image-based LMMs~\cite{lu2024deepseek, liu2024improved, li2024llavanext, chen2023sharegpt4v, dai2024instructblip, zhang2023internlm}, we reduce the input frames to 4 per video for these models. Notably, LLaVA-NeXT-Vicuna~\cite{li2024llavanext} series and XComposer~\cite{zhang2023internlm} fail to generate valid outputs on Video-MMLU with only 4 frames, so we further reduce the sampled frames to 2. For the visual QA track, we require models to provide concise responses, while for the video captioning track, we encourage generating the most detailed descriptions possible.
While proprietary models~\cite{openai2024gpt4o, google2024gemini, claude35sonnet2024} are widely used for evaluation, assessing large-scale benchmarks like~\benchmark~via API-based models is costly and not universally accessible. Additionally, the results are highly dependent on API versions. Therefore, we provide a free and reliable alternative by using Qwen2.5-72B~\cite{yang2024qwen2} as the LLM evaluation assistant with a temperature of 0. Given its strong reasoning capabilities, it ensures fair and accurate judgment, particularly for reasoning-based QA tasks. The evaluation is conducted using LMMs-Eval\cite{zhang2024lmms} and VLMEvalKit~\cite{duan2024vlmevalkit}.

\subsection{Leaderboard}
\begin{table*}[t]
\centering
\renewcommand{\arraystretch}{1.08}
\small
\caption{Results on~\benchmark~including vision-blind baselines, proprietary models, and open-source LMMs ($<$5B), across overall performance, detailed captioning (Notebook), and reasoning QA (Quiz) in different disciplines. Darker shades indicate better performance.}
\resizebox{\textwidth}{!}{%
\begin{tabular}{l cc| c |cccc| cccc }
\toprule
\multirow{2}{*}{\textbf{Models}} & \multirow{2}{*}{\textbf{LLM}} & \multirow{2}{*}{\textbf{Size}} & \multirow{2}{*}{\textbf{Overall}} & \multicolumn{4}{c|}{\textbf{Notebook}} & \multicolumn{4}{c}{\textbf{Quiz}} \\
& & & & \textbf{Avg.} & \textbf{Math} & \textbf{Physics} & \textbf{Chemistry} & \textbf{Avg.} & \textbf{Math} & \textbf{Physics} & \textbf{Chemistry}\\
\midrule
\textit{Vision-Blind Baselines} \\
\multirow{5}{*}{Qwen2.5~\cite{qwen2.5}} & -  & 0.5B & 4.29 & 2.25 &  2.31 & 4.44 & 0.01 & 6.33 & 4.92 & 8.88 & 5.19  \\
 & -  & 1.5B & 16.49 & 7.66 & 5.79 & 8.88 & 8.33 & 25.33 & 20.39 & 26.92 & 28.67  \\
 & -  & 3B & 21.64 & 11.31 & 11.88 & 14.24 & 7.81 & 31.97 & 27.58 & 32.12 & 36.23  \\
 & -  & 7B & 22.34 & 10.02 & 6.08 & 12.34 & 11.66 & 34.66 & 33.17 & 34.65 & 33.16  \\
 & -  & 32B & 24.76 & 13.65 & 9.85 & 17.45 & 13.66 & 35.87 & 34.14 & 36.15 & 37.33  \\
 & -  & 72B & 24.99 & 9.44 & 8.88 & 12.77 & 6.69 & 40.54 & 37.55 & 39.76 & 43.31  \\
\midrule  
\textit{Proprietary Models} \\
Gemini-1.5-Flash & -  & - & \rgbrank{43.63}{88} & \rgbrank{39.46}{86} & \rgbrank{27.69}{82} & \rgbrank{53.36}{92} & \rgbrank{37.33}{85} & \rgbrank{47.77}{92} & \rgbrank{44.36}{90} & \rgbrank{67.51}{94} & \rgbrank{31.43}{85}  \\
GPT-4o& -  & - & \rgbrank{49.41}{92} & \rgbrank{53.89}{94} & \rgbrank{55.23}{95} & \rgbrank{56.12}{94} & \rgbrank{50.33}{92} & \rgbrank{44.93}{90} & \rgbrank{33.08}{85} & \rgbrank{75.91}{96} & \rgbrank{25.79}{80}  \\
Claude-3.5-sonnet & -  & - & \rgbrank{69.34}{98} & \rgbrank{67.43}{98} & \rgbrank{63.74}{98} & \rgbrank{65.91}{98} & \rgbrank{72.66}{98} & \rgbrank{71.24}{98} & \rgbrank{68.29}{98} & \rgbrank{77.64}{98} & \rgbrank{67.80}{98}  \\
\midrule
\textit{Open-Source LMMs (\textasciitilde 5B)} \\
SmolVLM-256M~\cite{smolvlm2024} & SmolLM2~\cite{allal2025smollm2} & 135M & \rgbrank{8.95}{15} & \rgbrank{15.41}{35} & \rgbrank{11.87}{25} & \rgbrank{16.14}{35} & \rgbrank{18.24}{40} & \rgbrank{2.50}{8} & \rgbrank{1.62}{5} & \rgbrank{2.50}{8} & \rgbrank{3.40}{10} \\
VILA1.5-3B~\cite{lin2023vila} & Qwen2~\cite{wang2024qwen2} & 1.5B & \rgbrank{9.61}{18} & \rgbrank{18.71}{42} & \rgbrank{19.65}{48} & \rgbrank{15.83}{34} & \rgbrank{20.66}{45} & \rgbrank{0.51}{2} & \rgbrank{1.36}{4} & \rgbrank{0.13}{1} & \rgbrank{0.05}{1} \\
SmolVLM-500M~\cite{smolvlm2024} & SmolLM2~\cite{allal2025smollm2} & 360M & \rgbrank{11.05}{22} & \rgbrank{17.24}{38} & \rgbrank{11.29}{24} & \rgbrank{21.75}{48} & \rgbrank{18.68}{42} & \rgbrank{4.86}{12} & \rgbrank{3.65}{10} & \rgbrank{7.14}{18} & \rgbrank{3.81}{12} \\
SmolVLM~\cite{smolvlm2024} & SmolLM2~\cite{allal2025smollm2} & 1.7B & \rgbrank{14.14}{30} & \rgbrank{17.25}{38} & \rgbrank{14.91}{32} & \rgbrank{20.00}{45} & \rgbrank{16.86}{38} & \rgbrank{11.03}{28} & \rgbrank{6.09}{15} & \rgbrank{15.71}{38} & \rgbrank{11.29}{32} \\
DeepSeek-VL-1.3B~\cite{lu2024deepseek} & DeepSeek-LLM~\cite{bi2024deepseek} & 1.3B & \rgbrank{15.28}{34} & \rgbrank{20.59}{45} & \rgbrank{18.30}{42} & \rgbrank{20.51}{45} & \rgbrank{22.98}{50} & \rgbrank{9.98}{25} & \rgbrank{10.35}{28} & \rgbrank{8.57}{20} & \rgbrank{11.02}{30} \\
InternVL2-2B~\cite{chen2024far} & InternLM2~\cite{cai2024internlm2} & 1.8B & \rgbrank{15.60}{35} & \rgbrank{24.61}{55} & \rgbrank{22.44}{52} & \rgbrank{26.31}{58} & \rgbrank{25.09}{55} & \rgbrank{6.59}{15} & \rgbrank{5.68}{14} & \rgbrank{7.85}{19} & \rgbrank{6.25}{18} \\
Mini-InternVL-Chat-2B-V1.5~\cite{chen2024far} & InternLM2~\cite{cai2024internlm2} & 1.8B & \rgbrank{16.12}{36} & \rgbrank{21.19}{48} & \rgbrank{22.32}{51} & \rgbrank{20.00}{45} & \rgbrank{21.25}{48} & \rgbrank{11.05}{29} & \rgbrank{10.15}{26} & \rgbrank{10.35}{25} & \rgbrank{12.65}{34} \\
XinYuan-VL-2B~\cite{xinyuanvl2b} & Qwen2~\cite{wang2024qwen2} & 1.5B & \rgbrank{17.58}{40} & \rgbrank{29.65}{65} & \rgbrank{25.08}{58} & \rgbrank{32.63}{68} & \rgbrank{31.26}{68} & \rgbrank{5.52}{14} & \rgbrank{4.26}{12} & \rgbrank{6.07}{15} & \rgbrank{6.25}{18} \\
InternVL2-1B~\cite{chen2024far} & Qwen2~\cite{wang2024qwen2} & 0.5B & \rgbrank{18.59}{42} & \rgbrank{26.59}{58} & \rgbrank{22.77}{53} & \rgbrank{26.66}{59} & \rgbrank{30.34}{65} & \rgbrank{10.59}{27} & \rgbrank{10.35}{27} & \rgbrank{10.00}{24} & \rgbrank{11.43}{31} \\
Qwen2-VL-2B~\cite{wang2024qwen2} & Qwen2~\cite{wang2024qwen2} & 1.5B & \rgbrank{19.33}{44} & \rgbrank{30.19}{68} & \rgbrank{28.67}{62} & \rgbrank{32.98}{69} & \rgbrank{28.92}{62} & \rgbrank{8.47}{22} & \rgbrank{7.10}{18} & \rgbrank{10.71}{26} & \rgbrank{7.62}{20} \\
LLaVA-OneVision-OV~\cite{li2024llava} & Qwen2~\cite{wang2024qwen2} & 0.5B & \rgbrank{19.60}{45} & \rgbrank{23.77}{52} & \rgbrank{22.42}{52} & \rgbrank{20.89}{46} & \rgbrank{28.01}{60} & \rgbrank{15.43}{38} & \rgbrank{14.82}{35} & \rgbrank{13.92}{35} & \rgbrank{17.55}{45} \\
XComposer2-1.8B~\cite{dong2024internlm} & InternLM2~\cite{cai2024internlm2} & 1.8B & \rgbrank{19.77}{46} & \rgbrank{20.79}{46} & \rgbrank{14.21}{30} & \rgbrank{23.85}{52} & \rgbrank{24.32}{54} & \rgbrank{18.76}{45} & \rgbrank{13.60}{32} & \rgbrank{19.28}{48} & \rgbrank{23.40}{65} \\
InternVL2-4B~\cite{chen2024far} & Phi3-mini~\cite{abdin2024phi} & 3.8B & \rgbrank{20.44}{48} & \rgbrank{27.44}{60} & \rgbrank{26.28}{60} & \rgbrank{30.87}{65} & \rgbrank{25.19}{56} & \rgbrank{13.45}{34} & \rgbrank{11.16}{30} & \rgbrank{17.50}{42} & \rgbrank{11.70}{32} \\
Qwen2.5-VL-3B~\cite{bai2025qwen2} & Qwen2.5~\cite{qwen2.5} & 3B & \rgbrank{22.40}{50} & \rgbrank{31.06}{68} & \rgbrank{31.20}{68} & \rgbrank{32.63}{70} & \rgbrank{29.36}{65} & \rgbrank{13.74}{35} & \rgbrank{10.05}{27} & \rgbrank{17.85}{45} & \rgbrank{13.33}{35} \\
Aquila-VL-2B~\cite{gu2024infinity} & Qwen2.5~\cite{qwen2.5} & 1.5B & \rgbrank{23.94}{52} & \rgbrank{13.78}{30} & \rgbrank{13.14}{28} & \rgbrank{15.08}{32} & \rgbrank{13.14}{28} & \rgbrank{34.10}{75} & \rgbrank{30.45}{78} & \rgbrank{36.07}{78} & \rgbrank{35.78}{88} \\
Apollo-1.5B~\cite{zohar2024apollo} & Qwen2.5~\cite{qwen2.5} & 1.5B & \rgbrank{25.89}{55} & \rgbrank{26.43}{58} & \rgbrank{26.32}{60} & \rgbrank{21.66}{48} & \rgbrank{31.33}{68} & \rgbrank{25.35}{60} & \rgbrank{26.02}{68} & \rgbrank{20.01}{50} & \rgbrank{30.03}{82} \\
Apollo-3B~\cite{zohar2024apollo} & Qwen2.5~\cite{qwen2.5} & 3B & \rgbrank{27.27}{58} & \rgbrank{33.26}{72} & \rgbrank{32.30}{70} & \rgbrank{30.83}{65} & \rgbrank{36.66}{75} & \rgbrank{21.28}{52} & \rgbrank{17.12}{42} & \rgbrank{26.66}{62} & \rgbrank{20.07}{55} \\
InternVL2.5-1B~\cite{chen2024expanding} & Qwen2.5~\cite{qwen2.5} & 0.5B & \rgbrank{27.57}{60} & \rgbrank{31.71}{70} & \rgbrank{26.97}{62} & \rgbrank{34.38}{72} & \rgbrank{33.79}{72} & \rgbrank{23.43}{55} & \rgbrank{22.84}{55} & \rgbrank{23.92}{58} & \rgbrank{23.53}{68} \\
InternVL2.5-2B~\cite{chen2024expanding} & InternLM2.5~\cite{cai2024internlm2} & 1.8B & \rgbrank{28.62}{62} & \rgbrank{33.26}{72} & \rgbrank{27.94}{64} & \rgbrank{34.02}{71} & \rgbrank{37.83}{78} & \rgbrank{23.99}{58} & \rgbrank{22.43}{54} & \rgbrank{23.57}{57} & \rgbrank{25.98}{72} \\
Phi-3-Vision~\cite{abdin2024phi} & Phi3-mini~\cite{abdin2024phi} & 3.8B & \rgbrank{28.69}{62} & \rgbrank{21.85}{50} & \rgbrank{21.88}{50} & \rgbrank{23.85}{52} & \rgbrank{19.84}{44} & \rgbrank{35.54}{78} & \rgbrank{25.98}{65} & \rgbrank{41.07}{82} & \rgbrank{39.59}{90} \\
SAIL-VL-2B~\cite{dong2025scalable} & Qwen2.5~\cite{qwen2.5} & 1.5B & \rgbrank{28.86}{63} & \rgbrank{25.65}{56} & \rgbrank{23.27}{54} & \rgbrank{25.96}{57} & \rgbrank{27.74}{60} & \rgbrank{32.08}{72} & \rgbrank{27.71}{70} & \rgbrank{33.57}{75} & \rgbrank{34.96}{86} \\
Phi-3.5-Vision~\cite{abdin2024phi} & Phi3.5-mini~\cite{abdin2024phi} & 3.8B & \rgbrank{34.39}{72} & \rgbrank{29.55}{65} & \rgbrank{23.20}{54} & \rgbrank{32.38}{68} & \rgbrank{33.09}{70} & \rgbrank{39.23}{85} & \rgbrank{35.32}{88} & \rgbrank{40.35}{80} & \rgbrank{42.04}{92} \\
Mini-InternVL-Chat-4B-V1.5~\cite{chen2024far} & Phi3-mini~\cite{abdin2024phi} & 3.8B & \rgbrank{39.98}{82} & \rgbrank{25.76}{56} & \rgbrank{23.71}{55} & \rgbrank{30.17}{64} & \rgbrank{23.42}{52} & \rgbrank{54.20}{94} & \rgbrank{45.27}{90} & \rgbrank{61.42}{94} & \rgbrank{55.91}{94} \\
InternVL2.5-4B~\cite{chen2024expanding} & Qwen2.5~\cite{qwen2.5} & 3B & \rgbrank{40.74}{85} & \rgbrank{36.75}{75} & \rgbrank{31.35}{68} & \rgbrank{36.30}{75} & \rgbrank{42.61}{85} & \rgbrank{44.74}{90} & \rgbrank{41.82}{89} & \rgbrank{46.42}{88} & \rgbrank{45.98}{93} \\
Aria~\cite{li2024aria} & 66 Experts MoEs & 3.9B & \rgbrank{42.87}{87} & \rgbrank{45.09}{88} & \rgbrank{41.45}{85} & \rgbrank{45.83}{88} & \rgbrank{48.0}{90} & \rgbrank{40.65}{86} & \rgbrank{39.17}{87} & \rgbrank{42.66}{85} & \rgbrank{40.12}{91} \\
\bottomrule
\end{tabular}%
}
\label{tab:evu_tiny}
\end{table*}

\begin{table*}[t]
\centering
\renewcommand{\arraystretch}{1.05}
\small
\caption{Results on~\benchmark~including proprietary models, and open-source LMMs ($<$8B), across overall performance, detailed captioning (Notebook), and reasoning QA (Quiz) in different disciplines. Darker shades indicate better performance.}
\resizebox{\textwidth}{!}{%
\begin{tabular}{l c c | c |cccc| cccc }
\toprule
\multirow{2}{*}{\textbf{Models}} & \multirow{2}{*}{\textbf{LLM}} & \multirow{2}{*}{\textbf{Size}} & \multirow{2}{*}{\textbf{Overall}} & \multicolumn{4}{c|}{\textbf{Notebook}} & \multicolumn{4}{c}{\textbf{Quiz}} \\
& & & & \textbf{Avg.} & \textbf{Math} & \textbf{Physics} & \textbf{Chemistry} & \textbf{Avg.} & \textbf{Math} & \textbf{Physics} & \textbf{Chemistry}\\
\midrule
\textit{Proprietary Models} \\
Gemini-1.5-Flash & -  & - & \rgbrank{43.63}{88} & \rgbrank{39.46}{86} & \rgbrank{27.69}{82} & \rgbrank{53.36}{92} & \rgbrank{37.33}{85} & \rgbrank{47.77}{92} & \rgbrank{44.36}{90} & \rgbrank{67.51}{94} & \rgbrank{31.43}{85}  \\
GPT-4o& -  & - & \rgbrank{49.41}{92} & \rgbrank{53.89}{94} & \rgbrank{55.23}{95} & \rgbrank{56.12}{94} & \rgbrank{50.33}{92} & \rgbrank{44.93}{90} & \rgbrank{33.08}{85} & \rgbrank{75.91}{96} & \rgbrank{25.79}{80}  \\
Claude-3.5-sonnet & -  & - & \rgbrank{69.34}{98} & \rgbrank{67.43}{98} & \rgbrank{63.74}{98} & \rgbrank{65.91}{98} & \rgbrank{72.66}{98} & \rgbrank{71.24}{98} & \rgbrank{68.29}{98} & \rgbrank{77.64}{98} & \rgbrank{67.80}{98}  \\
\midrule
\textit{Open-Source LMMs (\textasciitilde 8B)} \\
XComposer~\cite{zhang2023internlm} & InternLM~\cite{dong2024internlm} & 7B & \rgbrank{10.91}{20} & \rgbrank{20.52}{45} & \rgbrank{12.96}{28} & \rgbrank{23.50}{51} & \rgbrank{25.10}{55} & \rgbrank{1.29}{4} & \rgbrank{1.42}{4} & \rgbrank{0.71}{2} & \rgbrank{1.76}{5} \\
InstructBLIP-7B~\cite{dai2024instructblip} & Vicuna~\cite{chiang2023vicuna} & 7B & \rgbrank{11.06}{22} & \rgbrank{19.26}{42} & \rgbrank{14.54}{31} & \rgbrank{21.75}{48} & \rgbrank{21.49}{48} & \rgbrank{2.86}{8} & \rgbrank{1.11}{3} & \rgbrank{4.64}{12} & \rgbrank{2.85}{8} \\
Mantis-8B-Fuyu~\cite{Jiang2024MANTISIM} & Fuyu~\cite{fuyu8b2024} & 8B & \rgbrank{12.31}{25} & \rgbrank{16.50}{36} & \rgbrank{12.41}{26} & \rgbrank{17.19}{38} & \rgbrank{19.91}{44} & \rgbrank{8.12}{20} & \rgbrank{6.09}{15} & \rgbrank{8.21}{20} & \rgbrank{10.06}{26} \\
Cambrian-8B~\cite{tong2024cambrian} & LLaMA3~\cite{llama3modelcard} & 8B & \rgbrank{12.68}{26} & \rgbrank{20.17}{44} & \rgbrank{20.38}{49} & \rgbrank{21.75}{48} & \rgbrank{18.38}{40} & \rgbrank{5.19}{13} & \rgbrank{3.35}{8} & \rgbrank{6.78}{16} & \rgbrank{5.44}{15} \\
Mantis-8B-siglip-llama3~\cite{Jiang2024MANTISIM} & LLaMA3~\cite{llama3modelcard} & 8B & \rgbrank{13.74}{28} & \rgbrank{23.95}{52} & \rgbrank{14.12}{30} & \rgbrank{28.42}{62} & \rgbrank{29.33}{64} & \rgbrank{3.54}{10} & \rgbrank{2.33}{6} & \rgbrank{1.78}{5} & \rgbrank{6.53}{18} \\
Mantis-8B-Idefics2~\cite{Jiang2024MANTISIM} & Mistral~\cite{ministral8b2024} & 7B & \rgbrank{14.19}{30} & \rgbrank{21.41}{48} & \rgbrank{16.81}{38} & \rgbrank{23.50}{51} & \rgbrank{23.94}{53} & \rgbrank{6.98}{18} & \rgbrank{6.70}{16} & \rgbrank{6.78}{16} & \rgbrank{7.48}{19} \\
LLaVA-1.5~\cite{liu2024improved} & Vicuna~\cite{chiang2023vicuna} & 7B & \rgbrank{15.71}{35} & \rgbrank{22.31}{50} & \rgbrank{15.30}{34} & \rgbrank{22.81}{50} & \rgbrank{28.84}{62} & \rgbrank{9.11}{24} & \rgbrank{7.81}{20} & \rgbrank{8.92}{22} & \rgbrank{10.61}{28} \\
Video-LlaVA-7B~\cite{lin2023video} & Vicuna~\cite{chiang2023vicuna} & 7B & \rgbrank{15.89}{35} & \rgbrank{15.32}{34} & \rgbrank{11.15}{24} & \rgbrank{16.14}{35} & \rgbrank{18.67}{41} & \rgbrank{16.47}{40} & \rgbrank{13.29}{31} & \rgbrank{17.5}{44} & \rgbrank{18.63}{48} \\
VideoChat2-HD~\cite{li2024mvbench} & Mistral~\cite{ministral8b2024} & 7B & \rgbrank{16.74}{38} & \rgbrank{18.07}{40} & \rgbrank{15.63}{35} & \rgbrank{19.65}{44} & \rgbrank{18.94}{42} & \rgbrank{15.40}{38} & \rgbrank{12.79}{30} & \rgbrank{13.15}{32} & \rgbrank{20.26}{55} \\
Mantis-8B-clip-llama3~\cite{Jiang2024MANTISIM}& LLaMA3~\cite{llama3modelcard} & 8B & \rgbrank{18.78}{42} & \rgbrank{21.62}{49} & \rgbrank{14.37}{31} & \rgbrank{24.56}{54} & \rgbrank{25.95}{58} & \rgbrank{15.95}{39} & \rgbrank{13.29}{31} & \rgbrank{13.21}{32} & \rgbrank{21.36}{58} \\
Video-ChatGPT~\cite{maaz2023video} & Vicuna~\cite{chiang2023vicuna} & 7B & \rgbrank{19.37}{44} & \rgbrank{16.30}{36} & \rgbrank{10.88}{23} & \rgbrank{15.78}{34} & \rgbrank{22.25}{50} & \rgbrank{22.45}{55} & \rgbrank{19.08}{44} & \rgbrank{20.00}{49} & \rgbrank{28.29}{78} \\
LLaVA-NeXT~\cite{li2024llavanext} & Vicuna~\cite{chiang2023vicuna} & 7B & \rgbrank{21.46}{48} & \rgbrank{18.06}{40} & \rgbrank{9.79}{22} & \rgbrank{21.40}{48} & \rgbrank{22.99}{52} & \rgbrank{24.87}{60} & \rgbrank{21.31}{52} & \rgbrank{22.85}{55} & \rgbrank{30.47}{82} \\
Qwen-VL~\cite{bai2023versatile} & Qwen~\cite{bai2023qwen} & 7B & \rgbrank{21.98}{50} & \rgbrank{24.35}{54} & \rgbrank{19.56}{47} & \rgbrank{25.61}{56} & \rgbrank{27.88}{60} & \rgbrank{19.62}{46} & \rgbrank{16.95}{40} & \rgbrank{16.07}{40} & \rgbrank{25.85}{70} \\
mPLUG-Owl3~\cite{ye2024mplug} & Qwen2~\cite{wang2024qwen2} & 7B & \rgbrank{22.59}{51} & \rgbrank{22.55}{51} & \rgbrank{18.61}{45} & \rgbrank{24.91}{55} & \rgbrank{24.13}{53} & \rgbrank{22.64}{56} & \rgbrank{18.57}{43} & \rgbrank{25.00}{60} & \rgbrank{24.35}{69} \\
ShareGPT4V-7B~\cite{chen2023sharegpt4v} & Vicuna~\cite{chiang2023vicuna} & 7B & \rgbrank{22.81}{51} & \rgbrank{23.48}{52} & \rgbrank{18.99}{45} & \rgbrank{25.16}{55} & \rgbrank{26.30}{58} & \rgbrank{22.15}{54} & \rgbrank{17.83}{42} & \rgbrank{22.24}{54} & \rgbrank{26.40}{72} \\
LLaVA-NeXT~\cite{li2024llavanext} & LLaMA3~\cite{llama3modelcard} & 8B & \rgbrank{23.29}{52} & \rgbrank{16.53}{37} & \rgbrank{8.97}{20} & \rgbrank{20.00}{45} & \rgbrank{20.64}{46} & \rgbrank{30.05}{70} & \rgbrank{24.06}{58} & \rgbrank{32.50}{74} & \rgbrank{33.60}{85} \\
PLLaVA~\cite{xu2024pllava} & Vicuna~\cite{chiang2023vicuna} & 7B & \rgbrank{23.85}{53} & \rgbrank{16.08}{35} & \rgbrank{12.75}{27} & \rgbrank{18.18}{40} & \rgbrank{17.33}{38} & \rgbrank{31.63}{71} & \rgbrank{21.55}{50} & \rgbrank{41.44}{83} & \rgbrank{31.91}{83} \\
InternVL2-8B~\cite{chen2024far} & InternLM2.5~\cite{cai2024internlm2} & 7B & \rgbrank{24.06}{54} & \rgbrank{31.43}{69} & \rgbrank{26.12}{59} & \rgbrank{33.33}{70} & \rgbrank{34.85}{74} & \rgbrank{16.69}{41} & \rgbrank{13.19}{31} & \rgbrank{21.78}{52} & \rgbrank{15.10}{40} \\
DeepSeek-VL-7B~\cite{lu2024deepseek} & DeepSeek-LLM~\cite{bi2024deepseek} & 7B & \rgbrank{24.12}{54} & \rgbrank{26.20}{58} & \rgbrank{25.62}{58} & \rgbrank{25.65}{56} & \rgbrank{27.33}{60} & \rgbrank{22.04}{54} & \rgbrank{20.50}{48} & \rgbrank{18.57}{46} & \rgbrank{27.07}{75} \\
VILA1.5-8B~\cite{lin2023vila}& LLaMA3~\cite{llama3modelcard} & 8B & \rgbrank{24.20}{54} & \rgbrank{27.95}{62} & \rgbrank{25.38}{58} & \rgbrank{25.83}{57} & \rgbrank{32.66}{70} & \rgbrank{20.45}{48} & \rgbrank{14.38}{34} & \rgbrank{26.73}{64} & \rgbrank{20.24}{54} \\
XComposer2~\cite{dong2024internlm} & InternLM2~\cite{cai2024internlm2} & 7B & \rgbrank{25.62}{56} & \rgbrank{16.24}{36} & \rgbrank{12.68}{27} & \rgbrank{17.89}{39} & \rgbrank{18.16}{40} & \rgbrank{35.00}{76} & \rgbrank{26.90}{69} & \rgbrank{38.92}{80} & \rgbrank{39.18}{89} \\
LLaVA-NeXT~\cite{li2024llavanext} & Mistral~\cite{ministral8b2024} & 7B & \rgbrank{25.83}{57} & \rgbrank{20.31}{45} & \rgbrank{18.48}{44} & \rgbrank{21.05}{47} & \rgbrank{21.42}{48} & \rgbrank{31.45}{70} & \rgbrank{26.09}{66} & \rgbrank{33.57}{75} & \rgbrank{34.69}{86} \\
Qwen2-VL-7B~\cite{wang2024qwen2} & Qwen2~\cite{wang2024qwen2} & 7B & \rgbrank{28.83}{63} & \rgbrank{34.22}{74} & \rgbrank{27.58}{63} & \rgbrank{35.37}{74} & \rgbrank{39.72}{80} & \rgbrank{23.44}{57} & \rgbrank{19.59}{46} & \rgbrank{24.07}{59} & \rgbrank{26.66}{74} \\
LLaVA-NeXT-Video-7B~\cite{zhang2024videoinstructiontuningsynthetic} & Qwen2~\cite{wang2024qwen2} & 7B & \rgbrank{31.55}{67} & \rgbrank{35.75}{74} & \rgbrank{30.03}{65} & \rgbrank{36.69}{76} & \rgbrank{40.54}{82} & \rgbrank{27.35}{65} & \rgbrank{29.32}{75} & \rgbrank{24.07}{59} & \rgbrank{28.66}{79} \\
LLaVA-OneVision-OV~\cite{li2024llava} & Qwen2~\cite{wang2024qwen2} & 7B & \rgbrank{33.99}{70} & \rgbrank{34.55}{73} & \rgbrank{29.53}{64} & \rgbrank{35.66}{74} & \rgbrank{38.46}{79} & \rgbrank{33.44}{73} & \rgbrank{30.35}{78} & \rgbrank{35.71}{77} & \rgbrank{34.28}{85} \\
Apollo-7B~\cite{zohar2024apollo} & Qwen2.5~\cite{qwen2.5} & 7B & \rgbrank{36.78}{75} & \rgbrank{38.22}{78} & \rgbrank{33.50}{71} & \rgbrank{39.16}{80} & \rgbrank{42.00}{84} & \rgbrank{35.33}{77} & \rgbrank{29.45}{76} & \rgbrank{26.56}{63} & \rgbrank{49.98}{93} \\
Qwen2.5-VL-7B\cite{bai2025qwen2} & Qwen2.5~\cite{qwen2.5} & 7B & \rgbrank{37.47}{78} & \rgbrank{42.02}{85} & \rgbrank{39.43}{82} & \rgbrank{44.91}{88} & \rgbrank{41.73}{84} & \rgbrank{32.93}{74} & \rgbrank{24.36}{60} & \rgbrank{41.78}{84} & \rgbrank{32.65}{84} \\
Valley-Eagle~\cite{wu2025valley2} & Qwen2.5~\cite{qwen2.5} & 7B & \rgbrank{37.96}{78} & \rgbrank{39.22}{80} & \rgbrank{33.21}{72} & \rgbrank{41.40}{85} & \rgbrank{43.07}{85} & \rgbrank{36.71}{80} & \rgbrank{28.40}{72} & \rgbrank{37.14}{78} & \rgbrank{44.58}{92} \\
MiniCPM-V 2.6~\cite{yao2024minicpm} & Qwen2~\cite{wang2024qwen2} & 7B & \rgbrank{39.31}{80} & \rgbrank{43.57}{88} & \rgbrank{36.13}{75} & \rgbrank{47.01}{90} & \rgbrank{47.59}{90} & \rgbrank{35.06}{76} & \rgbrank{28.79}{74} & \rgbrank{37.66}{79} & \rgbrank{38.73}{89} \\
InternVL2.5-8B~\cite{chen2024expanding} & InternLM2.5~\cite{cai2024internlm2} & 7B & \rgbrank{39.51}{80} & \rgbrank{34.51}{73} & \rgbrank{30.48}{66} & \rgbrank{34.53}{72} & \rgbrank{38.52}{79} & \rgbrank{44.51}{91} & \rgbrank{39.91}{88} & \rgbrank{46.29}{87} & \rgbrank{47.33}{94} \\
MiniCPM-o 2.6~\cite{yao2024minicpm} & Qwen2.5~\cite{qwen2.5} & 7B & \rgbrank{44.89}{90} & \rgbrank{54.83}{95} & \rgbrank{47.86}{92} & \rgbrank{57.54}{95} & \rgbrank{59.10}{95} & \rgbrank{34.95}{76} & \rgbrank{39.28}{87} & \rgbrank{22.85}{55} & \rgbrank{42.72}{92} \\
\bottomrule
\end{tabular}%
}
\label{tab:evu_small}
\end{table*}

\definecolor{small}{RGB}{255, 255, 255} 
\definecolor{big}{RGB}{148, 153, 192}

\begin{table*}[t]
\centering
\renewcommand{\arraystretch}{1.08}
\small
\caption{Results on~\benchmark~including proprietary models, and open-source LMMs ($<$40B), across overall performance, detailed captioning (Notebook), and reasoning QA (Quiz) in different disciplines. Darker shades indicate better performance.}
\vspace{-3pt}
\resizebox{\textwidth}{!}{%
\begin{tabular}{l cc| c |cccc| cccc }
\toprule
\multirow{2}{*}{\textbf{Models}} & \multirow{2}{*}{\textbf{LLM}} & \multirow{2}{*}{\textbf{Size}} & \multirow{2}{*}{\textbf{Overall}} & \multicolumn{4}{c|}{\textbf{Notebook}} & \multicolumn{4}{c}{\textbf{Quiz}} \\
& & & & \textbf{Avg.} & \textbf{Math} & \textbf{Physics} & \textbf{Chemistry} & \textbf{Avg.} & \textbf{Math} & \textbf{Physics} & \textbf{Chemistry}\\
\midrule
\textit{Proprietary Models} \\
Gemini-1.5-Flash & -  & - & \rgbrank{43.63}{85} & \rgbrank{39.46}{80} & \rgbrank{27.69}{62} & \rgbrank{53.36}{92} & \rgbrank{37.33}{78} & \rgbrank{47.77}{95} & \rgbrank{44.36}{92} & \rgbrank{67.51}{95} & \rgbrank{31.43}{85}  \\
GPT-4o & -  & - & \rgbrank{49.41}{90} & \rgbrank{53.89}{92} & \rgbrank{55.23}{94} & \rgbrank{56.12}{94} & \rgbrank{50.33}{90} & \rgbrank{44.93}{92} & \rgbrank{33.08}{85} & \rgbrank{75.91}{98} & \rgbrank{25.79}{75}  \\
Claude-3.5-sonnet & -  & - & \rgbrank{69.34}{98} & \rgbrank{67.43}{96} & \rgbrank{63.74}{96} & \rgbrank{65.91}{96} & \rgbrank{72.66}{98} & \rgbrank{71.24}{96} & \rgbrank{68.29}{95} & \rgbrank{77.64}{98} & \rgbrank{67.80}{95}  \\
\midrule
\textit{Open-Source LMMs (\textasciitilde 20B)} \\
LLaVA-NeXT~\cite{li2024llavanext} & Vicuna~\cite{chiang2023vicuna} & 13B & \rgbrank{8.13}{15} & \rgbrank{16.27}{35} & \rgbrank{9.95}{22} & \rgbrank{22.55}{50} & \rgbrank{16.32}{35} & \rgbrank{0.0}{1} & \rgbrank{0.0}{1} & \rgbrank{0.0}{1} & \rgbrank{0.0}{1} \\
ShareGPT4V-13B~\cite{chen2023sharegpt4v} & Vicuna~\cite{chiang2023vicuna} & 13B & \rgbrank{11.57}{24} & \rgbrank{18.37}{40} & \rgbrank{17.13}{38} & \rgbrank{19.31}{42} & \rgbrank{18.69}{42} & \rgbrank{4.78}{12} & \rgbrank{4.06}{10} & \rgbrank{5.00}{12} & \rgbrank{5.30}{15} \\
Cambrian-13B~\cite{tong2024cambrian} & Vicuna~\cite{chiang2023vicuna} & 13B & \rgbrank{14.56}{32} & \rgbrank{21.77}{48} & \rgbrank{20.14}{45} & \rgbrank{24.21}{54} & \rgbrank{20.97}{46} & \rgbrank{7.36}{18} & \rgbrank{4.16}{11} & \rgbrank{10.71}{25} & \rgbrank{7.21}{20} \\
VILA1.5-13B~\cite{lin2023vila} & Vicuna~\cite{chiang2023vicuna} & 13B & \rgbrank{15.71}{35} & \rgbrank{24.95}{55} & \rgbrank{24.21}{54} & \rgbrank{22.66}{51} & \rgbrank{28.00}{62} & \rgbrank{6.48}{16} & \rgbrank{2.73}{8} & \rgbrank{6.68}{16} & \rgbrank{10.05}{28} \\
InstructBLIP-13B~\cite{dai2024instructblip} & Vicuna~\cite{chiang2023vicuna} & 13B & \rgbrank{15.89}{36} & \rgbrank{22.32}{50} & \rgbrank{14.96}{32} & \rgbrank{26.51}{58} & \rgbrank{25.49}{56} & \rgbrank{9.47}{24} & \rgbrank{5.17}{14} & \rgbrank{15.35}{38} & \rgbrank{7.89}{22} \\
InternVL-Chat-V1-1~\cite{chen2024far} & LLaMA2~\cite{touvron2023llama} & 13B & \rgbrank{21.53}{48} & \rgbrank{24.83}{54} & \rgbrank{22.30}{50} & \rgbrank{26.31}{58} & \rgbrank{25.88}{57} & \rgbrank{18.22}{45} & \rgbrank{13.29}{32} & \rgbrank{21.78}{52} & \rgbrank{19.59}{52} \\
LLaVA-1.5~\cite{liu2024improved} & Vicuna~\cite{chiang2023vicuna} & 13B & \rgbrank{21.58}{48} & \rgbrank{16.74}{36} & \rgbrank{12.75}{28} & \rgbrank{21.75}{48} & \rgbrank{15.72}{35} & \rgbrank{26.42}{65} & \rgbrank{23.14}{58} & \rgbrank{26.07}{62} & \rgbrank{30.06}{82} \\
OmChat-v2.0-13B~\cite{zhao2024omchat} & Qwen2~\cite{wang2024qwen2} & 7B & \rgbrank{21.91}{49} & \rgbrank{24.57}{54} & \rgbrank{21.26}{48} & \rgbrank{25.61}{56} & \rgbrank{26.85}{59} & \rgbrank{19.26}{48} & \rgbrank{15.32}{38} & \rgbrank{20.71}{50} & \rgbrank{21.76}{58} \\
PLLaVA-13B~\cite{xu2024pllava} & Vicuna~\cite{chiang2023vicuna} & 13B & \rgbrank{26.25}{58} & \rgbrank{21.08}{47} & \rgbrank{17.75}{40} & \rgbrank{21.42}{48} & \rgbrank{24.07}{53} & \rgbrank{31.43}{72} & \rgbrank{25.27}{62} & \rgbrank{28.21}{68} & \rgbrank{40.81}{90} \\
InternVL-Chat-V1-5~\cite{chen2024far} & InternLM~\cite{dong2024internlm} & 20B & \rgbrank{28.76}{64} & \rgbrank{26.00}{57} & \rgbrank{21.35}{48} & \rgbrank{26.31}{58} & \rgbrank{30.34}{68} & \rgbrank{31.53}{73} & \rgbrank{32.33}{82} & \rgbrank{29.62}{70} & \rgbrank{32.66}{84} \\
CogVLM2-LLaMA3-Chat-19B~\cite{hong2024cogvlm2} & LLaMA3~\cite{llama3modelcard} & 8B & \rgbrank{31.99}{70} & \rgbrank{24.08}{53} & \rgbrank{21.33}{48} & \rgbrank{24.91}{55} & \rgbrank{26.01}{57} & \rgbrank{39.90}{85} & \rgbrank{32.58}{83} & \rgbrank{41.42}{88} & \rgbrank{45.71}{92} \\
\midrule
\textit{Open-Source LMMs (\textasciitilde 40B)} \\
Cambrian-34B~\cite{tong2024cambrian} & Nous-Hermes-2-Yi~\cite{noushermes2yi2023} & 34B & \rgbrank{12.73}{28} & \rgbrank{19.90}{44} & \rgbrank{20.52}{46} & \rgbrank{22.10}{49} & \rgbrank{17.08}{38} & \rgbrank{5.56}{14} & \rgbrank{4.06}{10} & \rgbrank{6.78}{16} & \rgbrank{5.85}{16} \\
InternVL2-26B~\cite{chen2024far} & InternLM2~\cite{cai2024internlm2} & 20B & \rgbrank{21.33}{48} & \rgbrank{29.68}{65} & \rgbrank{26.19}{58} & \rgbrank{28.77}{63} & \rgbrank{34.01}{75} & \rgbrank{12.98}{32} & \rgbrank{13.84}{34} & \rgbrank{11.11}{26} & \rgbrank{14.00}{38} \\
InternVL-Chat-V1-2-Plus~\cite{chen2024far} & Nous-Hermes-2-Yi~\cite{noushermes2yi2023} & 34B & \rgbrank{24.25}{54} & \rgbrank{18.88}{42} & \rgbrank{21.23}{48} & \rgbrank{21.05}{47} & \rgbrank{14.36}{32} & \rgbrank{29.62}{70} & \rgbrank{22.13}{55} & \rgbrank{38.57}{85} & \rgbrank{28.16}{78} \\
InternVL2-40B~\cite{chen2024far} & Nous-Hermes-2-Yi~\cite{noushermes2yi2023} & 34B & \rgbrank{27.44}{60} & \rgbrank{32.74}{70} & \rgbrank{28.67}{64} & \rgbrank{33.58}{70} & \rgbrank{35.99}{75} & \rgbrank{22.15}{55} & \rgbrank{19.79}{48} & \rgbrank{25.71}{60} & \rgbrank{20.95}{56} \\
InternVL-Chat-V1-2~\cite{chen2024far} & Nous-Hermes-2-Yi~\cite{noushermes2yi2023} & 34B & \rgbrank{29.00}{65} & \rgbrank{21.42}{48} & \rgbrank{23.14}{52} & \rgbrank{26.66}{59} & \rgbrank{14.47}{32} & \rgbrank{36.58}{80} & \rgbrank{23.85}{58} & \rgbrank{50.00}{90} & \rgbrank{35.91}{88} \\
VILA1.5-40B~\cite{lin2023vila} & Nous-Hermes-2-Yi~\cite{noushermes2yi2023} & 34B & \rgbrank{30.72}{68} & \rgbrank{32.30}{70} & \rgbrank{29.91}{65} & \rgbrank{28.33}{62} & \rgbrank{38.66}{85} & \rgbrank{29.13}{68} & \rgbrank{31.51}{80} & \rgbrank{20.15}{48} & \rgbrank{35.72}{87} \\
PLLaVA-34B~\cite{xu2024pllava} & Nous-Hermes-2-Yi~\cite{noushermes2yi2023} & 34B & \rgbrank{30.91}{68} & \rgbrank{21.09}{47} & \rgbrank{20.53}{46} & \rgbrank{22.10}{49} & \rgbrank{20.65}{45} & \rgbrank{40.74}{88} & \rgbrank{30.62}{78} & \rgbrank{47.22}{89} & \rgbrank{44.38}{91} \\
LLaVA-NeXT~\cite{li2024llavanext} & Nous-Hermes-2-Yi~\cite{noushermes2yi2023} & 34B & \rgbrank{34.16}{72} & \rgbrank{25.07}{55} & \rgbrank{23.58}{52} & \rgbrank{25.72}{56} & \rgbrank{25.93}{58} & \rgbrank{43.25}{90} & \rgbrank{21.42}{52} & \rgbrank{58.33}{93} & \rgbrank{50.00}{94} \\
LLaVA-NeXT~\cite{li2024llavanext} & Qwen1.5~\cite{qwen} & 32B & \rgbrank{40.43}{85} & \rgbrank{26.98}{59} & \rgbrank{23.10}{52} & \rgbrank{22.96}{51} & \rgbrank{34.90}{76} & \rgbrank{53.88}{94} & \rgbrank{46.10}{93} & \rgbrank{55.55}{92} & \rgbrank{60.00}{95} \\
InternVL2.5-26B~\cite{chen2024expanding} & InternLM2.5~\cite{cai2024internlm2} & 20B & \rgbrank{44.39}{90} & \rgbrank{39.71}{85} & \rgbrank{32.90}{72} & \rgbrank{47.91}{92} & \rgbrank{38.33}{84} & \rgbrank{49.07}{93} & \rgbrank{47.21}{94} & \rgbrank{50.00}{90} & \rgbrank{50.00}{94} \\
InternVL2.5-38B~\cite{chen2024expanding} & Qwen2.5~\cite{qwen2.5} & 32B & \rgbrank{49.35}{95} & \rgbrank{37.01}{80} & \rgbrank{34.25}{75} & \rgbrank{35.13}{75} & \rgbrank{41.66}{90} & \rgbrank{61.68}{96} & \rgbrank{60.47}{96} & \rgbrank{59.25}{94} & \rgbrank{65.33}{98} \\
\bottomrule
\end{tabular}%
}
\label{tab:evu_big}
\end{table*}
\begin{figure}[!h]
    \centering
    \begin{minipage}{0.49\textwidth}
        \centering
        \includegraphics[width=\textwidth]{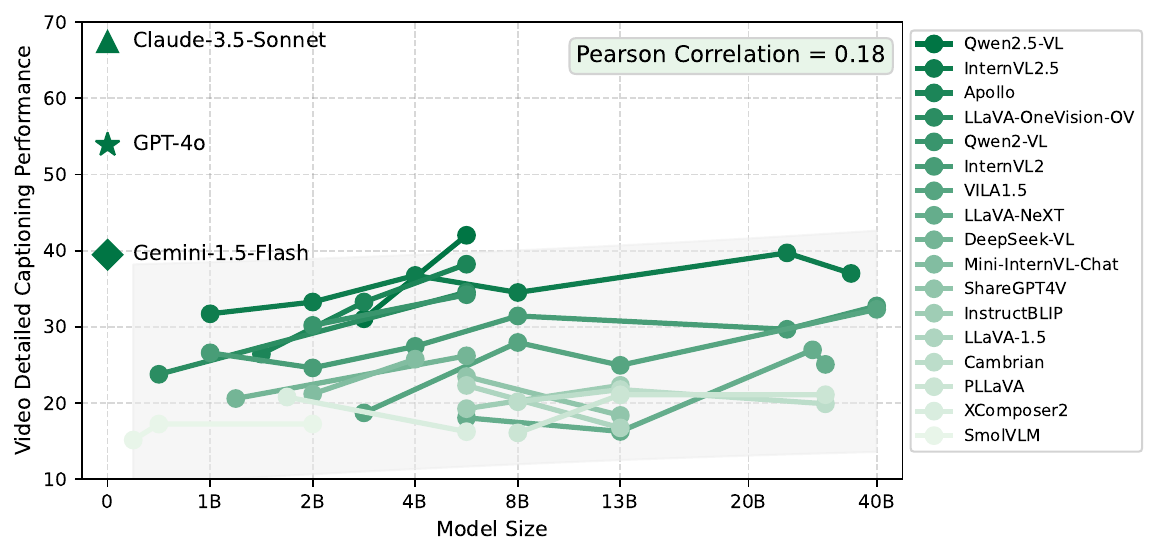}
        \caption{\textbf{Relationship between model size and video captioning performance.} The shaded region shows the confidence interval, with darker colors indicating better performance.}
        \label{caption_performance}
    \end{minipage}
    \hfill
    \begin{minipage}{0.49\textwidth}
        \centering
        \includegraphics[width=\textwidth]{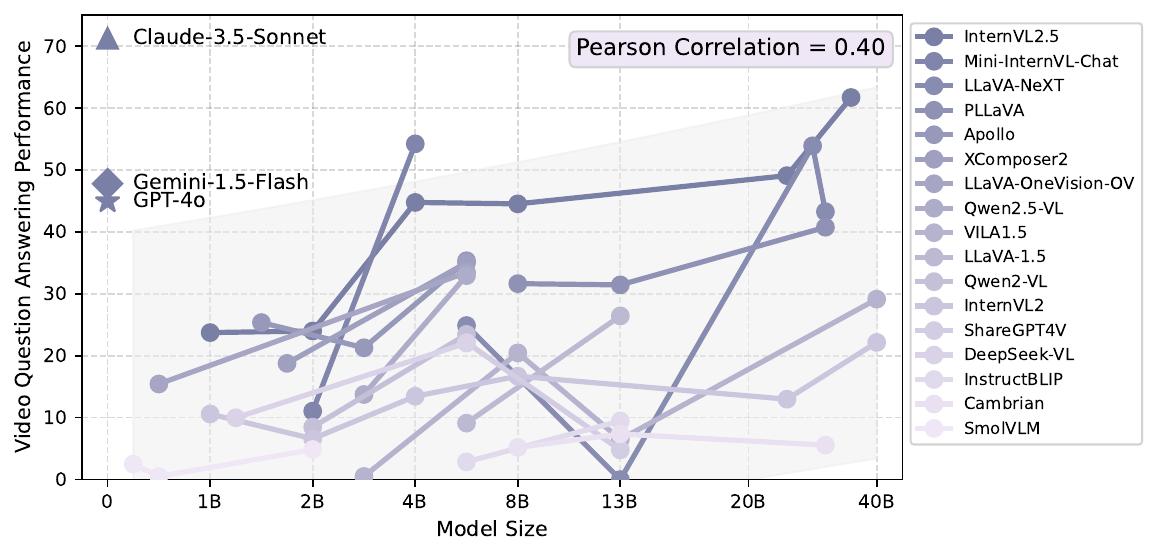}
        \caption{\textbf{Relationship between model size and video QA performance.} The shaded region shows the confidence interval, with darker colors indicating better performance.}
        \label{qa_performance}
    \end{minipage}
\end{figure}

\begin{figure}[h]
  \centering
    \begin{minipage}{0.32\textwidth}
        \centering
        \includegraphics[width=0.98\linewidth]{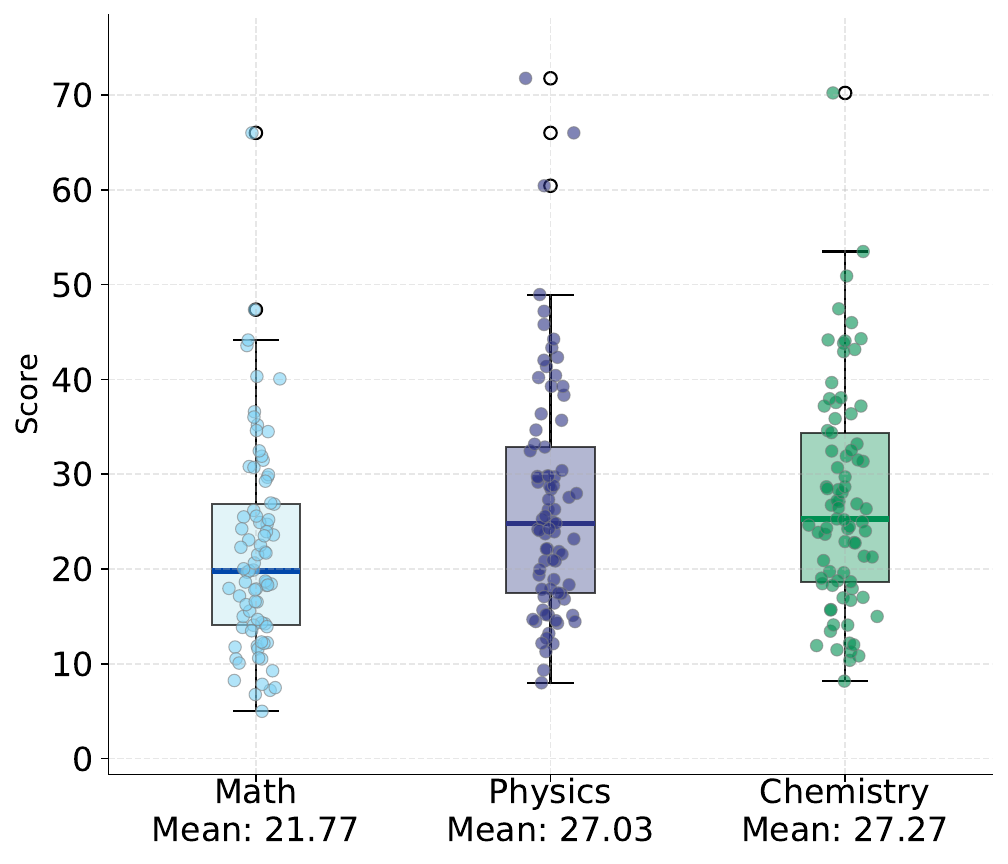}
        \caption{Score distribution.}
        \label{discipline}
    \end{minipage}
    \hfill
    \begin{minipage}{0.32\textwidth}
        \centering
        \includegraphics[width=0.98\linewidth]{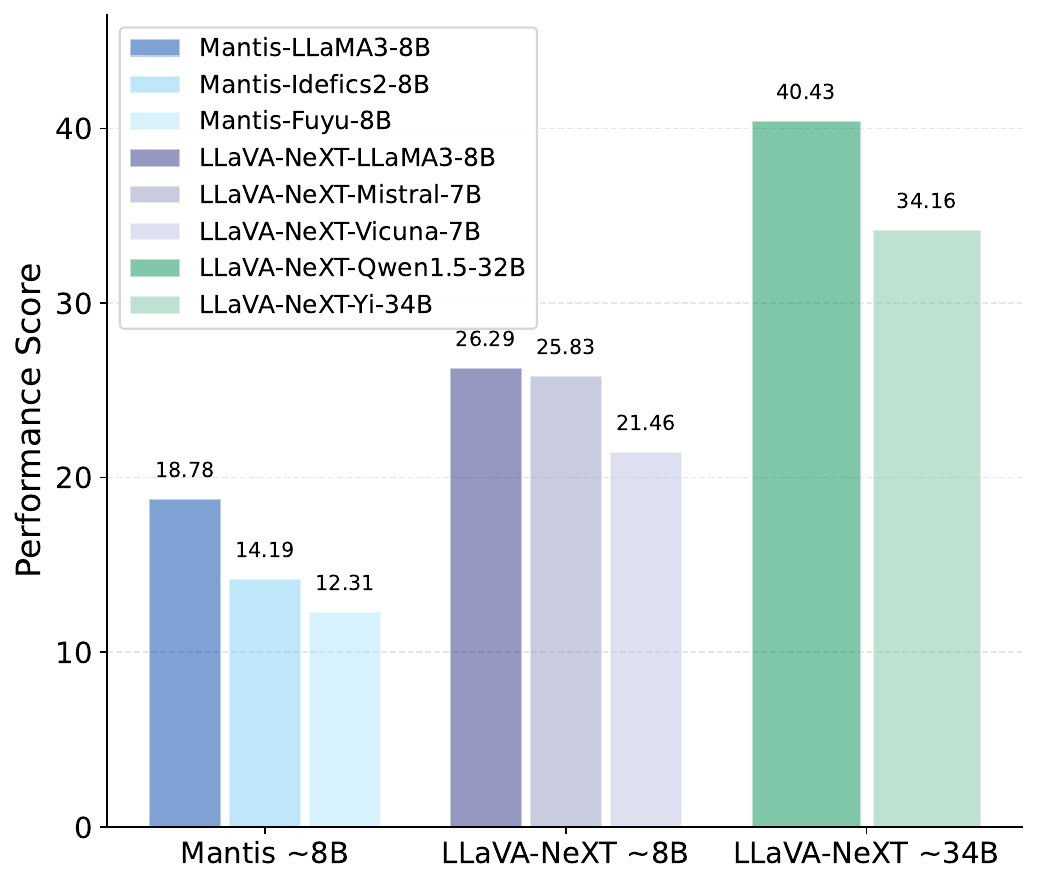}
        \caption{Impact of LLM backbones.}
        \label{llm_backbone}
    \end{minipage}
    \hfill
    \begin{minipage}{0.32\textwidth}
        \centering
        \includegraphics[width=0.98\linewidth]{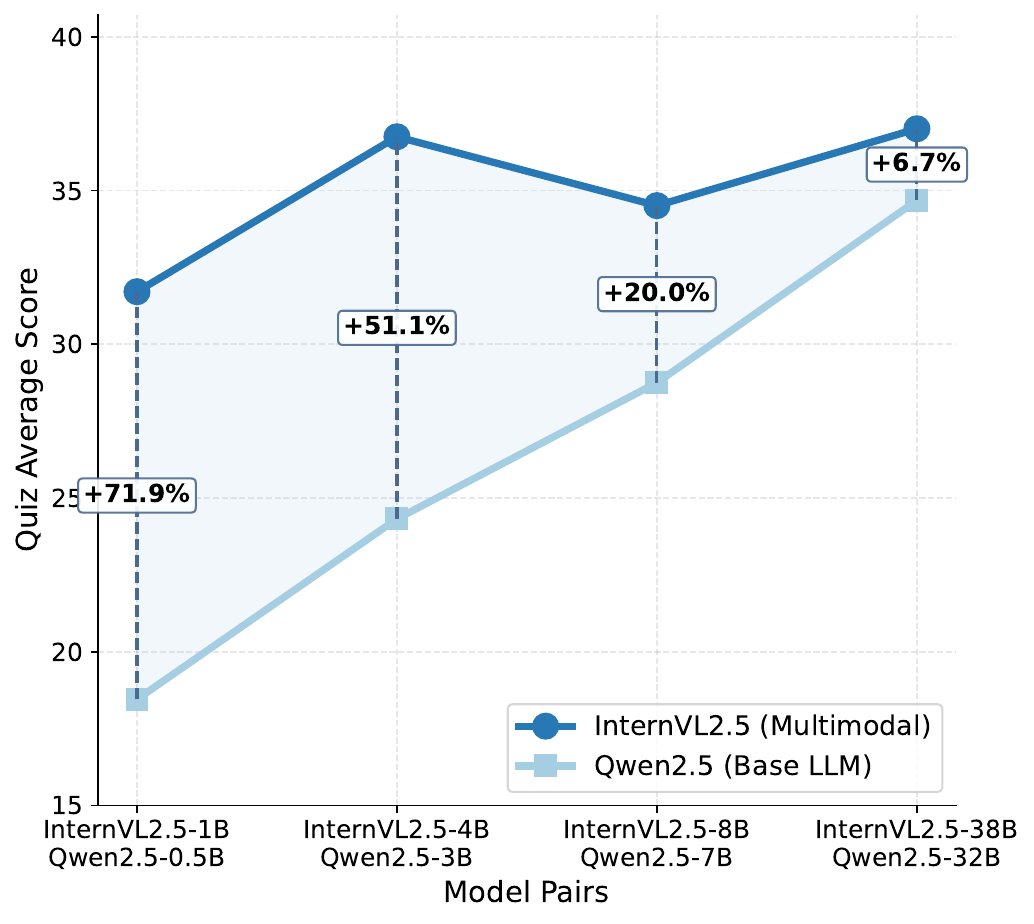}
        \caption{Impact of LLM size.}
        \label{vision_blind}
    \end{minipage}
\end{figure}

\begin{figure}[ht]
    \centering
    \includegraphics[width=\linewidth]{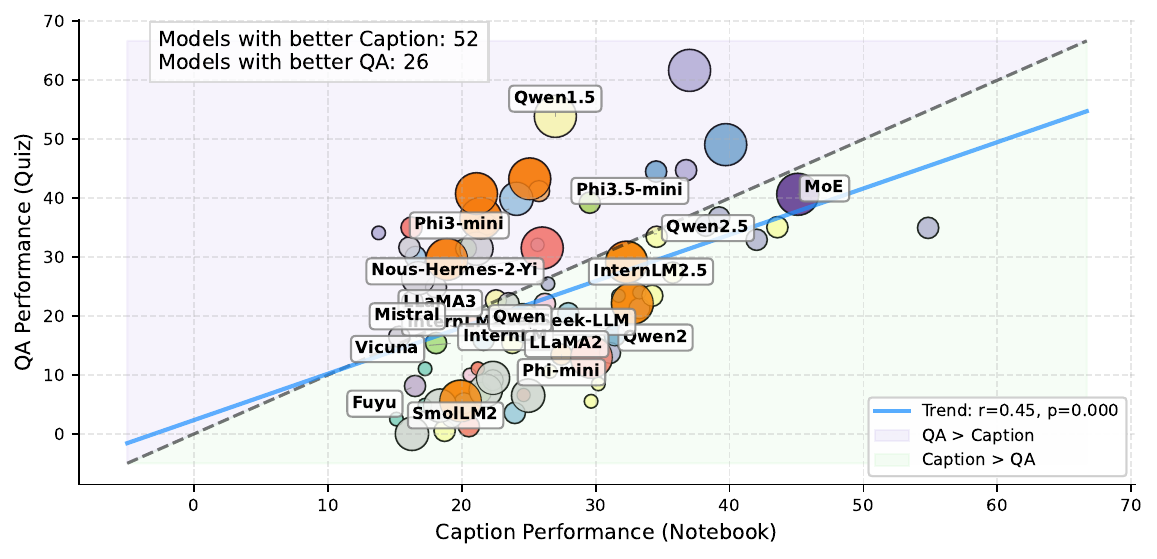}
    \caption{Relationship between captioning and QA performance across LLM.}
    \label{llm_structure}
\end{figure}

\begin{figure}[ht]
    \centering
    \includegraphics[width=0.7\linewidth]{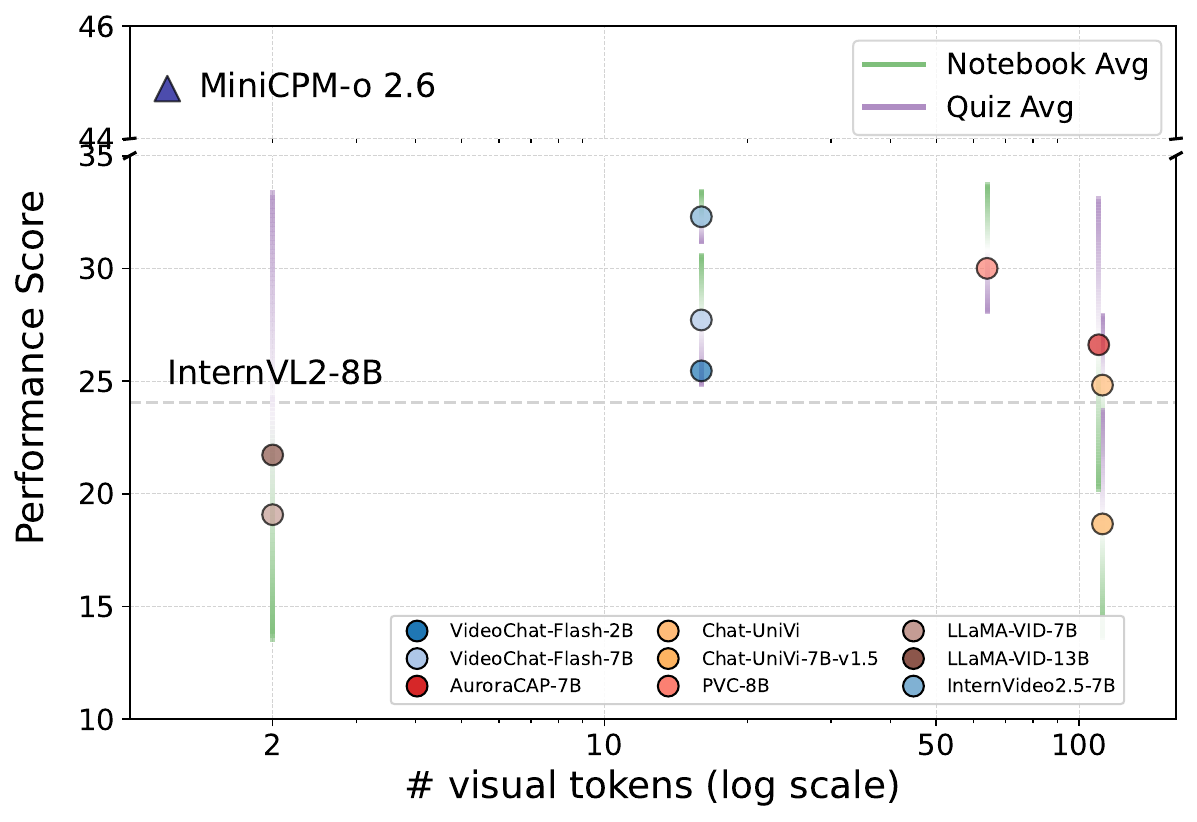}
    \caption{Impact of visual tokens number.}
    \label{token_compression}
\end{figure}

We select LMMs encompassing various design aspects such as architecture, training strategies and data mixtures. Our evaluation brings several important findings, as follows:

\paragraph{1) Proprietary models consistently outperform open-source models.}
As shown in Table~\ref{tab:evu_tiny}, Table~\ref{tab:evu_small} and Table~\ref{tab:evu_big}, both proprietary and open-source models perform poorly on~\benchmark, with accuracy mostly between 10\% and 50\%. Despite demonstrating competitive results in various video understanding and image OCR tasks, open-source models fall significantly behind in handling videos in~\benchmark, particularly in video detailed captioning. Among all evaluated models, \texttt{Claude-3.5-sonnet} achieves the highest performance across all tasks.



\paragraph{2) Lecture understanding in models relies more on textual content in frames than on animations.}
To comprehensively assess performance across different disciplines, we compute the average scores for both notebook and quiz tasks. Figure~\ref{discipline} shows that models significantly excel in physics and chemistry over mathematics.
Lecture videos in~\benchmark~reveals that physics and chemistry lectures contain more textual explanations, while mathematics lectures emphasizes formulas and dynamic visualizations. This suggests that existing LMMs primarily extract textual information but struggle with inferring complex logical relationships from animations. Consequently, their weaker performance in mathematics highlights challenges in handling dynamic abstract symbolic representations.


\vspace{-5pt}

\paragraph{3) Open-source video LMMs may not exhibit clear advantages over image LMMs.}
Most video LMMs~\cite{chai2024auroracap, chen2024internvl} are initialized with pre-trained image model weights and fine-tuned on video-text data to enhance temporal modeling without additional parameters. However, in~\benchmark, under the same architecture, ShareGPT4V-13B~\cite{chen2023sharegpt4v} (video LMM) underperforms LLaVA-1.5-Vicuna-13B~\cite{liu2024improved} (image LMM), despite additional video training. This may be due to the lack of OCR and visual knowledge reasoning tasks in video-text datasets, which are common in image-text training. Therefore, video LMMs struggle to transfer these capabilities to dynamic scenes, limiting their reasoning potential and temporal modeling advantages.

\paragraph{4) Large scale LMMs do not show clear advantages over smaller ones.}
The scaling law of LMMs~\cite{zohar2024apollo, deitke2024molmo} suggests that increasing model size significantly improves performance. While this trend persists in~\benchmark, its effect is less pronounced. Aria~\cite{li2024aria} (best under 5B) and MiniCPM-o-2.6~\cite{yao2024minicpm} (best under 8B) outperform CogVLM2~\cite{hong2024cogvlm2}, the strongest model around 20B. 
The size-performance scaling of the same model in~\benchmark~is not linear, as seen in InternVL2.5~\cite{cai2024internlm2}, which improves from 1B to 38B without proportional gains.  
In video captioning (Figure~\ref{caption_performance}), model size has a weak correlation with performance ($r=0.18$), suggesting that larger models do not necessarily generate better captions. Contrastly, video QA (Figure~\ref{qa_performance}) shows a stronger correlation ($r=0.40$), indicating that reasoning abilities benefit more from scaling. However, the trend remains inconsistent, with several models experiencing performance drops around 13B. A possible explanation is that larger models overfit training data focused on open-world understanding, limiting their generalization to multi-discipline lecture comprehension.


\paragraph{5) Larger LLMs enhance lecture understanding but with diminishing returns.}
Figure~\ref{vision_blind} compares QA performance in~\benchmark~between vision-blind Qwen2.5~\cite{qwen2.5} and selected InternVL2.5~\cite{chen2024expanding} trained on the same dataset with Qwen2.5~\cite{qwen2.5} as the LLM backbone. As model size increases, Qwen2.5~\cite{qwen2.5} exhibits continuous performance gains, while InternVL2.5~\cite{chen2024expanding} follows a similar trend, indicating that scaling up the LLMs enhances lecture understanding.
However, the performance gains gradually diminish with increasing model size, tapering to just a 6.7\% improvement at the largest scale. This indicates that while larger LLMs offer notable improvements, their advantage weakens at higher scales, highlighting opportunities to optimize training strategies for better leveraging strong LLMs in large-scale LMMs.


\paragraph{6) LLM Architecture shapes LMMs' balance between perception and reasoning.}
Figure~\ref{llm_backbone} confirms that LMMs performance is directly influenced by the ability of LLMs. Table~\ref{tab:evu_small} and Table~\ref{tab:evu_big} indicate that LMM performance in lecture captioning and QA tasks is likely shaped by the LLM architecture. Figure~\ref{llm_structure} visualizes performance distribution, with point colors representing architectures and sizes reflecting model scale.  
Most models excel in captioning over QA, highlighting the greater reasoning challenge lecture QA in~\benchmark. LMMs built on Qwen2.5~\cite{qwen2.5} and InternLM2.5~\cite{cai2024internlm2} achieve strong and balanced performance, while MoE-based LLMs also perform well. In contrast, ultra-small LLMs (SmolLM2~\cite{allal2025smollm2}) and decoder-only models (Fuyu~\cite{fuyu8b2024}) struggle with these tasks. Earlier architectures like Vicuna~\cite{chiang2023vicuna} and LLaMA2~\cite{li2023llama} perform poorly in QA for its weaker reasoning and instruction-following capabilities. This underscores the need to prioritize reasoning-focused LLMs (e.g., QwQ-32B~\cite{qwq32b}) for future advancements in lecture understanding.

\subsection{Variation Analysis}

As Large Multimodal Models (LMMs) excel across various tasks, model efficiency has gained increasing attention. Due to widespread redundancy in images and videos, most approaches~\cite{fu2024framefusion, vasu2024fastvlm, yang2024visionzip, han2024rethinking, han2025adafv} employ visual token compression based on similarity or attention scores. Experiments show these methods achieve competitive or better performance using only a few visual tokens compared to full-token models. However, it remains unclear: \textbf{Can LMMs with visual token compression sustain strong performance in complex, context-rich lecture understanding tasks like~\benchmark?} To explore this, we evaluate representative models on~\benchmark.

Figure~\ref{token_compression} presents a schematic comparison of performance and efficiency among visual token compression models. Since inference time varies significantly across architectures, models, deployment frameworks, and output lengths, we use the number of visual tokens per frame as a proxy of efficiency.
The wide performance variation among models with similar token counts suggests that architecture and compression strategy are as crucial as token quantity. Our findings indicate that significant token reduction is feasible while maintaining or even surpassing the performance of the base model. For instance, PVC-8B~\cite{yang2024pvc} (64 tokens per frame) achieves a 24.7\% performance improvement over its base model (InternVL2-8B~\cite{chen2024internvl}). However, models with ultra-low token counts suffer significant performance drops, like LLaMA-VID~\cite{li2023llamavid} (2 tokens). 
An optimal range of 16–300 tokens per frame balances efficiency and performance, with PVC-8B~\cite{yang2024pvc} (64 tokens) and VideoChat-Flash-7B~\cite{li2024videochat} (16 tokens) being particularly effective. The non-linear performance curve of AuroraCap-7B~\cite{chai2024auroracap} across different token counts underscores the need for domain-specific token compression optimization. 
Despite efficiency gains, a substantial gap remains between token-compressed models and the state-of-the-art. Even the best-performing compressed model, PVC-8B, lags far behind MiniCPM-o2.6~\cite{yao2024minicpm} (the leading 8B model), indicating challenges in preserving fine-grained details (e.g., formulas, numbers) essential for complex lecture reasoning with existing token-compressed models. Additionally, results from the LLaMA-VID~\cite{li2023llamavid} and VideoChat-Flash~\cite{li2024videochat} series suggest that larger model sizes can partially mitigate the information loss from compression.

\section{Conclusion}

In this paper, we introduce~\benchmark, a large-scale video-based benchmark for multi-discipline lecture understanding, designed to evaluate Large Multimodal Models (LMMs) in multimodal perception and reasoning within lecture comprehension. 
Our results show that both proprietary and open-source LMMs perform poorly, highlighting significant challenges in lecture understanding. Our analysis of visual token strategies and base LLM architectures provides valuable insights for guiding future research.

\section*{Acknowledgments} 
We acknowledge the
support of Lambda, Inc. for providing compute resources
for this project.


\clearpage
\bibliographystyle{plainnat}
\bibliography{main}

\clearpage
\beginsupplement
\clearpage
\renewcommand\thefigure{\Alph{section}\arabic{figure}}
\renewcommand\thetable{\Alph{section}\arabic{table}}
\setcounter{figure}{0}
\setcounter{table}{0}

\begin{center}
     \Large\textbf{Supplementary Material}
\end{center}

\noindent The supplementary material is structured as follows:

\begin{itemize}[leftmargin=7.5mm]
\setlength{\itemsep}{2pt}
\item Literature review about the related works in Section~\ref{supp:more_related}.
\item More details about the dataset construction in~\ref{sec:dataset_construction}.
\item Surface question-answer pairs generation prompt template in~\ref{sec:qa_generation}
\item Answer extraction from predicted captions prompt template in~\ref{sec:answer_generation}
\item Video reasoning-based question-answering generation prompt template in~\ref{sec:qa_generation}
\item More details about the dataset statistics in~\ref{sec:statistics}.
\item More details about the evaluation strategies in~\ref{sec:evaluation_strategy}.
\item Surface question-answering judgement prompt template  in~\ref{sec:caption_judge}.
\item Reasoning-based question-answering judgement prompt template  in~\ref{sec:qa_judge}.
\item More analysis about the model size and overall performance in~\ref{sec:overall}.
\item More analysis about the visual token reduction in~\ref{sec:visual_token}.

\end{itemize}
\section{Related Works}
\label{supp:more_related}
\subsection{Visual Token Compression}
Visual token compression has been widely employed to enhance efficiency in image and video LMMs~\cite{zhao2024stitch, zhao2024accelerating, ye2024atp, wu2024accelerating, huang2024efficient, xu2025learning, shao2025growing, li2025improving, zhang2025token, qi2025beyond, zeng2025skip, wang2025adaretake, yuan2025shortv}, reducing computational cost during training and inference.  Most strategies apply visual token compression before the LLM.
Applying pooling~\cite{maaz2023video, vasu2024fastvlm}, downsampling~\cite{xu2024pllava}, convolution~\cite{cheng2024videollama, zhang2025videollama3frontiermultimodal} via additional-training is straightforward but may introduce additional training cost. For question-answering tasks, ~\cite{lu2024b, huang2024prunevid, cheng2024focuschat, yamao2024iqvic} jointly train the compression module with the model to integrate question-related features.
Conversely, training-free methods leverage token similarity~\cite{chai2024auroracap, jeddi2025similarity}, relevance to the query~\cite{zhang2024fastervlm, zhong2024lyra}, or information content~\cite{wang2024llava} for compression. Feather~\cite{endo2024feather} is specifically designed for grounding tasks, where traditional compression methods struggle. DyCoke~\cite{tao2024dycoke} introduces a dynamic temporal token merging strategy. Additionally,~\cite{zhong2024aim} explores similarity-based token merging at the LLM layer to optimize compression further.
KVTP~\cite{liu2025keyframe} enhances efficiency on long-form video processing via keyframe-oriented vision token pruning

\section{More Details About~\benchmark~Construction}
\label{sec:dataset_construction}

\begin{figure}[ht]
    \centering
    \includegraphics[width=\textwidth]{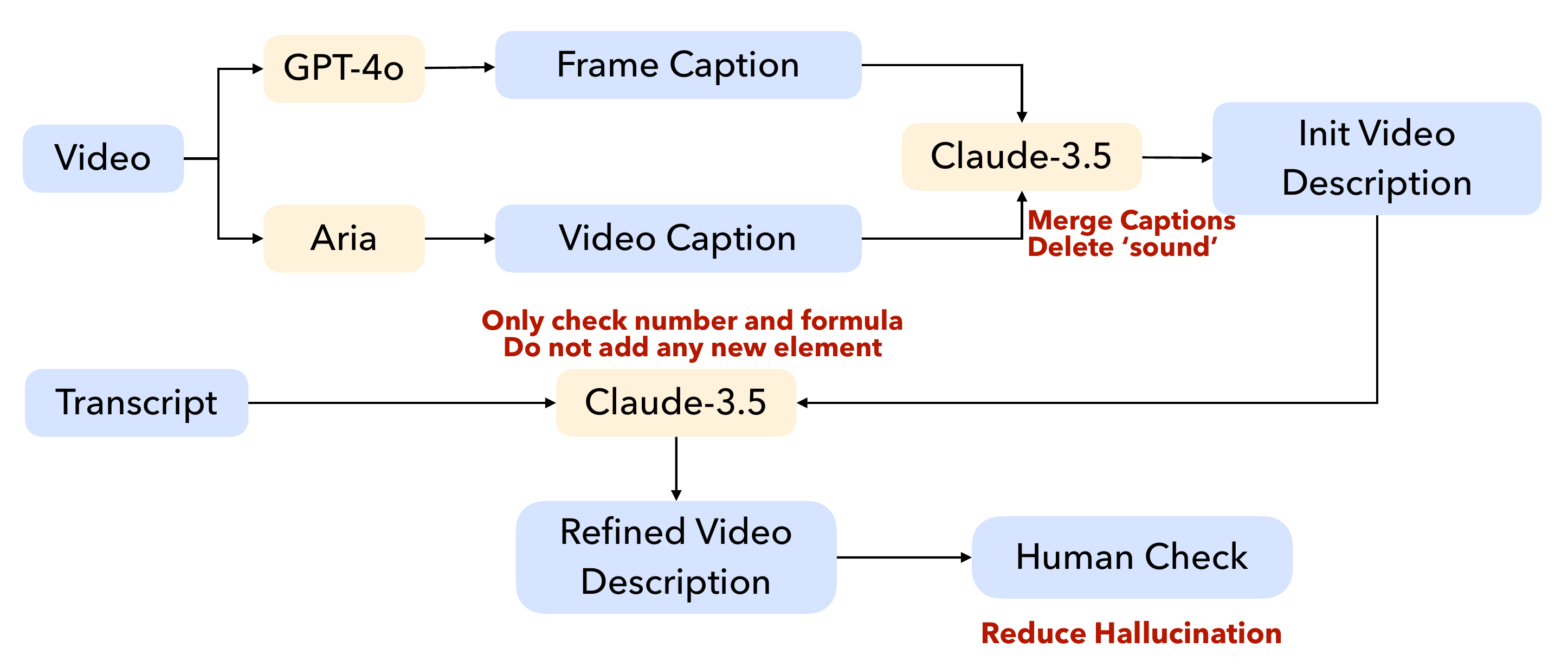}
    \caption{\benchmark~construction pipeline.}
    \label{fig:framework}
\end{figure}

For video caption, we first use Aria~\cite{li2024aria} to capture the temporal motion of the video from a global perspective. We set the sampling rate to 1 frame per second, with a resolution of 980.
We present the framework of ~\benchmark~construction process as shown in Figure~\ref{fig:framework} .

\section{Question-answer Pairs Generation Prompt Template of~\benchmark}
\label{sec:qa_generation}

To decompose the ground-truth structured detailed captions in~\benchmark, we utilize \texttt{Claude-3.5-sonnet} as the LLM assistant to generate numerous short question-answer pairs for subsequent evaluation. The full prompt and example cases are presented as followings:

\newpage

\vspace{3pt}
\begin{itemize}[leftmargin=12.5mm]
\setlength{\itemsep}{2pt}
    \item[\textbf{Type}] \textbf{Prompt}
    \vspace{3pt}
    \small{
    \item[SYSTEM] You are an intelligent chatbot designed for generating 20 question-answer pairs given a detailed description of a video or image. You are describing the video. 

    Here's how you can accomplish the task:
    INSTRUCTIONS: 
    - Cover the main objects and actions in the video or image.
    
    - The questions should be open-ended and start with 'What', 'Who', 'Where', 'When', 'Why', 'How', etc.
    
    - The answer should be a short sentence or phrase.
    
    - Generate 20 question-answer pairs.

    \item[User] Please generate 15 question-answer pairs given a detailed description of a video or image: detailed description: In this comprehensive mathematical exploration, the lecture delves into the fundamental nature of the exponential function $e^t$ by examining its defining properties and unique characteristics. The presentation begins with a crucial questioning approach, shifting from "What is $e^t$?\" to the more profound inquiry "What properties define $e^t$?" This methodological shift proves instrumental in understanding the function's essence …

    Please generate the response in the form of a Python list of tuple with the question and the corresponding answer. DO NOT PROVIDE ANY OTHER OUTPUT TEXT OR EXPLANATION. Only provide the Python list of tuple. For example, your response should look like this: [(the question 1, the answer 1), (the question 2, the answer 2), …].

    \item[\texttt{Claude}] 
    [{"question": "What background color is used in the visual presentation?", "answer": "Dark background"}, 
     {"question": "What is the initial condition of $e^t$ mentioned in the lecture?", "answer": "$e^0 = 1$"}, 
     {"question": "What is the approximate value of $e^{-2}$ shown in the presentation?", "answer": "0.135"}, 
     {"question": "What is the value of $e^{0.675}$ shown in the lecture?", "answer": "314.19"}, 
     {"question": "What is the approximate value of $e^{1.27}$?", "answer": "3.58"}, 
     {"question": "What is the approximate value of $e^{3.07}$?", "answer": "21.47"}, 
     {"question": "What is the approximate value of $e^{2.17}$?", "answer": "8.73"}, 
     {"question": "What are the strategic marker positions shown on the number line?", "answer": "25, 50, and 75"}, 
     {"question": "What is the approximate value of $e^{4.86}$?", "answer": "128.81"}, 
     {"question": "What is the approximate value of $e^{3.96}$?", "answer": "52.37"}, 
     {"question": "What is the approximate value of $e^{0.37}$?", "answer": "1.45"}, 
     {"question": "What is the typical span of the number line shown?", "answer": "0 to 9, sometimes extending to 800"}, 
     {"question": "What label is given to the horizontal number line?", "answer": "Position"}, 
     {"question": "What visual elements are used to show relationships on the number line?", "answer": "Arrows pointing to positions and velocities"}, 
     {"question": "What is the derivative of $e^{2t}$?", "answer": "$2 \cdot e^{2t}$"}]

    \item[User] [Manually check and refine]
}
\end{itemize}

\section{Predicted Answer Extraction Prompt Template}
\label{sec:answer_generation}

Given the question-answer pairs based on the ground truth caption, we utilize \texttt{Claude-3.5-sonnet} to extract predicted answers based on the generated caption by our designed prompt template. The complete prompt is shown as followings:

\vspace{3pt}
\begin{itemize}[leftmargin=12.5mm]
\setlength{\itemsep}{2pt}
    \item[\textbf{Type}] \textbf{Prompt}
    \vspace{3pt}
    \small{
    \item[SYSTEM] You are an intelligent chatbot designed for providing accurate answers to questions related to the content based on a detailed description of a video or image.
    
    Here's how you can accomplish the task:"
    
    ------
    
    \#\#INSTRUCTIONS: 
    
    - Read the detailed description carefully.
    
    - Answer the question only based on the detailed description.
    
    - The answer should be a short sentence or phrase.

    \item [User] Please provide accurate answers to questions related to the content based on a detailed description of a video or image:
    
    detailed description: This detailed mathematics tutorial video provides comprehensive instruction on applying the product rule for derivatives, featuring a consistent bright green background and an engaging male instructor positioned in the lower right corner, dressed in a dark jacket. The instructor maintains an enthusiastic and approachable teaching style throughout the presentation, making complex calculus concepts more accessible to viewers.
    
    question: What color is the video background?

    DO NOT PROVIDE ANY OTHER OUTPUT TEXT OR EXPLANATION. Only provide short but accurate answer.

    \item[\texttt{Claude}] Bright green.
}
\end{itemize}

\section{More details about~\benchmark~Statistics}
\label{sec:statistics}

We collect videos from the open-source platform YouTube, primarily sourced from ten video creators. Their channel homepage links are as follows:
\begin{itemize}
    \item \href{https://www.youtube.com/@brain_station_videos}{brain-station-videos}
    \item \href{https://www.youtube.com/@cpmrana}{cpmrana}
    \item \href{https://www.youtube.com/@AndyMath}{AndyMath}
    \item \href{https://www.youtube.com/@aleksandr-physics}{aleksandr-physics}
    \item \href{https://www.youtube.com/@3blue1brown}{3blue1brown}
    \item \href{https://www.youtube.com/@LittleSaigonWack}{LittleSaigonWack}
    \item \href{https://www.youtube.com/@ProfessorDaveExplains}{ProfessorDaveExplains}
    \item \href{https://www.youtube.com/@DrTrefor}{DrTrefor}
    \item \href{https://www.youtube.com/@MathVisualProofs}{MathVisualProofs}
    \item \href{https://www.youtube.com/@TheOrganicChemistryTutor}{TheOrganicChemistryTutor}
\end{itemize}

Figure~\ref{fig:benchmark_inf} and Figure~ indicates the word distribution of the video detailed captions and reasoning question-answering in~\benchmark. 
\begin{figure}[ht]
    \centering
    \includegraphics[width=0.5\textwidth]{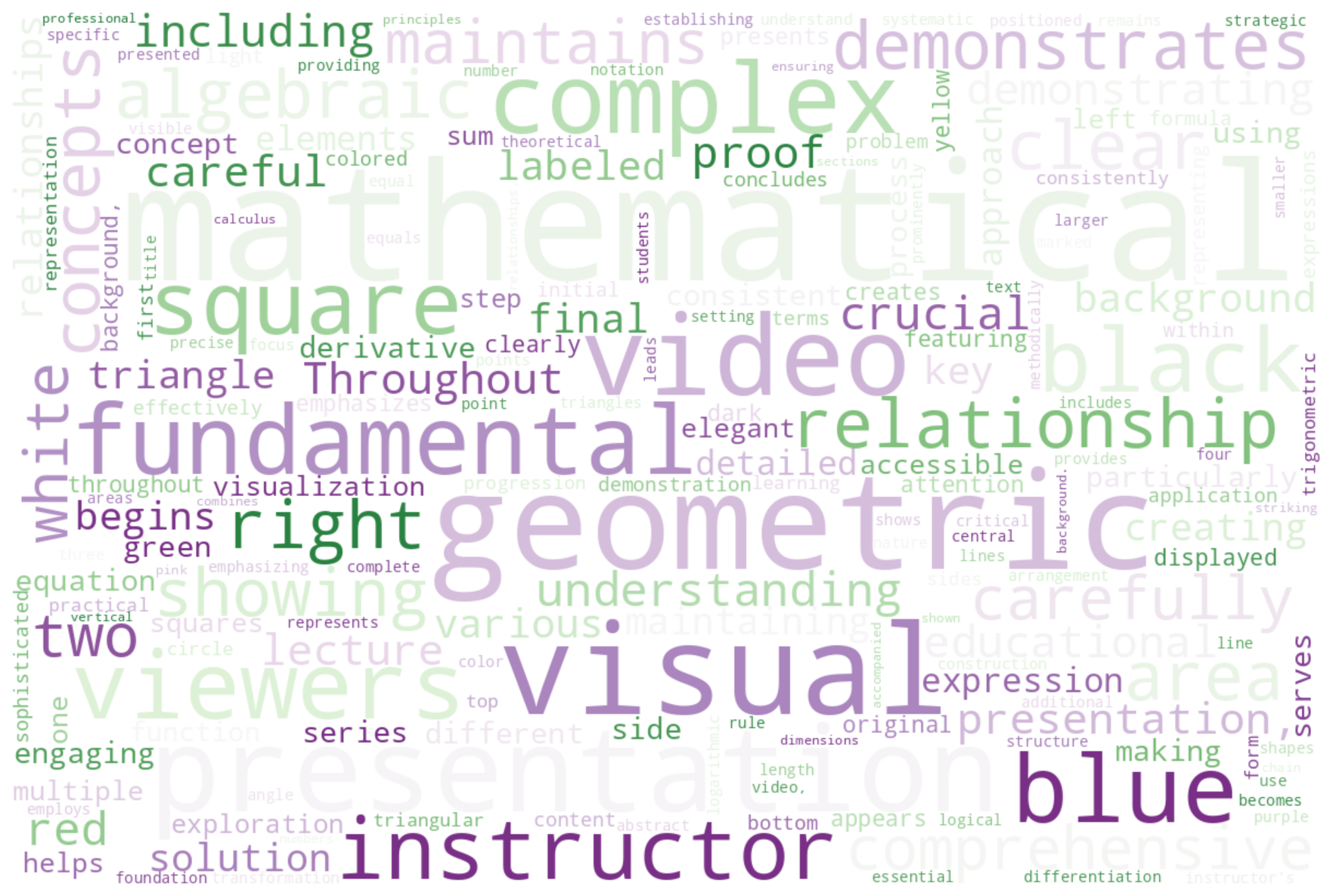}
    \caption{Word cloud of lecture detailed captions in~\benchmark, showing the diversity.}
    \label{fig:benchmark_inf}
\end{figure}

\begin{figure*}[ht]
    \centering
    \begin{tabular}{cccc}
        \subfloat[Surface Question]{\includegraphics[width=0.4\textwidth]{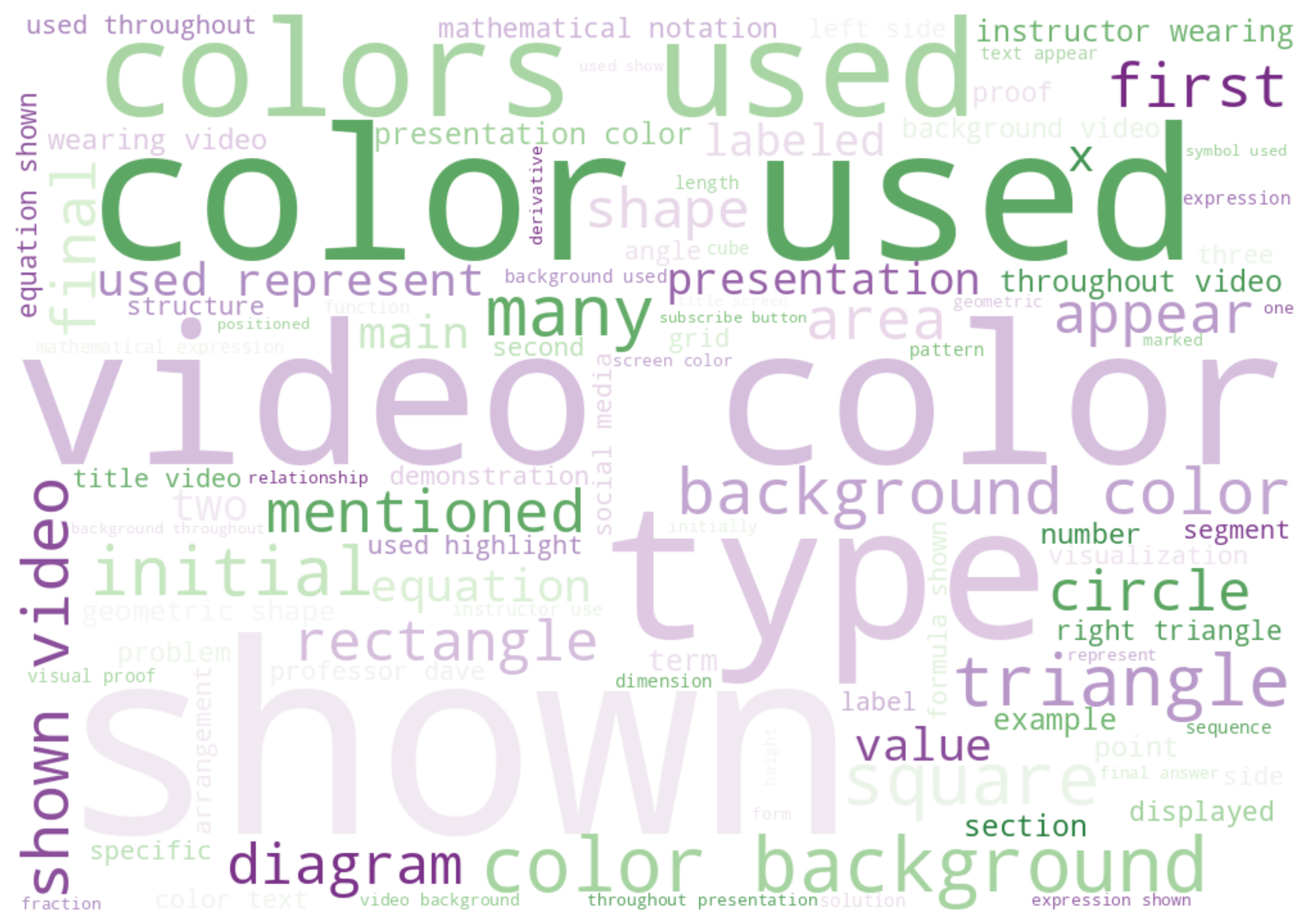}} &
        \subfloat[Surface Answer]{\includegraphics[width=0.4\textwidth]{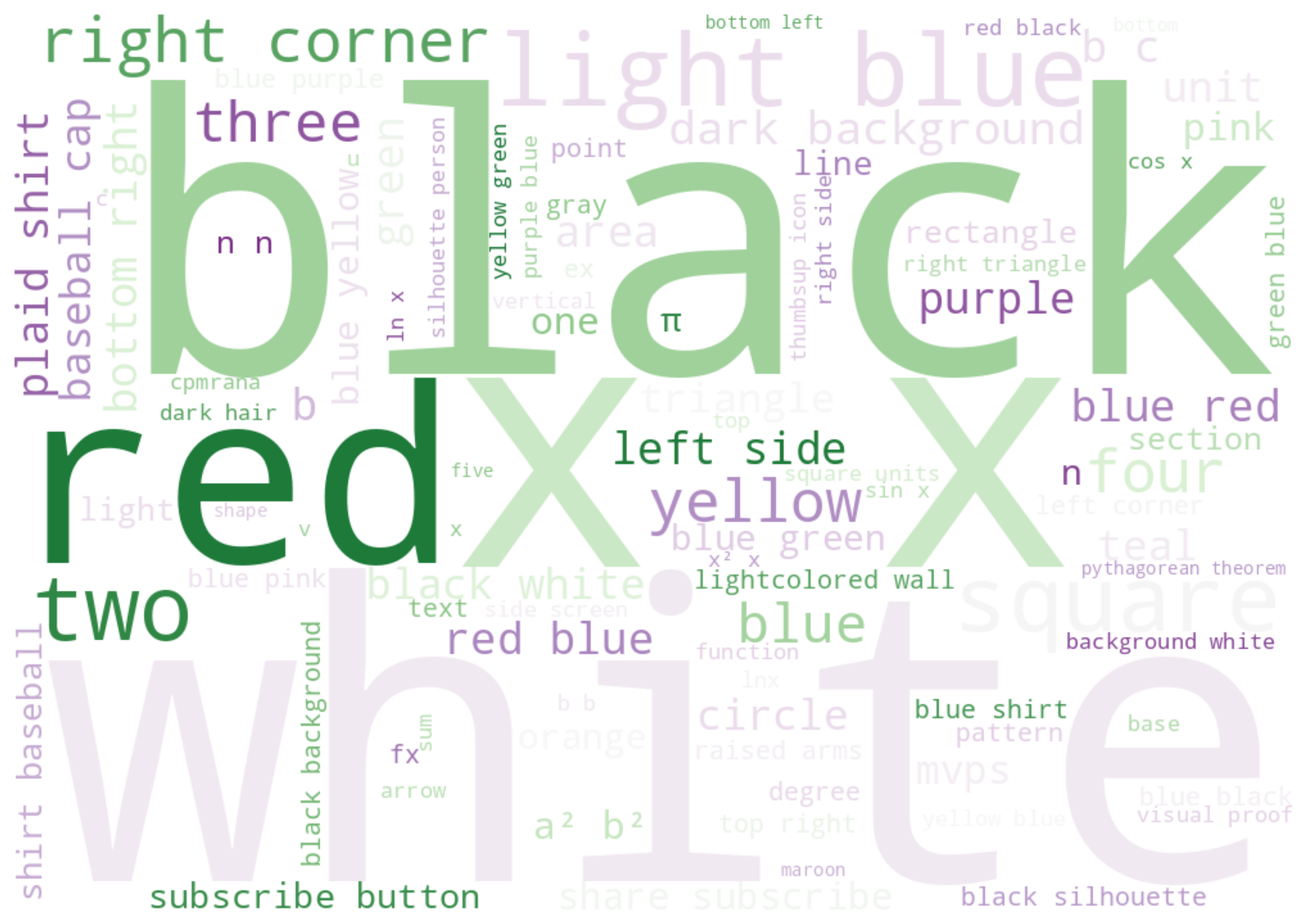}} \\
        \subfloat[Deeper Question]{\includegraphics[width=0.4\textwidth]{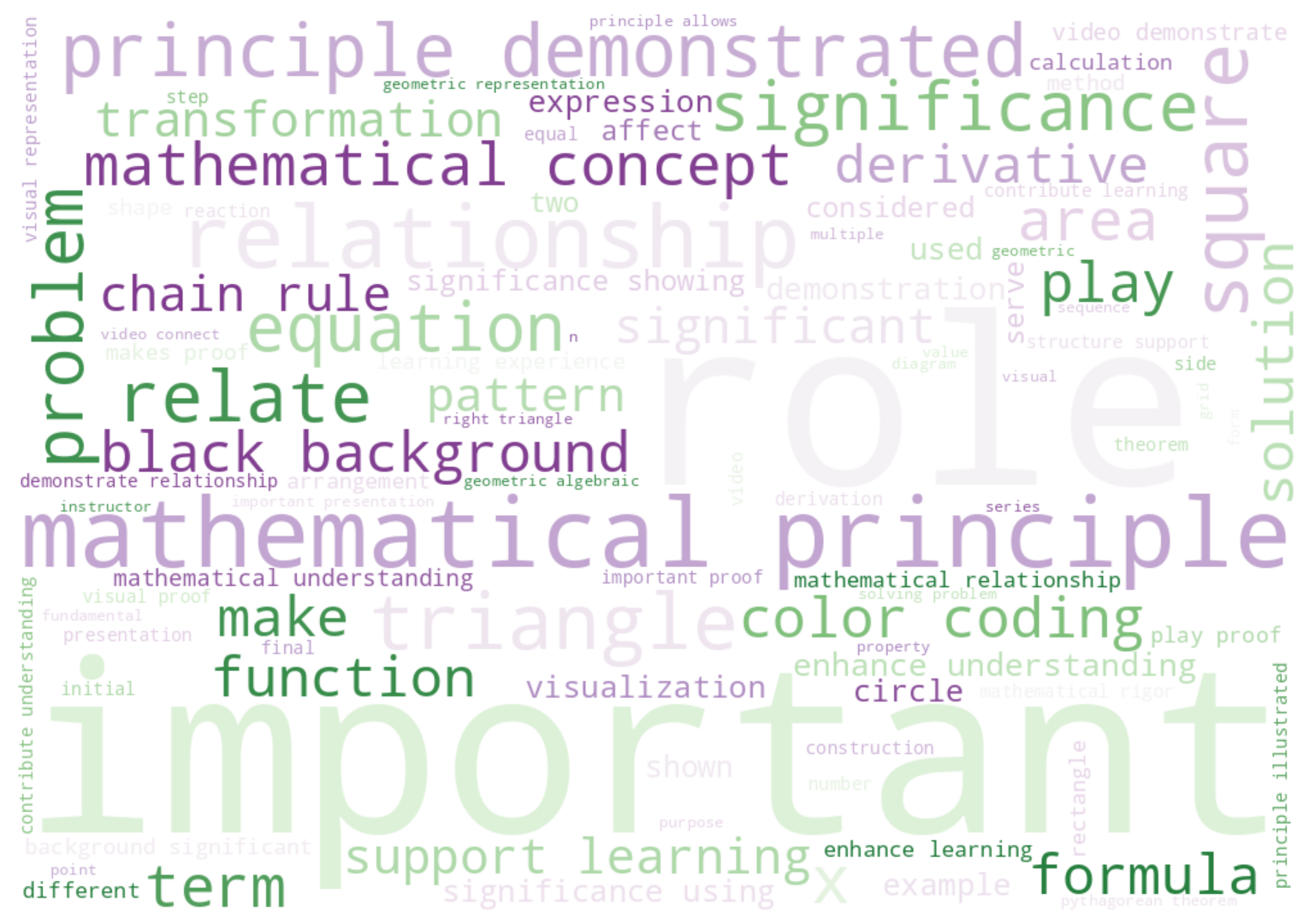}} &
        \subfloat[Deeper Answer]{\includegraphics[width=0.4\textwidth]{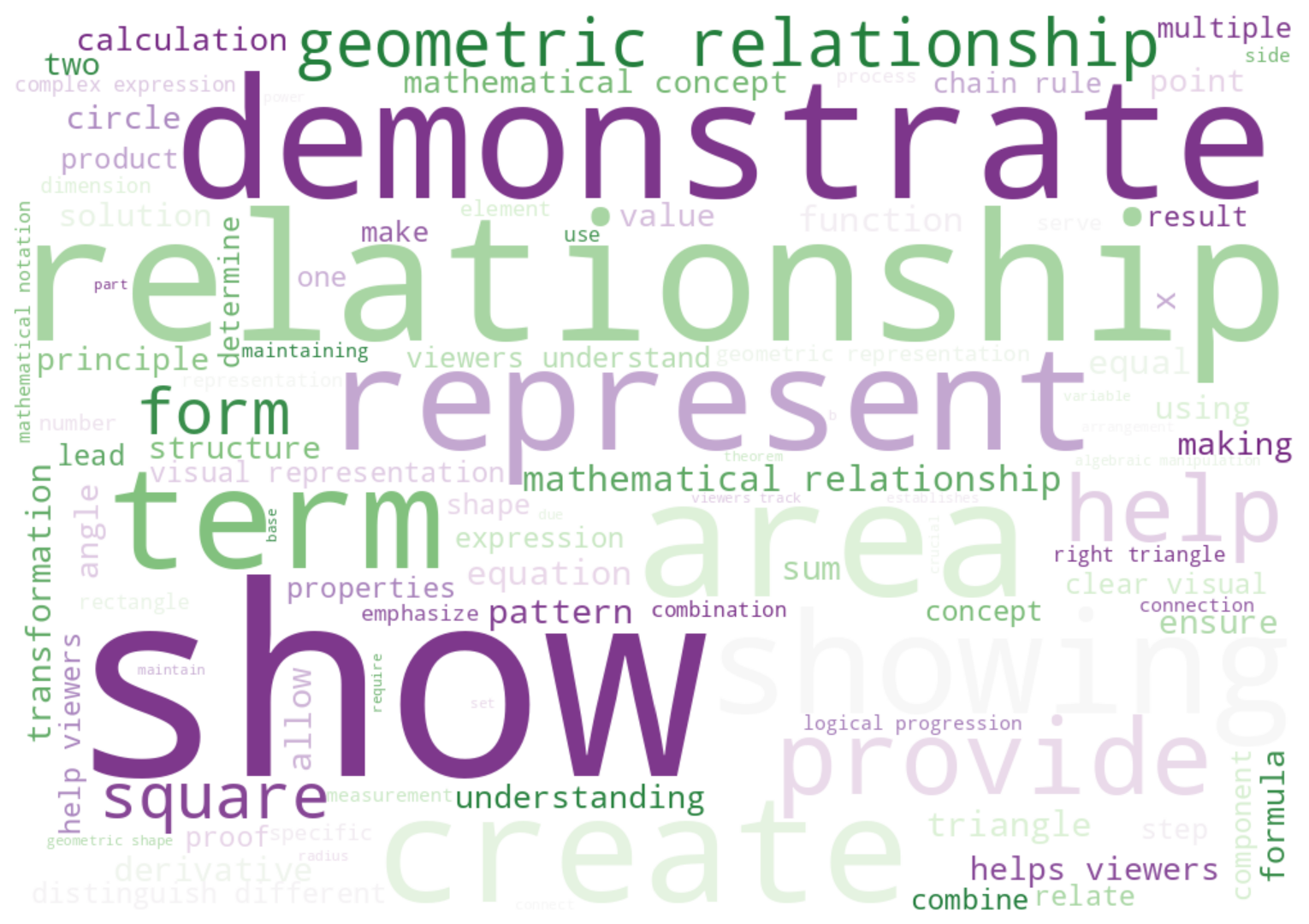}} \\
    \end{tabular}
    \caption{Word cloud of different question-answering pairs in~\benchmark, showing the diversity.}
    \label{fig:benchmark_inf}
\end{figure*}

\section{More Details about the Evaluation Strategies}
\label{sec:evaluation_strategy}

The latest versions of the evaluated properietary models before March 2025 were \texttt{Gemini-1.5-Flash-002}, \texttt{GPT-4o-2024-05-13}, and \texttt{Claude-3.5-sonnet-20241022}.

For the visual QA track, we require models to provide concise responses via the system prompt `Answer briefly and directly in one sentence.' and limit \texttt{max\_new\_tokens} to 64. For the video captioning track, we allow \texttt{max\_new\_tokens} to follow each model’s default setting and prompt with the designed instruction in different length randomly as follow:
\begin{itemize}
 \item The images are given containing equally spaced video frames. Please imagine the video based on the sequence of frames, and provide a faithfully detailed description of this video in more than three sentences.
 \item You are given a sequence of equally spaced video frames. Based on these frames, imagine the full video and provide a detailed description of what is happening in more than three sentences.
 \item The following set contains equally spaced video frames. Imagine the video from which these frames were taken and describe it in detail in at least three sentences.
 \item Below are equally spaced frames from a video. Use these frames to visualize the entire video and provide a detailed description in more than three sentences.
 \item A sequence of equally spaced video frames is presented. Please imagine the full video and write a faithfully detailed description of the events in more than three sentences.
 \item The images provided include equally spaced frames from a video. Based on these frames, imagine the video and describe it comprehensively in at least three sentences.
 \item You are given equally spaced frames from a video. Use these frames to envision the entire video and provide a detailed description of the events in more than three sentences.
 \item The sequence includes equally spaced frames from a video. Imagine the full video based on these frames and provide a detailed description in more than three sentences.
 \item The provided images contain equally spaced frames from a video. Visualize the video from these frames and describe it in detail in more than three sentences.
 \item Here are equally spaced frames from a video. Based on these frames, imagine the video and provide a detailed, faithful description of it in more than three sentences.
 \item The set of images includes equally spaced video frames. Please imagine the video these frames come from and describe it comprehensively in at least three sentences.
 \item Describe the video based on these frames in a few sentences.
 \item Explain the video using these frames.
 \item Imagine the video from these frames and describe it in detail in a few sentences.
 \item Based on these frames, provide a narrative of the video in more than three sentences.
 \item Describe the events in the video shown by these frames in at least three sentences.
 \item Visualize the video from these frames and explain what is happening in more than three sentences.
 \item Describe the sequence of events in the video depicted by these frames in a detailed manner.
 \item Given these equally spaced frames, imagine the entire video and provide a detailed description of the events, including the setting, characters, and actions, in more than three sentences.
 \item Visualize the video based on these frames and write a comprehensive description of what happens, describing the beginning, middle, and end in at least three sentences.
 \item Using these frames as a reference, imagine the full video and provide a thorough description of the plot, including key details and actions, in more than three sentences.
 \item Based on the sequence of these frames, describe the entire video in detail, mentioning important aspects such as the context, movements, and transitions in more than three sentences.
 \item Imagine the video that corresponds to these frames and provide an elaborate description, covering the storyline, visual elements, and any notable features in at least three sentences.
\end{itemize}

We use Qwen2.5-72B~\cite{yang2024qwen2} as the LLM evaluation assistant and accelerate with LMDeploy~\cite{2023lmdeploy}.

\section{Correctness Evaluation for Detailed Captioning Prompt Template}
\label{sec:caption_judge}

Following~\cite{maaz2023video}, we evaluate the correctness and score of the predicted answers with the assistant of Qwen2.5-72B~\cite{qwen2.5}. Given the question, correct answer, and predicted answer from the generated caption, \texttt{Qwen2.5-72B~\cite{qwen2.5}} should return the \textit{True} or \textit{False} judgement and relative score ($0$ to $5$). We specially design a strict prompt for OCR-related question-answering evaluation. The complete prompt is shown as followings:

\vspace{3pt}
\begin{itemize}[leftmargin=12.5mm]
\setlength{\itemsep}{2pt}
    \item[\textbf{Type}] \textbf{Prompt}
    \vspace{3pt}
    \small{
    \item[SYSTEM] You are an intelligent chatbot designed for evaluating the correctness of generative outputs for question-answer pairs.

    Your task is to compare the predicted answer with the correct answer and determine if they match meaningfully. The evaluation criteria differ based on the type of question:
    
    ------
    
    \#\#INSTRUCTIONS: 

    1. \textbf{For OCR-related questions}:
    
    - Perform a strict letter-by-letter comparison.
    
    - Any difference in characters (including case, punctuation, or letter substitution) must result in 'no'.
    
    - Minor spelling errors or missing characters should not be accepted.

    2. For \textbf{non-OCR-related questions}:
    
    - Focus on the meaningful match between the predicted answer and the correct answer.
    
    - Synonyms or paraphrases can be considered valid matches.
    
    - Minor spelling differences or alternative expressions should not be penalized.

    \item [User] Please evaluate the following video-based question-answer pair:
    
    Question: What specific DNA sequence is shown in the presentation?
    
    Correct Answer: TCCGTGCAGTAAATGC
    
    Predicted Answer: TTCCGTAATACGACTGCGC
    
    Provide your evaluation only as a yes/no and score where the score is an integer value between 0 and 5, with 5 indicating the highest meaningful match. 
    
    Please generate the response in the form of a Python dictionary string with keys 'pred' and 'score', where value of 'pred' is  a string of 'yes' or 'no' and value of 'score' is in INTEGER, not STRING.
    
    DO NOT PROVIDE ANY OTHER OUTPUT TEXT OR EXPLANATION. Only provide the Python dictionary string. 
    
    For example, your response should look like this: \{'pred': 'yes', 'score': 4.8\}.
    
    \item[Qwen2.5] \{'pred': 'no', 'score': 0\}
}
\end{itemize}

\section{Correctness Evaluation for Reasoning QA Prompt Template}
\label{sec:qa_judge}

Following~\cite{maaz2023video}, we evaluate the correctness and score of the predicted answers with the assistant of Qwen2.5-72B~\cite{qwen2.5}. Given the question, correct answer, and predicted answer from the generated caption, Qwen2.5-72B~\cite{qwen2.5} should return the \textit{True} or \textit{False} judgement and relative score ($0$ to $5$). We ask the LLM assistan focus on the evaluation of the reasoning process. The complete prompt is shown as followings:

\vspace{3pt}
\begin{itemize}[leftmargin=12.5mm]
\setlength{\itemsep}{2pt}
    \item[\textbf{Type}] \textbf{Prompt}
    \vspace{3pt}
    \small{
    \item[SYSTEM] You are an intelligent chatbot designed for evaluating the correctness of generative outputs for reasoning-based question-answer pairs.

    Your task is to compare the predicted answer with the correct answer based on the following rules:
    
    ------
    
    \#\#INSTRUCTIONS: 
    
    1. \*\*Evaluate Reasoning Tasks Strictly:\*\*
    
    - The predicted answer must capture all critical concepts and details mentioned in the correct answer.
    
    - If the correct answer mentions specific concepts or examples (e.g., 'odd numbers accumulate to form perfect squares'), the predicted answer must include these concepts or examples.
    
    - Even if the phrasing differs, the key meaning and concepts must be preserved. However, omitting or altering key concepts or examples is not acceptable.

     - \textbf{Example 1}: If the correct answer is 'The construction method shows how odd numbers accumulate to form perfect squares,' the predicted answer must include 'odd numbers' and 'perfect squares.'
    - \textbf{Example 2: }If the correct answer is 'To eliminate HBr and form an alkene,' the predicted answer must address the elimination of HBr as well.
    - Minor differences in phrasing are acceptable as long as the key information is retained.
    - \textbf{Critical Detail:} If any essential element (e.g., key terms, concepts, or examples) is missing from the predicted answer, the answer is considered incorrect.
    - Do \textbf{not} introduce new, unrelated information in the predicted answer.

    \item [User] Please evaluate the following video-based question-answer pair:
    
    Question: What role does RNA polymerase play in the lac operon system?
    
    Correct Answer: It initiates transcription of the structural genes when allowed access to the promoter region
    
    Predicted Answer: RNA polymerase binds to the promoter gene and transcribes the structural genes when lac operon is active.
    
    Provide your evaluation only as a yes/no and score where the score is an integer value between 0 and 5, with 5 indicating the highest meaningful match. 
    
    Please generate the response in the form of a Python dictionary string with keys 'pred' and 'score', where value of 'pred' is  a string of 'yes' or 'no' and value of 'score' is in INTEGER, not STRING.
    
    DO NOT PROVIDE ANY OTHER OUTPUT TEXT OR EXPLANATION. Only provide the Python dictionary string. 
    
    For example, your response should look like this: \{'pred': 'yes', 'score': 4.8\}.
    
    \item[Qwen2.5] \{'pred': 'yes', 'score': 5\}
}
\end{itemize}

\section{More Analysis About the Model Size and Overall Performance}
\label{sec:overall}
\begin{figure*}[htbp]
    \centering
    \includegraphics[width=\textwidth]{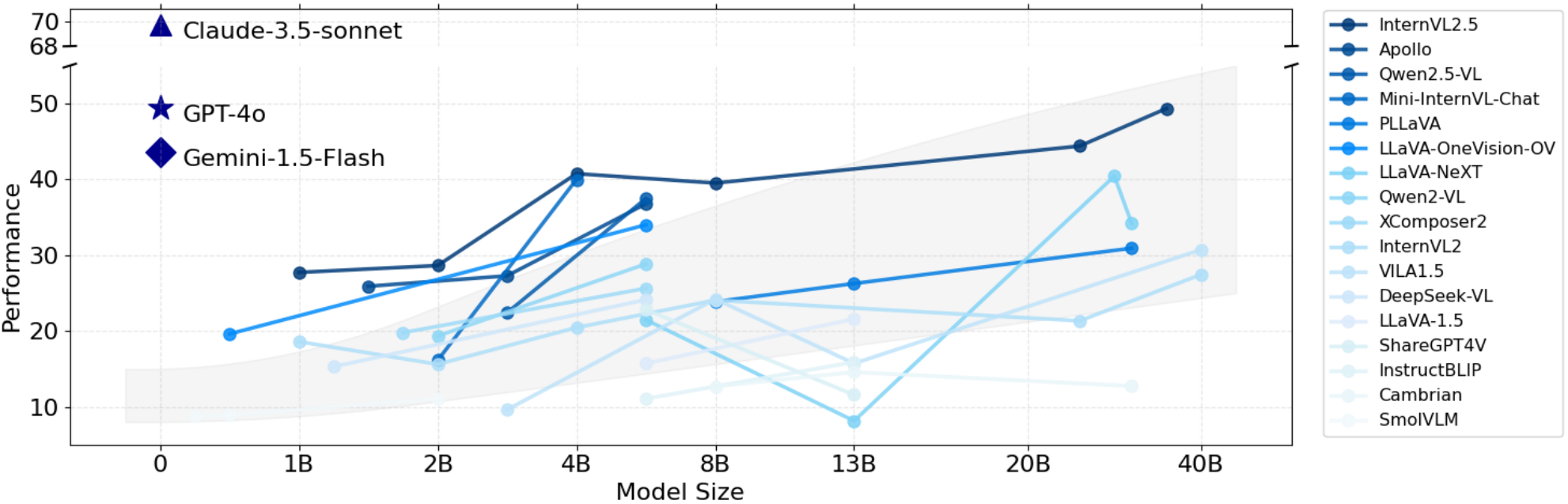}
    \caption{\textbf{Relationship between model size and the average video captioning and question-answering performance.} The shaded region shows the confidence interval, with darker colors indicating better performance.}
    \label{overall_performance}
\end{figure*}

As shown in Figure~\ref{sec:overall}, we visualize the relationship between model size and the average performance across video detailed captioning and question-answering tasks.
Generally, larger models (over 20B) tend to achieve better performance, but the scaling trend is not strictly linear. While some models, such as InternVL2.5~\cite{chen2024expanding}, show consistent improvements as size increases, others~\cite{tong2024cambrian, li2024llavanext} exhibit fluctuating gains. Notably, certain mid-sized models (e.g., around 8B–13B parameters) might outperform larger ones, suggesting that beyond model size, architecture and training strategies play a crucial role in lecture understanding. Additionally, proprietary models like \texttt{Gemini-1.5-Flash}, \texttt{GPT-4o}, and \texttt{Claude-3.5-sonnet} significantly outperform open-source models, highlighting a substantial gap in LMM capabilities. 
\section{More Analysis about the Visual token Reduction}
\label{sec:visual_token}



\begin{table*}[t]
\centering
\renewcommand{\arraystretch}{1.08}
\small
\caption{Results of visual token compression models on~\benchmark.}
\resizebox{\textwidth}{!}{%
\begin{tabular}{l cc| c |cccc| cccc }
\toprule
\multirow{2}{*}{\textbf{Models}} & \multirow{2}{*}{\textbf{LLM}} & \multirow{2}{*}{\textbf{\#~Tokens Per Frame}} & \multirow{2}{*}{\textbf{Overall}} & \multicolumn{4}{c|}{\textbf{Notebook}} & \multicolumn{4}{c}{\textbf{Quiz}} \\
& & & & \textbf{Avg.} & \textbf{Math} & \textbf{Physics} & \textbf{Chemistry} & \textbf{Avg.} & \textbf{Math} & \textbf{Physics} & \textbf{Chemistry}\\
\midrule
Chat-UniVi~\cite{jin2024chat} & Vicuna~\cite{chiang2023vicuna}-7B & 112 & 24.82 & 21.73 & 16.68 & 26.97 & 21.55 & 27.91 & 23.85 & 25.35 & 34.55  \\
Chat-UniVi-7B-v1.5~\cite{jin2024chat} & Vicuna~\cite{chiang2023vicuna}-7B & 112 & 18.66 & 13.62 & 10.30 & 16.84 & 13.74 & 23.70 & 21.21 & 23.92 & 25.98 \\
\multirow{2}{*}{LLaMA-VID~\cite{li2023llamavid}} & Vicuna~\cite{chiang2023vicuna}-7B & 2 & 19.07 & 13.87 & 8.11 & 14.16 & 19.33 & 24.27 & 12.32 & 20.24 & 40.25 \\
 & Vicuna~\cite{chiang2023vicuna}-13B & 2 & 21.72 & 13.53 & 10.08 & 12.50 & 18.00 & 33.35 & 28.08 & 26.63 & 45.34 \\
\multirow{12}{*}{AuroraCap~\cite{chai2024auroracap}} & \multirow{11}{*}{Vicuna~\cite{chiang2023vicuna}-7B} & 48 & 21.45 & 18.36 & 16.75 & 15.00 & 23.33 & 23.65 & 19.46 & 12.24 & 39.26 \\
&  & 79 & 22.19 & 19.55 & 16.49 & 17.50 & 24.66 & 24.84 & 21.23 & 13.34 & 39.97  \\
&  & 110 & 26.61 & 20.18 & 16.06 & 19.16 & 25.33 & 33.09 & 21.08 & 38.08 & 40.11  \\
&  & 172 & 23.31 & 19.62 & 17.35 & 17.50 & 24.00 & 27.00 & 19.17 & 26.67 & 35.17  \\
&  & 265 & 22.90 & 19.55 & 16.74 & 26.66 & 15.25 & 25.14 & 25.34 & 20.12 & 29.96   \\
&  & 327 & 27.53 & 17.84 & 15.64 & 22.56 & 15.33 & 37.21 & 28.08 & 33.31 & 50.23   \\
&  & 389 & 26.21 & 18.97 & 17.00 & 22.57 & 17.35 & 33.44 & 23.28 & 26.93 & 50.12  \\
&  & 451 & 25.86 & 19.31 & 13.93 & 21.34 & 22.65 & 31.44 & 23.27 & 20.73 & 50.33   \\
&  & 544 & 21.60 & 19.17 & 16.32 & 20.83 & 20.35 & 26.32 & 20.54 & 13.43 & 45.00  \\
&  & 606 & 21.87 & 18.78 & 15.21 & 20.83 & 20.31 & 23.95 & 23.54 & 12.97 &  35.35 \\
&  & 668 & 17.10 & 14.13 & 12.56 & 15.83 & 14.00 & 20.07 & 22.61 & 6.57 & 31.03  \\
&  & 730 & 20.83 & 16.78 & 14.95 & 21.33 & 14.07 & 25.22 & 23.30 & 6.64 & 45.71  \\
\multirow{2}{*}{VideoChat-Flash~\cite{li2024videochat}} & Qwen2.5~\cite{yang2024qwen2}-2B & 16 & 25.45 & 27.03 & 25.09 & 25.15 & 30.86 & 25.53 & 21.64 & 9.93 & 45.02 \\
 & Qwen2~\cite{wang2024qwen2}-7B & 16 & 27.71 & 30.58 & 30.17 & 33.43 & 28.15 & 24.83 & 17.82 & 13.39 & 43.29 \\
InternVideo2.5~\cite{wang2025internvideo2} & InternLM2.5~\cite{cai2024internlm2}-7B & 16 & 32.29 & 33.40 & 29.74 & 30.05 & 40.42 & 31.18 & 24.65 & 33.27 & 35.62 \\
PVC~\cite{yang2024pvc} & InternLM2.5~\cite{cai2024internlm2}-7B & 64 & 30.00 & 33.70 & 27.43 & 38.33 & 35.33& 26.29 & 28.07 & 20.53 & 30.29  \\
\bottomrule
\end{tabular}%
}
\label{tab:token}
\end{table*}

\begin{figure*}[!ht]
    \centering
    \begin{tabular}{ccc}
        \subfloat{\includegraphics[width=0.3\textwidth]{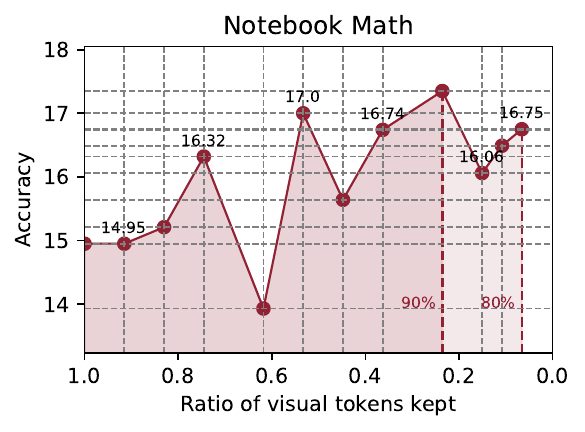}} &
        \subfloat{\includegraphics[width=0.3\textwidth]{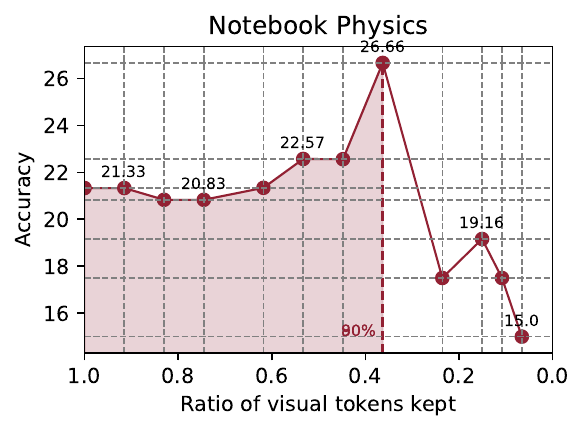}} &
        \subfloat{\includegraphics[width=0.3\textwidth]{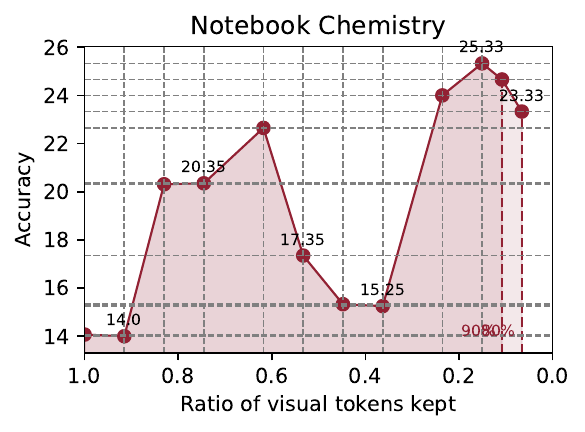}} \\
        \subfloat{\includegraphics[width=0.3\textwidth]{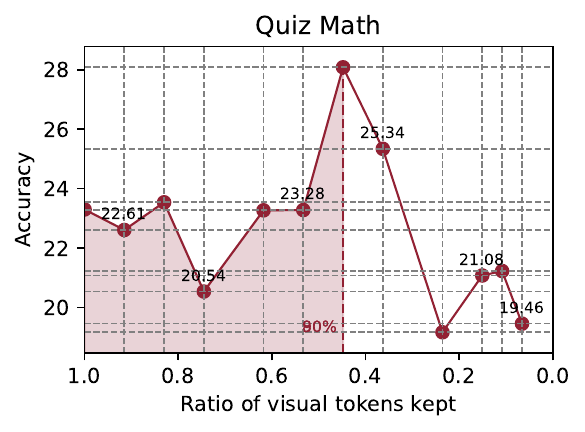}} &
        \subfloat{\includegraphics[width=0.3\textwidth]{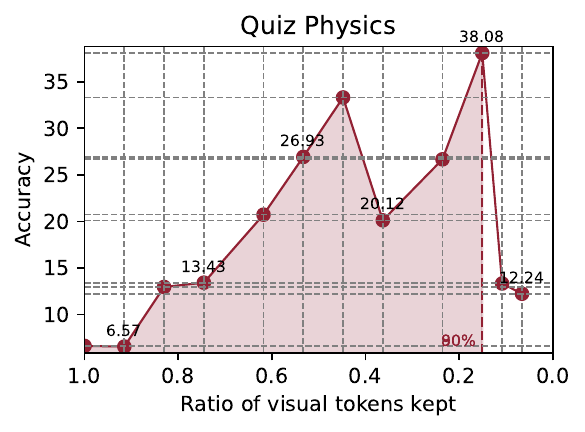}} &
        \subfloat{\includegraphics[width=0.3\textwidth]{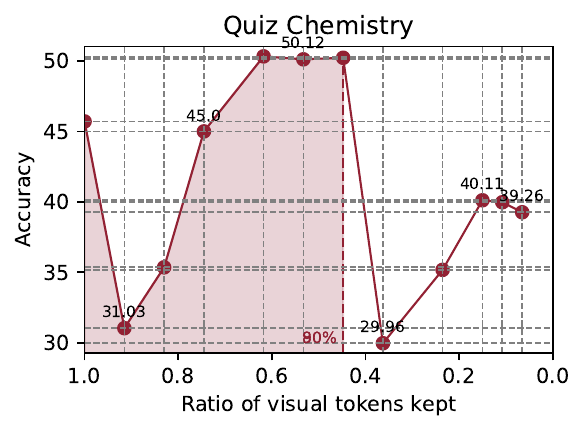}} \\
        \subfloat{\includegraphics[width=0.3\textwidth]{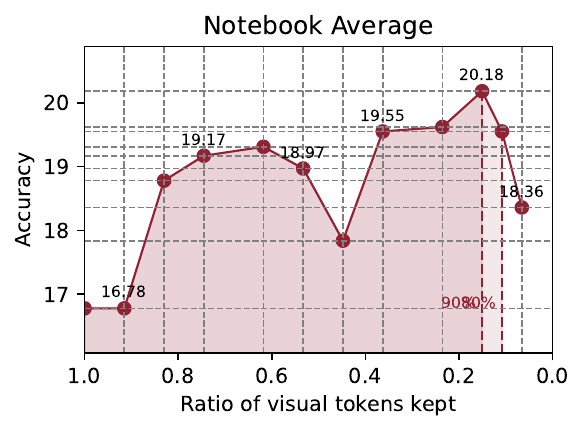}} &
        \subfloat{\includegraphics[width=0.3\textwidth]{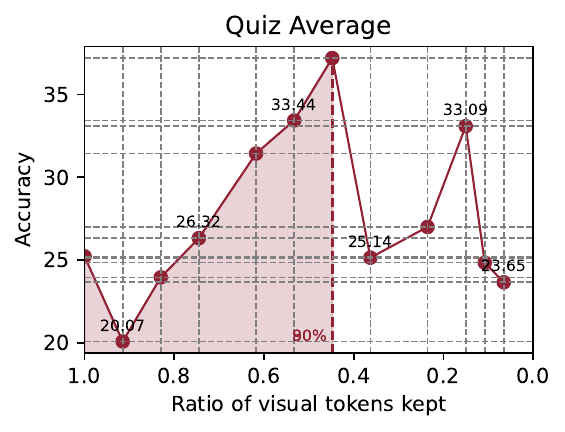}} &
        \subfloat{\includegraphics[width=0.3\textwidth]{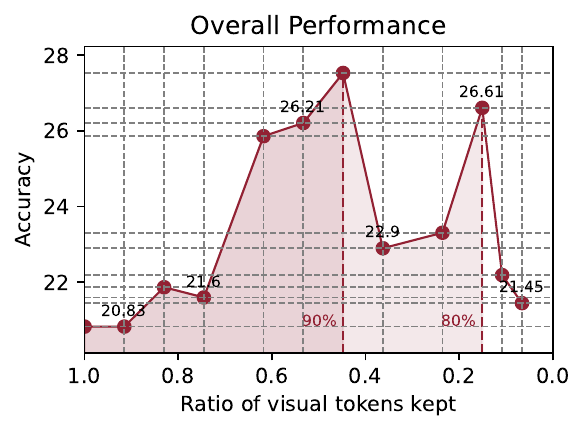}} \\
    \end{tabular}
    \caption{Ablation study of token merging in AuroraCap~\cite{chai2024auroracap} on~\benchmark. }
    \label{fig:vdc_tome_ablation}
\end{figure*}

We supplement Table~\ref{tab:token} with the performance of various visual-token reduction models on video detailed captioning and reasoning-based QA tasks in Video-MMLU. As a core design of AuroaCap~\cite{chai2024auroracap}, token merging plays a crucial role in reducing visual token redundancy. Therefore, we systematically test AuroraCap~\cite{chai2024auroracap} under different visual token kept ratios to analyze how varying compression levels affect video detailed captioning capability to understand its impact, where the number of remaining visual tokens per frame varies from 49 to 730.

Figure~\ref{fig:vdc_tome_ablation} illustrates the impact of visual token reduction on performance across video detailed captioning and reasoning-based QA tasks in different disciplines. Overall, most models retain over 80\% of their peak performance even when keeping only 20\%–40\% of visual tokens, suggesting that significant token compression is feasible without severe degradation. However, performance does not follow a strict monotonic trend, indicating that the sequential nature of videos introduces additional complexity in token merging, leading to non-trivial effects on performance.

When comparing captioning and QA tasks, we observe that captioning performance remains relatively stable across different token kept ratios, particularly in mathematics and chemistry. This suggests that structural elements like formulas and static visual cues are more resilient to compression. In contrast, QA performance, especially in physics and chemistry, exhibits sharp fluctuations, highlighting the greater sensitivity of reasoning-based tasks to token reduction. The varying performance across disciplines further reinforces that subjects relying on dynamic visual elements, such as physics and chemistry, require a higher number of retained tokens to maintain accuracy.

Interestingly, optimal performance does not always occur at the highest kept ratio but rather at a mid-range level. This indicates that moderate token merging can improve efficiency without significantly compromising performance, but excessive compression leads to information loss, particularly in reasoning-heavy tasks. Since AuroraCap~\cite{chai2024auroracap} primarily focuses on spatial token merging without explicitly addressing temporal dependencies, these results suggest that additional strategies are needed to optimize token merging for time-dependent visual reasoning. The results also imply that different tasks and disciplines may benefit from customized token compression strategies rather than a one-size-fits-all approach.

\begin{figure*}[!ht]
    \centering
    \begin{tabular}{ccc}
        \subfloat{\includegraphics[width=0.48\textwidth]{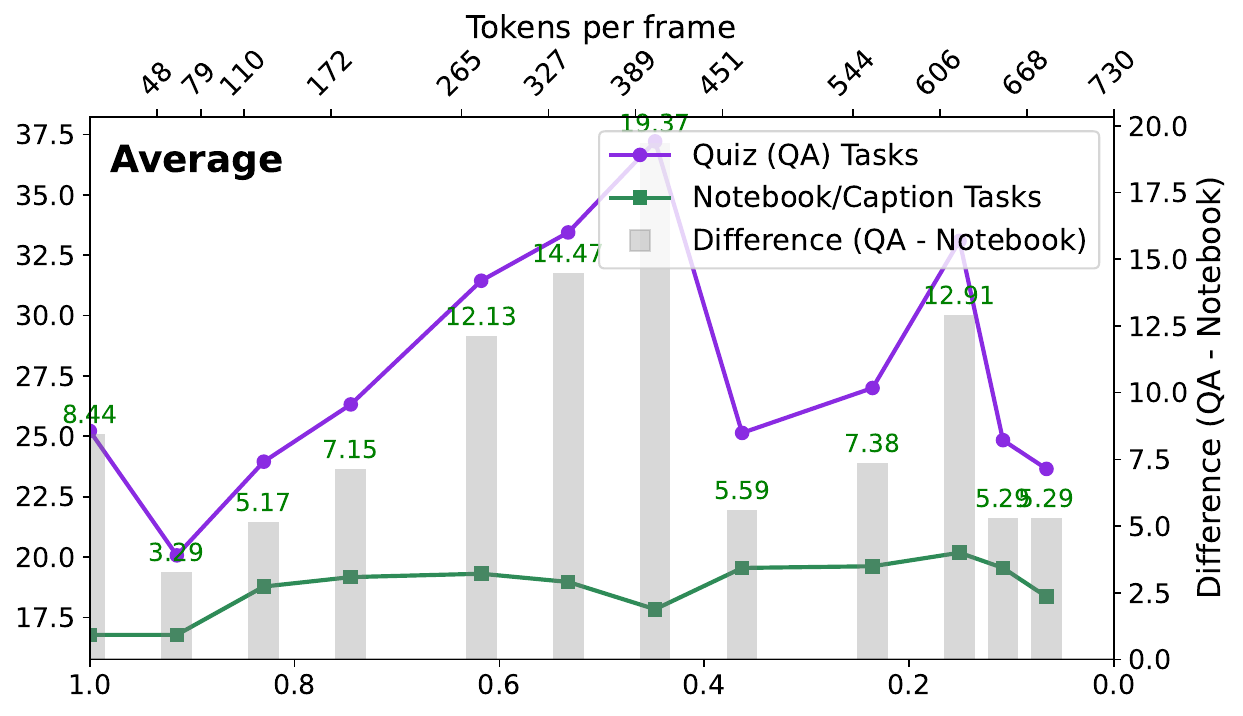}} &
        \subfloat{\includegraphics[width=0.48\textwidth]{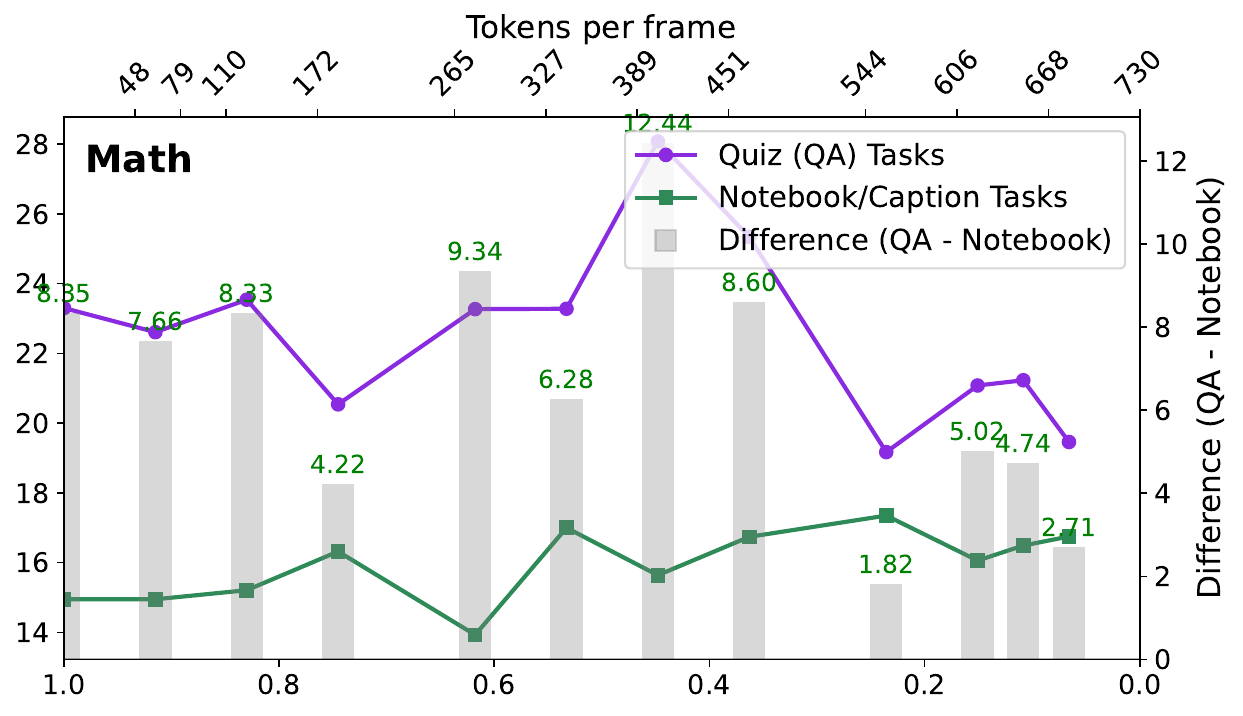}} \\
        \subfloat{\includegraphics[width=0.48\textwidth]{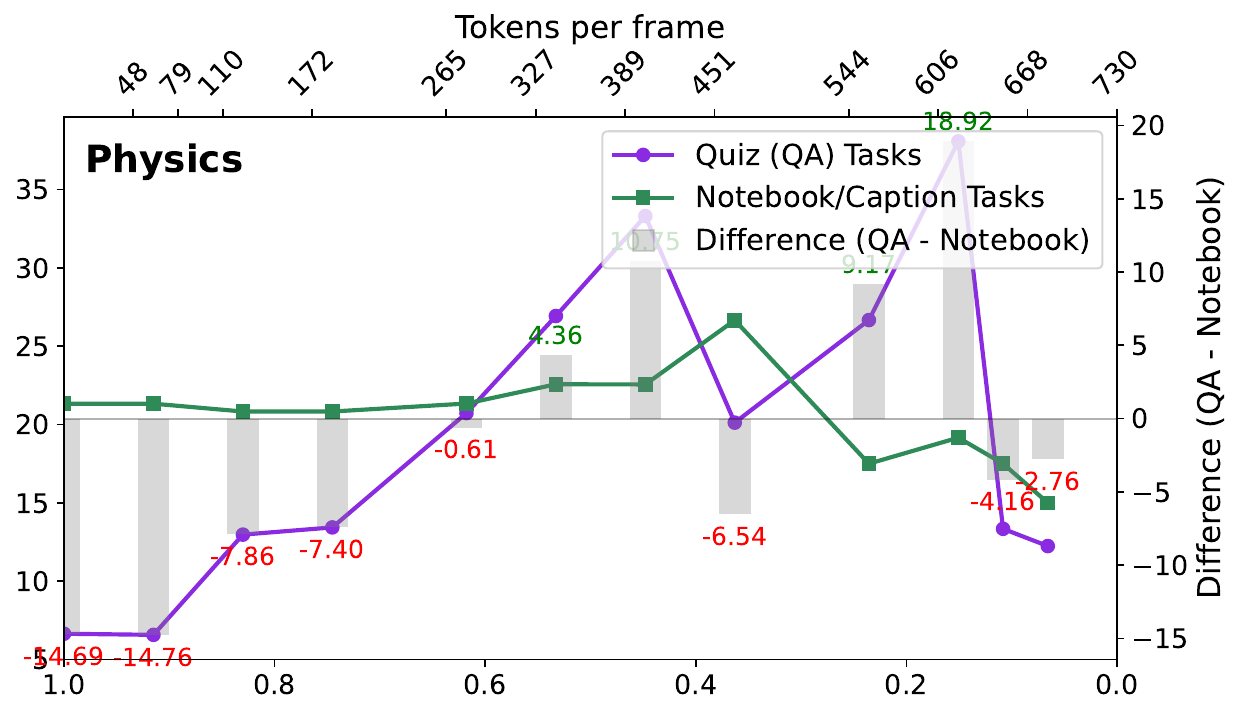}} &
        \subfloat{\includegraphics[width=0.48\textwidth]{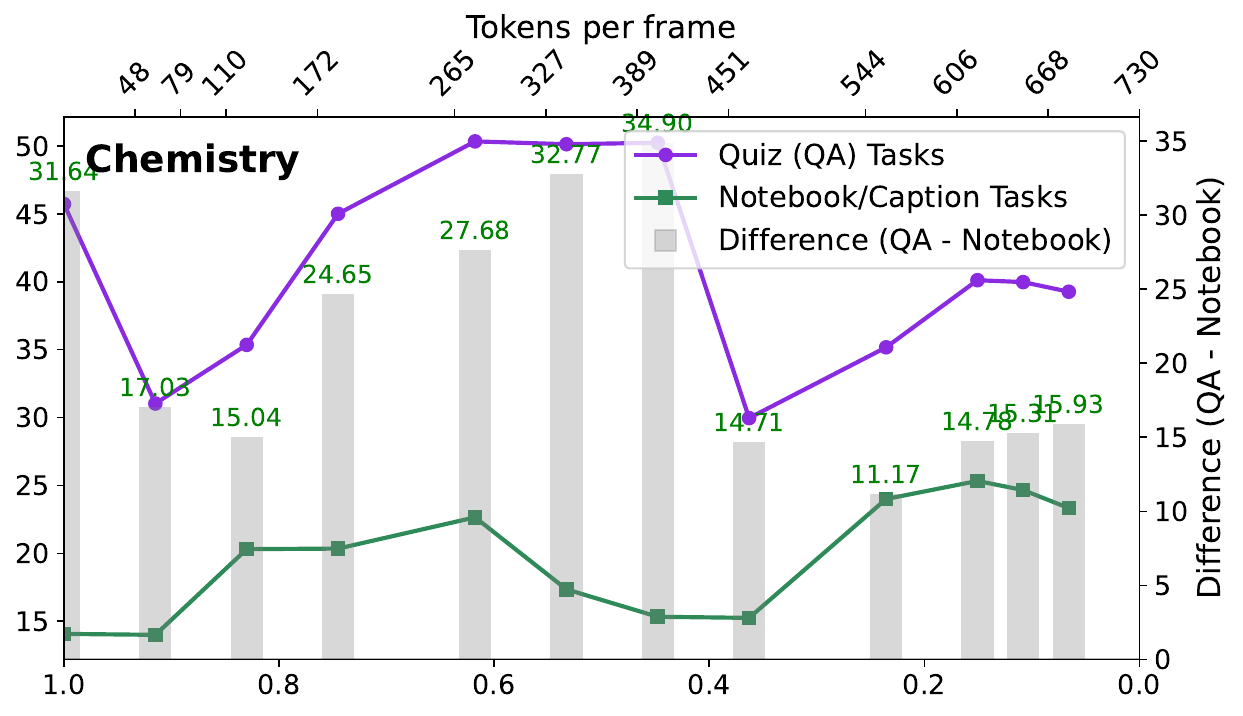}} \\
    \end{tabular}
    \caption{Performance comparsion of token merging in AuroraCap~\cite{chai2024auroracap} on~\benchmark across different discipline. }
    \label{fig:dis_tome}
\end{figure*}

We also compare its effectiveness in video detailed captioning and reasoning-based QA tasks across different desciplines in Figure~\ref{fig:dis_tome}. 
In the overall performance plot, QA scores (purple line) generally exceed captioning scores (green line) across most token kept ratios, with the largest gap (gray bars) appearing at mid-range ratios (0.4–0.6). Although AuroraCap~\cite{chai2024auroracap} is designed as a video captioning model, its captioning performance on~\benchmark~lags behind its QA performance. We believe that the results is relative to its token merging strategy, which is based on token similarity. In lecture videos, crucial text patches containing key information may be merged due to embedding similarity, leading to degraded caption quality. Reasoning tasks initially benefit from retaining more visual tokens but do not necessarily improve at the highest kept ratios. However, when extreme compression is applied (ratios below 0.2), both QA and captioning performance drop sharply, reinforcing the importance of maintaining a sufficient number of tokens.
Across different disciplines, mathematics shows a relatively small gap between QA and captioning across all token ratios, suggesting that both tasks require a similar level of visual information. In physics, QA performance starts significantly lower than captioning at high token counts but surpasses it as fewer tokens are retained, indicating that sequential reasoning in physics may require a more refined token selection strategy. Chemistry consistently shows the widest gap favoring QA, particularly at mid-range token ratios, suggesting that chemistry reasoning benefits more from retaining structured visual elements and textual annotations.

To provide a more intuitive comparison, we present AuroraCap~\cite{chai2024auroracap}’s video captioning results at different token kept ratios on a video about the Arithmetic Mean-Root Mean Square Inequality demonstration. We observe that as the visual token kept ratio increases (retaining more tokens), the generated captions become progressively shorter.

\begin{figure*}[!ht]
    \centering
    \begin{tabular}{ccccc}
        \subfloat{\includegraphics[width=0.23\textwidth]{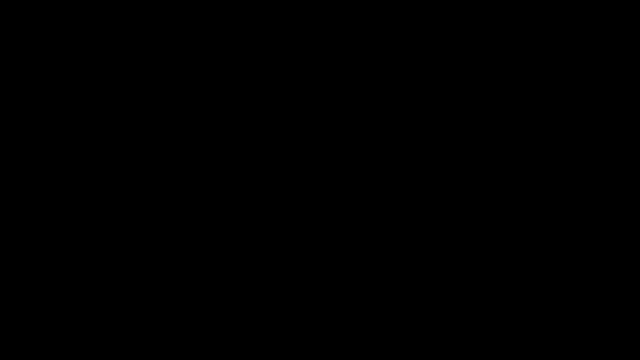}} &
        \subfloat{\includegraphics[width=0.23\textwidth]{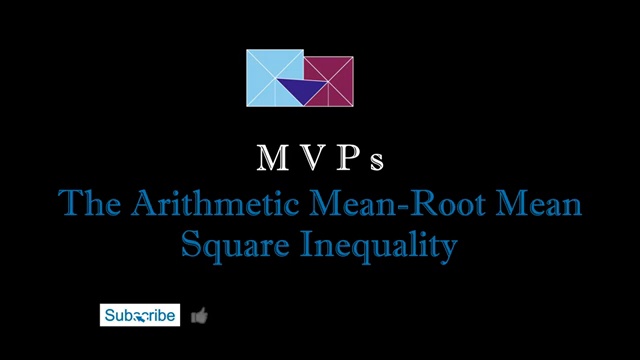}} &
        \subfloat{\includegraphics[width=0.23\textwidth]{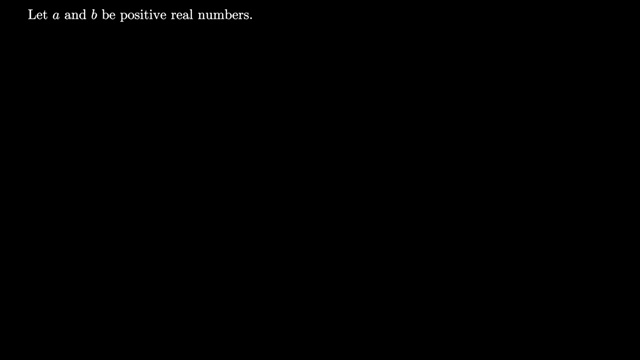}} &
        \subfloat{\includegraphics[width=0.23\textwidth]{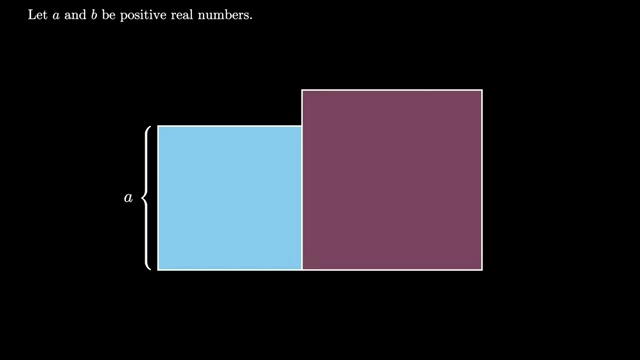}} \\
        \subfloat{\includegraphics[width=0.23\textwidth]{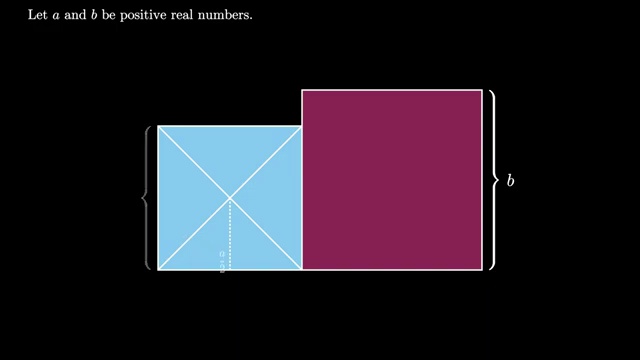}} &
        \subfloat{\includegraphics[width=0.23\textwidth]{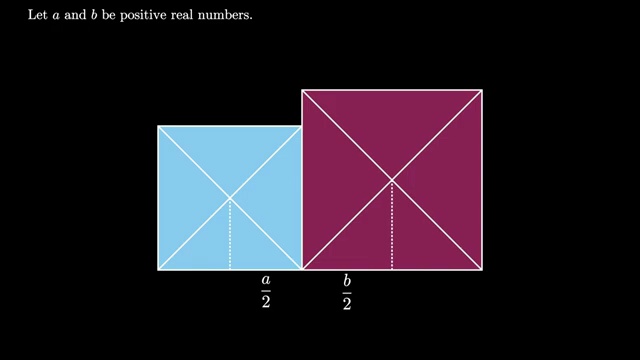}} &
        \subfloat{\includegraphics[width=0.23\textwidth]{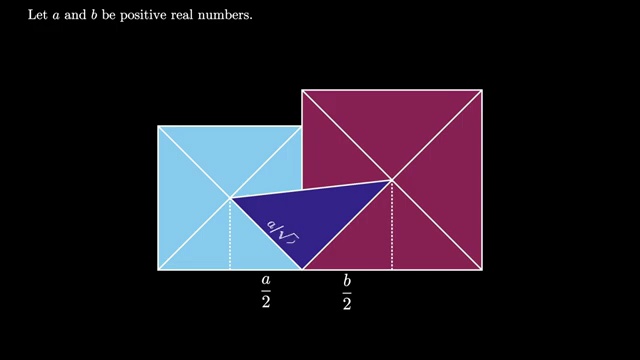}} &
        \subfloat{\includegraphics[width=0.23\textwidth]{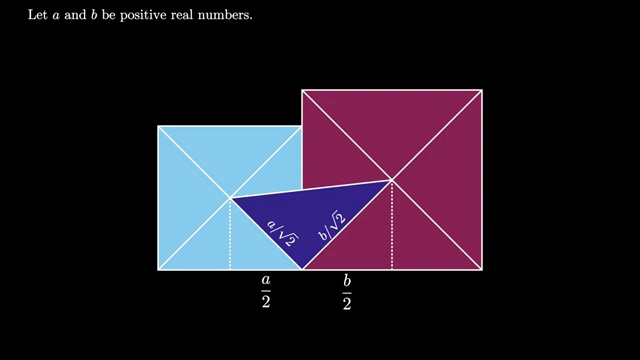}} \\
        \subfloat{\includegraphics[width=0.23\textwidth]{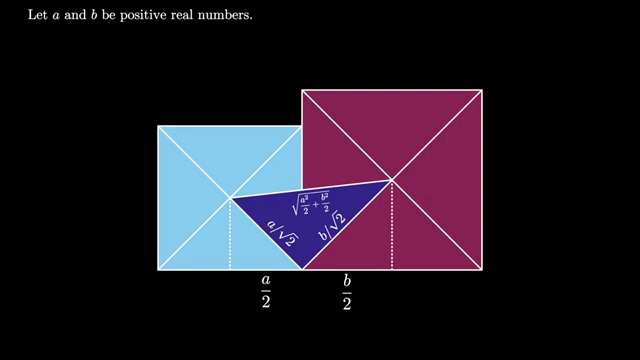}} &
        \subfloat{\includegraphics[width=0.23\textwidth]{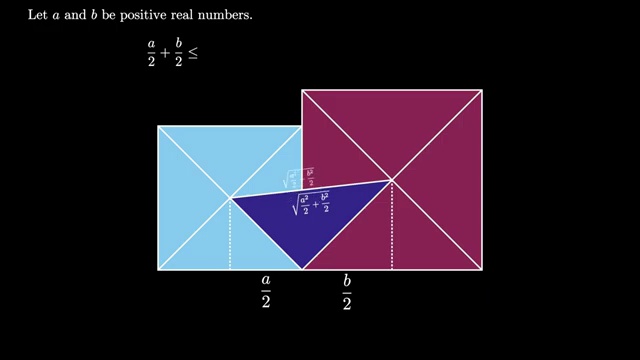}} &
        \subfloat{\includegraphics[width=0.23\textwidth]{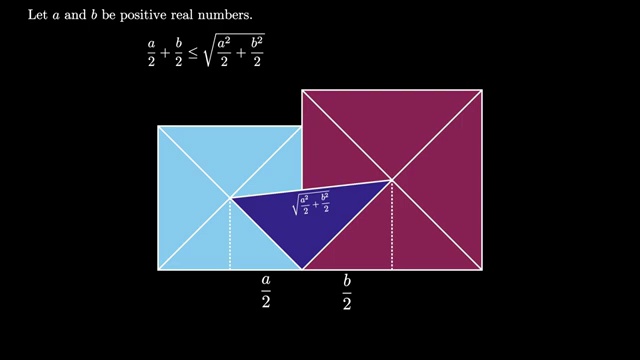}} &
        \subfloat{\includegraphics[width=0.23\textwidth]{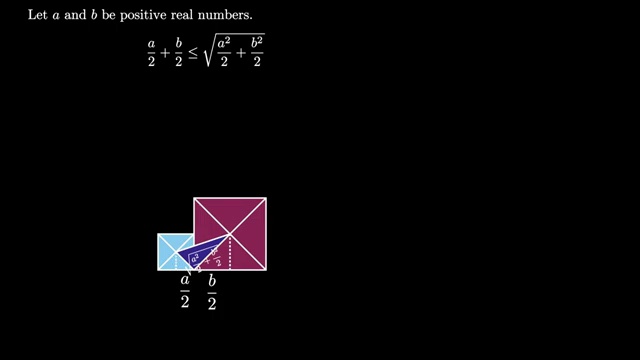}} \\
        \subfloat{\includegraphics[width=0.23\textwidth]{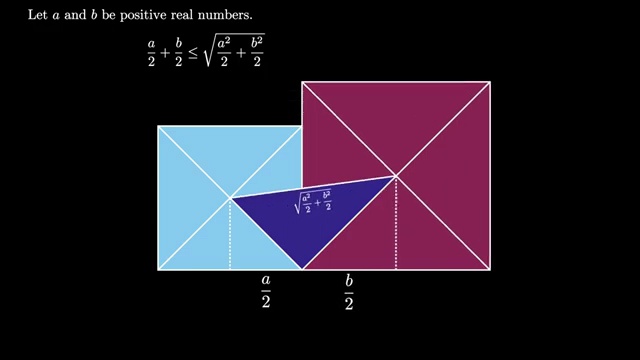}} &
        \subfloat{\includegraphics[width=0.23\textwidth]{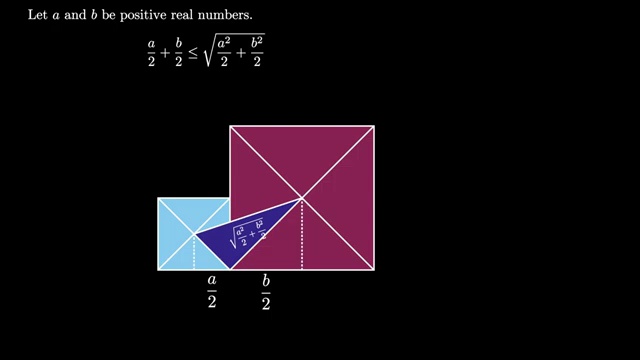}} &
        \subfloat{\includegraphics[width=0.23\textwidth]{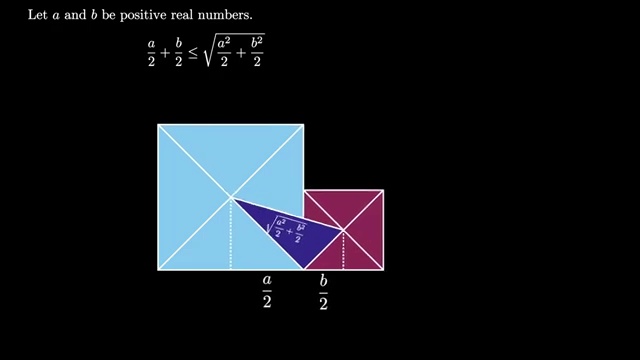}} &
        \subfloat{\includegraphics[width=0.23\textwidth]{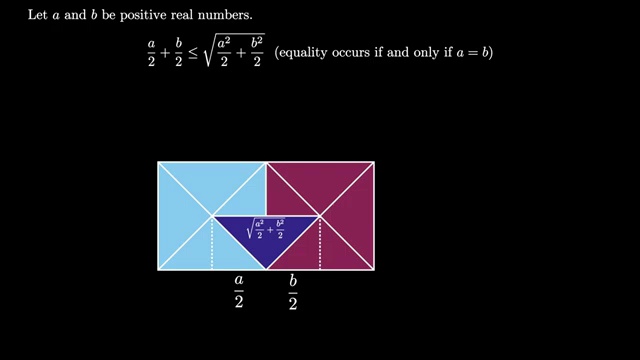}} \\
        \subfloat{\includegraphics[width=0.23\textwidth]{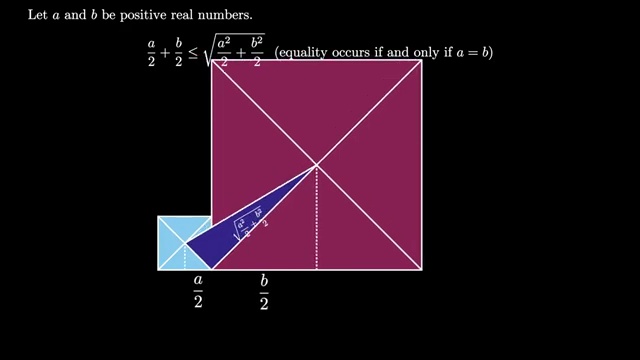}} &
        \subfloat{\includegraphics[width=0.23\textwidth]{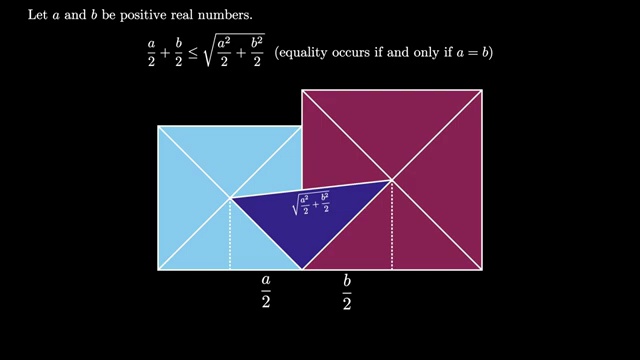}} &
        \subfloat{\includegraphics[width=0.23\textwidth]{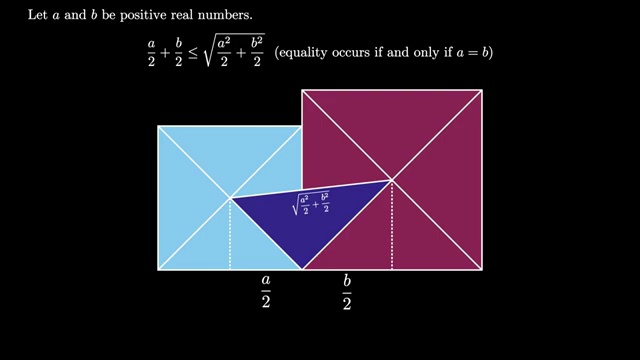}} &
        \subfloat{\includegraphics[width=0.23\textwidth]{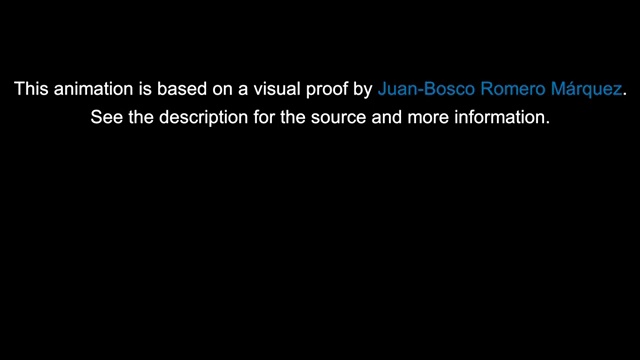}} \\
    \end{tabular}
    \caption{Example video frames in~\benchmark. The video YouTubeID is $9Qyba_swKTI$, which is focus on the demonstration of the Arithmetic Mean-Root Mean Square Inequality, provided by \textit{Mathemativs Visual Proofs}.  }
    \label{fig:auroracap}
\end{figure*}

\begin{itemize}[leftmargin=12.5mm]
\setlength{\itemsep}{2pt}
    \item[\textbf{\# Token}] \textbf{Describe this video in detail.~(Figure~\ref{fig:auroracap})}
    \vspace{3pt}
    \small{
    \item[\textbf{GT}] The educational video opens with a distinctive title screen displaying \textbf{``MVPs''} and \textbf{``The Arithmetic Mean-Root Mean Square Inequality''} against a dark background, accompanied by a geometric logo in blue and purple tones and a subscribe button with a thumbs-up icon.

    The mathematical presentation establishes its fundamental premise by introducing two positive real numbers, $a$ and $b$, displayed against a black background. The geometric construction begins with two adjacent squares: a blue square with side length $a$ and a maroon square with side length $b$. This fundamental construction serves as the foundation for demonstrating a profound relationship between arithmetic and quadratic means. Each square is methodically divided by diagonal lines intersecting at their respective centers, creating four congruent triangular sections. This division is crucial as it creates isosceles right triangles within each square, with legs measuring $a/2$ in the blue square and $b/2$ in the maroon square.
    
    The geometric visualization advances by connecting the center points of the two squares, forming a right triangle with legs measuring $a/\sqrt{2}$ and $b/\sqrt{2}$. These measurements arise from the fact that these segments are hypotenuses of the isosceles right triangles formed within each square. Through the application of the Pythagorean theorem to this connecting triangle, the hypotenuse measures $\sqrt{a^2/2 + b^2/2}$. The construction creates a trapezoid formed by two isosceles right triangles and the connecting triangle, which proves fundamental to establishing the inequality: $\frac{a}{2} + \frac{b}{2} \leq \sqrt{\frac{a^2}{2} + \frac{b^2}{2}}.$
    
    The proof, credited to Juan-Bosco Romero Márquez, elegantly demonstrates that equality occurs if and only if $a = b$. This condition becomes geometrically evident as the squares would be identical in this case, creating a symmetric configuration. Throughout the presentation, the squares maintain their distinct blue and maroon/purple coloring, with their dimensions consistently labeled. When the squares overlap, the resulting purple-shaded intersection area provides additional visual insight into the mathematical relationship.
    
    The visualization effectively combines algebraic expressions with geometric representations, demonstrating how the arithmetic mean (represented by $(a/2 + b/2)$) relates to the root mean square ($\sqrt{a^2/2 + b^2/2}$) through the careful arrangement and interaction of colored squares and triangles. This geometric approach provides a compelling visual proof of the arithmetic mean-root mean square inequality, showing that the arithmetic mean of two positive numbers is always less than or equal to their root mean square, with equality occurring only when the numbers are identical.
    
    Multiple variations of the core geometric arrangement are presented, each emphasizing different aspects of the inequality. The black background provides strong contrast for the blue, maroon, and purple geometric elements, ensuring the mathematical relationships remain clear and visually striking. Every frame builds upon the previous understanding, with careful attention to labeling, color coding, and the precise geometric relationships between the squares and triangles, maintaining consistent use of variables throughout while demonstrating their relationships both algebraically and geometrically.
    
    \item[48] The video presents a vibrant and engaging exploration of mathematical concepts, focusing on the properties of triangles. It begins with a bold statement, "A triangle is a real number," setting the stage for a deep dive into the world of geometric shapes. The scene transitions to a black background, where a large, brightly colored triangle dominates the frame. The triangle is divided into two distinct sections, each filled with a different color, creating a striking visual contrast.
    
    As the camera pans across the triangle, it reveals a series of mathematical equations and expressions that highlight the relationship between the lengths of the sides and the angles of the triangle. The text, "A triangle is a real number. (equivalently, if f is a function, then f(a) is a real number)," appears, further emphasizing the connection between mathematical concepts and real-world applications. The camera then zooms in on the triangle, allowing viewers to appreciate the intricate details of the shapes and the precise placement of the mathematical notations.
    
    The video continues to explore the triangle's properties, showcasing various angles and their corresponding expressions. The camera pans slowly, allowing viewers to absorb the information and appreciate the beauty of the geometric shapes. The background transitions from black to a gradient of blue and purple, enhancing the visual appeal and drawing attention to the mathematical content.
    
    As the video progresses, it delves deeper into the triangle's attributes, including its area and the sum of its angles. The camera captures the angles with precision, highlighting their significance in the triangle's structure. The text, "A triangle is a real number. (equivalently, if f is a function, then f(a) is a real number)," reappears, reinforcing the connection between mathematical concepts and real-world applications.
    
    The video concludes with a thoughtful reflection on the triangle's significance, emphasizing its role as a fundamental shape in mathematics. The camera zooms in on the triangle, allowing viewers to appreciate the intricate details of the shapes and the precise placement of the mathematical notations. The background transitions to a gradient of blue and purple, enhancing the visual appeal and drawing attention to the mathematical content.
    
    Throughout the video, the camera work is smooth and deliberate, allowing viewers to absorb the information and appreciate the beauty of the geometric shapes. The use of color, lighting, and camera angles creates a visually engaging experience that highlights the importance of mathematical concepts in our daily lives.
    
    \item [79] The video presents a vibrant and dynamic scene set in a lively urban environment. The main focus is on a bustling street filled with a diverse crowd of people, each engaged in various activities. Some are walking briskly, while others are seen enjoying leisurely strolls, creating a lively atmosphere. The street is adorned with colorful buildings, each showcasing unique architectural styles, and the atmosphere is enhanced by the presence of street vendors and parked cars.
    
    As the camera pans through the scene, it captures the essence of city life, highlighting the interplay of movement and stillness. The sound of chatter, laughter, and the occasional car horn fills the air, enhancing the sense of realism. The camera work is smooth, transitioning seamlessly between different angles and perspectives, allowing viewers to appreciate the depth and complexity of the urban landscape.
    
    The video transitions to a cozy indoor setting, featuring a warmly lit living room where a group of friends are gathered, engaged in animated conversation. The atmosphere is inviting, with soft lighting and comfortable furniture, creating a sense of intimacy and camaraderie. The friends exhibit a range of emotions, from laughter to contemplation, reflecting the dynamic nature of human interaction.
    
    As the video progresses, it shifts focus to a serene outdoor setting, where a person is seen enjoying a peaceful moment by the water. The tranquil environment contrasts with the earlier bustling street scene, providing a calming visual experience. The camera captures the beauty of the natural surroundings, with gentle ripples on the water and lush greenery in the background, evoking a sense of calm and reflection.
    
    The video concludes with a return to the urban environment, showcasing a vibrant street scene filled with people, vehicles, and colorful buildings. The camera work is dynamic, capturing the essence of city life with a mix of close-up shots and wider angles, allowing viewers to appreciate the intricate details of the urban landscape. The sound of traffic, chatter, and laughter fills the air, encapsulating the essence of a lively, dynamic city.

    \item [110] The video presents a series of mathematical diagrams and explanations, focusing on the concept of real numbers. The initial frame displays a large, bold text that reads, "In this video, we explore the basics of real numbers." This sets the stage for an educational series that aims to delve into the intricacies of mathematical concepts.
    
    As the video progresses, the visuals transition to a vibrant display of geometric shapes and lines, creating a striking contrast against a dark background. The shapes are primarily triangles, with one large triangle in a deep purple color and a smaller one in a lighter shade, both outlined in white. The triangles are interconnected by lines, forming a complex network that suggests a visual representation of mathematical relationships.
    
    The narrative is enhanced by a clear, informative text overlay that explains the significance of the shapes and their connections. The text is presented in a clean, white font that stands out against the dark backdrop, ensuring readability. The combination of visual elements and text creates a dynamic learning experience, making the subject matter accessible and engaging.
    
    Throughout the video, the camera work is smooth, with a steady focus on the shapes and text, allowing viewers to absorb the information without distraction. The transitions between frames are seamless, maintaining a cohesive flow that keeps the viewer's attention on the educational content.
    
    As the video nears its conclusion, the text shifts to emphasize the importance of understanding the source of real numbers, as it is the foundation of all mathematical concepts. The final frame features a concluding statement that reinforces the significance of the topic, leaving viewers with a deeper appreciation for the intricate world of real numbers.

    \item [172] The video presents a series of mathematical diagrams and text, focusing on the concept of real numbers and their properties. The initial frame features a large, bold text that reads, "This is a visual proof by Juan-Borrego of the real numbers." Below this text, a geometric figure is displayed, consisting of a large square with a smaller square inside it, creating a visual representation of the concept of a real number line. The figure is color-coded, with the larger square in blue and the smaller square in purple, with a dotted line indicating the real number line.
    
    As the video progresses, the focus shifts to a more complex geometric figure, where the same square and line are now enclosed within a larger square, creating a three-dimensional perspective. The text below this frame reads, "The description for the source information." This frame is followed by a mathematical expression, "a = b = 2," which is highlighted in blue, indicating the equality of the variables 'a' and 'b' with the number '2'. The next frame introduces a new element, a right-angled triangle with the hypotenuse labeled 'a' and the legs labeled 'b' and 'c'. The text below this frame states, "The description for the source information."
    
    The video then transitions to a frame where the triangle is rotated, revealing a new perspective. The text below this frame reads, "The description for the source information." The final frame presents a concluding statement, "Juan-Borrego's proof of the real numbers." The text is accompanied by a mathematical expression, "a = b = c = 2," which is highlighted in blue, emphasizing the equality of the variables 'a', 'b', and 'c' with the number '2'. The video concludes with a frame that features a large, bold text stating, "This is a visual proof by Juan-Borrego of the real numbers."
    
    Throughout the video, the background is a solid black, providing a stark contrast to the vibrant colors of the geometric figures and text. The text is presented in a clear, sans-serif font, ensuring readability. The overall layout is organized and methodical, guiding the viewer through the mathematical concepts being presented.

    \item [265] The video presents a mathematical explanation focusing on the concept of real numbers, as described by the visual elements of a geometric figure. The figure is a large square divided into smaller squares, with a diagonal line creating a right-angled triangle. The triangle is highlighted in blue, with its hypotenuse labeled as 'a' and the legs as 'b' and 'c'. The square is labeled 'a' and 'b', while the smaller squares are labeled 'a' and 'b' as well. The background is a deep blue, providing a contrast that emphasizes the geometric shapes.
    
    At the top of the video, there is a text overlay in white that reads, "This is based on a visual proof by Juan-Bosco Rodriguez." Below this, in a larger font, the text states, "The real numbers are based on a visual proof by Juan-Bosco Rodriguez." The text is clear and legible, set against the dark blue background. The overall layout is clean and organized, with the geometric figure and text providing a clear visual representation of the mathematical concept being explained.

    \item [327] The video presents a mathematical concept, specifically focusing on the relationship between the number of real numbers and their representation. It features a geometric diagram with a large purple square and a smaller blue square, both sharing a common side. The purple square is divided into four smaller triangles, each labeled with a number from 1 to 4, while the blue square is divided into two triangles, labeled with 'a' and 'b'. The triangles are arranged in a way that suggests a visual proof of a mathematical concept, with the purple square representing the set of real numbers, while the blue square represents a specific subset of these numbers. The video is set against a black background, emphasizing the vibrant colors of the geometric shapes. The text overlay, in white font, provides context to the visual, stating that the number of real numbers is based on a visual proof by Juan-Bosco Rojas, and it describes the source of information as more informative.

    \item [389] The video presents a mathematical concept, focusing on the relationship between the number of real numbers and their representation. It features a geometric diagram with a large purple square and a smaller blue square, both sharing a common side. The purple square is divided into smaller triangles, each labeled with a mathematical expression. The expressions include the square root of a number, a fraction, and a variable 'b'. The blue square is a smaller representation of the purple square, with a similar division into triangles, each labeled with a different expression. The expressions are mathematical in nature, involving the square root of 'a' and 'b', a fraction 'a/b', and a variable 'b'. The video is educational, aiming to explain the concept of real numbers and their visual representation.

    \item [451] The video presents a series of geometric shapes and mathematical expressions, primarily focusing on the concept of real numbers. It begins with a large blue triangle, labeled 'a', which is divided into two smaller triangles, 'b' and 'c', creating a visual representation of the number line. The number '2' is placed at the bottom of 'a', while '1' is at the top, indicating the scale of the number line. The shapes are set against a black background, enhancing their visibility.
    
    A mathematical expression, '(2a + b) = a + b', is displayed above the triangles, suggesting a relationship between the lengths of the sides. The expression is highlighted in white text, contrasting with the dark background. The video transitions to a purple square, 'd', with a smaller square 'e' inside it, creating a visual representation of the square root symbol. The expression '\u221a(2a) = a' is shown, indicating the square root property. The purple square is labeled 'd', while 'e' is the square root symbol. The video concludes with a text overlay in white, stating, "Juan-Bosco Roque based on a visual proof by Jian-Bosco Roque," which credits the creator of the visual representation.

    \item [544] The video presents a mathematical concept, focusing on the properties of triangles and their relationships with real numbers. It begins with a title that reads, "Juan-Bosco Ro." Below the title, a statement explains that the video is based on a visual proof by Juan-Bosco Ro. The main visual element is a geometric diagram featuring a large triangle with a smaller triangle inside it, both sharing a common side. The larger triangle is colored in shades of blue, while the smaller one is in purple. The shared side is highlighted with a dotted line, indicating its significance.
    
    The video then transitions to a black background where the text continues, stating that the description is based on the source and more information can be found. The text is white, providing a clear contrast against the dark backdrop. The overall layout is clean and organized, with the text and diagram clearly separated, allowing for easy comprehension of the mathematical content.

    \item [606] The video presents a mathematical concept, focusing on the relationship between the number of real numbers and their representation. It features a geometric diagram with a large purple square and a smaller blue square, both sharing a common side. The purple square is divided into four smaller squares, each labeled with a number from 1 to 4. The blue square is also divided into four smaller squares, with the top left square labeled 'a' and the bottom left 'b'. The video explains that the number of real numbers is based on a visual proof by Juan-Bosco, which is described as a source for more information. The text is overlaid on a black background, enhancing readability. The video is educational, aiming to convey the mathematical concept of real numbers through visual representation.

    \item [668] The video presents a vibrant and dynamic scene featuring a bustling cityscape at dusk. The sky is painted with warm hues of orange and pink, transitioning into a deep blue as the sun sets. Skyscrapers with illuminated windows rise against the backdrop of the fading daylight, creating a striking contrast. The streets below are alive with the movement of people, vehicles, and the glow of streetlights. The atmosphere is further enhanced by the presence of a river reflecting the city lights, adding a serene touch to the lively urban environment.

    }
\end{itemize}
\vspace{3pt}

\end{document}